\documentclass[11pt]{article}

\usepackage[margin=1in]{geometry}
\usepackage{amsmath,amssymb,amsthm,mathtools,bm,algorithm,algpseudocode}
\usepackage{booktabs}
\usepackage{longtable}
\usepackage{subcaption}
\usepackage{enumitem}
\usepackage[colorlinks=true,linkcolor=blue,citecolor=blue,urlcolor=blue]{hyperref}
\usepackage{natbib}
\usepackage{authblk}

\newtheorem{theorem}{Theorem}

\newtheorem{remark}{Remark}
\newtheorem{proposition}{Proposition}
\newtheorem{corollary}{Corollary}
\newtheorem{assumption}{Assumption}

\def\argmin{\mathop{\rm argmin}}

\def\bfx{\mathop{\bf x}}
\def\bfx{\mathop{\bf X}}

\def\bfx{\boldsymbol{x}}

\def\bfu{\boldsymbol{u}}

\newcommand {\bfphi} {\mbox{\boldmath $\phi$}}

\newcommand {\bfw} {\mbox{\boldmath $\omega$}}

\title{Adaptive Regularization for Random Features: A Neighboring Early-Stopping Rule with Oracle-Rate Guarantees}
\author{Caixing Wang\textsuperscript{a}, Zhibo Chen\textsuperscript{a} and
Yue Wang\textsuperscript{b}}
\affil{\textsuperscript{a}School of Statistics and Data Science, Southeast University\\
\textsuperscript{b}Faculty of Business for Science and Technology, School of Management, University of Science and Technology of China}

\date{}
\begin{document}
\maketitle

\begin{abstract}
Random feature methods provide a scalable approximation to kernel ridge regression (KRR), but the regularization parameter that yields the oracle learning rate depends on unknown smoothness and capacity parameters. In this work, we propose a neighboring early-stopping rule for adaptive regularization in KRR with random features (KRR-RF). The method uses a grid that is uniform in inverse regularization and compares only adjacent estimators, reducing the number of discrepancy comparisons relative to standard all-pairs Lepskii-type procedures. Both the neighboring discrepancy and its empirical complexity term can be computed directly in the random feature space, without constructing the exact kernel Gram matrix.

We establish a high-probability comparison bound for neighboring KRR-RF estimators and show that, under standard source and capacity conditions together with suitable grid and random feature budget conditions, the selected estimator attains the oracle polynomial learning rate up to logarithmic factors. The result allows the regularization parameter to be selected without prior knowledge of the source and capacity exponents and covers both well-specified and partially misspecified regimes. Our analysis is based on an empirical random feature effective dimension that connects the observable stopping threshold with the population complexity of the random feature model. Simulation and real-data experiments illustrate the prediction performance and computational behavior of the proposed method in comparison with standard tuning procedures.
\end{abstract}

\noindent%

\section{Introduction}\label{sec-intro}

Kernel methods are a fundamental class of nonparametric learning techniques that have played a central role in statistical learning theory and machine learning \cite{wahba1990spline,hofmann2008kernel,scholkopf2002learning}. By embedding data into a reproducing kernel Hilbert space (RKHS), they enable flexible nonlinear modeling while preserving a tractable linear structure in the feature space. Specifically, let $\bfx, \bfx^{\prime} \in \mathcal{X}$ be two input samples, and $\psi(\cdot): \mathcal{X} \rightarrow \mathcal{H}$ be a nonlinear feature map that transforms the input space $\mathcal{X}$ to the RKHS $\mathcal{H}$. The associated kernel function is defined by $K(\bfx, \bfx^{\prime})=\langle \psi(\bfx), \psi(\bfx^{\prime}) \rangle_{\mathcal{H}}$, which implicitly characterizes the inner product in $\cal H$. This allows one to perform linear operations in a potentially high- or infinite-dimensional feature space without explicitly evaluating the feature map. Kernel methods have been successfully applied to a wide range of problems, including regression \cite{smale2007learning,caponnetto2007optimal}, classification \cite{steinwart2008support,camps2005kernel}, and clustering \cite{camastra2005novel,zhao2009multiple}. Despite their theoretical elegance and practical success, kernel methods often suffer from computational challenges when applied to large-scale datasets \cite{sonnenburg2006large}. The computational complexity of kernel methods typically scales quadratically or cubically with the number of samples, making them infeasible for large datasets \cite{zhang2015divide,rudi2015less,rudi2017falkon}. 

\begin{figure}
    \centering
    \includegraphics[width=\textwidth]{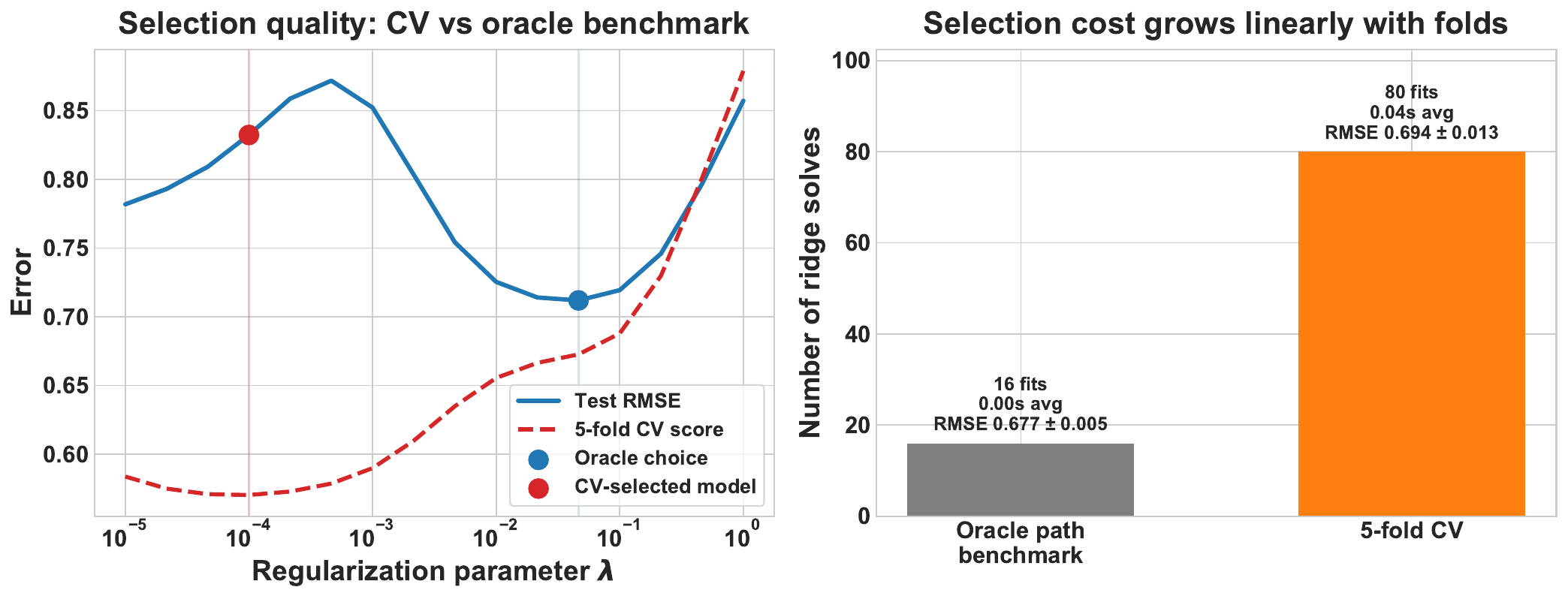}
    \caption{Left: The results for KRR-RF over 16 candidate values of $\lambda$. The blue curve is the test RMSE on an independent test set after training on the full sample and its minimizer is the oracle choice. The red dashed curve is the 5-fold cross-validation score, obtained by refitting the model on training folds and evaluating it on validation folds. Cross-validation selects a smaller $\lambda$ and incurs a larger test error than the oracle choice. Right: averaged over 12 trials, the oracle path benchmark requires 16 solves, while 5-fold cross-validation requires 80 solves. Despite this 5× higher cost, cross-validation still gives a worse average test RMSE (0.694 ± 0.013 vs. 0.677 ± 0.005).
}
    \label{fig:comparison}
\end{figure}

Random feature methods address this challenge by replacing the implicit kernel representation with an explicit low-dimensional randomized feature map, leading to scalable approximations of kernel algorithms \cite{rahimi2007random,rudin2017Fourier}. A large body of work has been devoted to understanding the statistical properties of random feature methods, including their approximation capabilities \cite{rahimi2007random,bach2017equivalence}, generalization performance \cite{rudi2017generalization,li2021towards,wang2026generalization}, and fast computation \cite{avron2017faster,ma2025generalization}. In the context of kernel ridge regression with random features (KRR-RF), it has been shown that under suitable source conditions on the target function and capacity conditions on the RKHS, KRR-RF can achieve minimax optimal rates of convergence \cite{rudi2017generalization,li2021towards}. However, these optimal results typically require careful tuning of the regularization parameter, which depends on unknown smoothness parameters of the target function and capacity parameters of the RKHS. In practice, the regularization parameter is often selected by cross-validation (CV) or grid search methods, where the regularization parameter is chosen by minimizing a validation criterion over a prespecified grid \cite{browne2000cross,gyorfi2002distribution,caponnetto2010cross}. However, such approaches face several limitations: first, they are computationally expensive, as they require refitting the model multiple times for different regularization parameters; second, they can be sensitive to the choice of the search grid and may not guarantee optimal performance if the grid is not sufficiently fine or does not cover the optimal parameter; third, the data-splitting inherent in cross-validation can lead to a loss of statistical efficiency, especially in small-sample regimes. Moreover, from a theoretical perspective, cross-validation and grid search methods often lack sharp finite-sample guarantees in the random feature setting, making it difficult to understand their performance in practice. In Figure \ref{fig:comparison}, we empirically compare the performance of cross-validation with an oracle tuning strategy that selects the regularization parameter based on the true test error. The results show that cross-validation can be significantly more expensive than oracle tuning and may still yield worse performance.

\paragraph*{Related literature on adaptive parameter selection.}
Adaptive regularization has been studied extensively in inverse problems, nonparametric regression, and kernel learning. One line of work treats regularized learning as a statistical inverse problem and analyzes spectral regularization methods, including Tikhonov regularization, gradient methods, spectral cut-off, and conjugate-gradient-type procedures \cite{devito2005learning,bauer2007regularization,gerfo2008spectral, blanchard2018optimal,lin2020spectral}. These results describe the bias--variance trade-off and establish optimal rates for appropriately chosen regularization parameters, but the theoretically optimal choice generally depends on unknown smoothness and capacity parameters.

A second line of work uses data-dependent criteria to select the regularization level. Classical examples include hold-out, cross-validation, leave-one-out, generalized cross-validation, and related predictive-risk criteria \cite{golub1979generalized,browne2000cross,gyorfi2002distribution, arlot2010survey,caponnetto2010cross}. Another family of methods relies on residual, discrepancy, or stability information along the regularization path. Discrepancy principles originate in inverse problems \cite{engl1996regularization} and have been extended to statistical inverse problems and kernelized spectral algorithms \cite{blanchard2012discrepancy,celisse2021analyzing}. Related stopping rules for gradient descent, boosting, and inverse problems select an iteration number rather than an explicit ridge parameter \cite{yao2007early,raskutti2014early,blanchard2018early, blanchard2018optimaladapt,wei2019early}.

A particularly relevant approach is the Lepskii or balancing principle, which selects a scale by comparing estimators at different levels of regularization. The original Lepskii principle was introduced for adaptive estimation in Gaussian white noise models \cite{lepskii1991problem}. In kernel learning, \cite{de2010adaptive} developed a balancing principle for KRR, \cite{lu2020balancing} incorporated effective-dimension information for more general regularization schemes, and \cite{blanchard2019lepskii} studied adaptation to source smoothness, eigenvalue decay, and different error norms. The Goldenshluger--Lepski method has also been considered for constrained least-squares estimation over RKHSs \cite{page2018goldenshluger}.

More recent work has adapted these ideas to particular kernel algorithms. For exact KRR, \cite{lin2024adaptive} introduced a uniformly subdivided regularization grid that reduces repeated cross-scale comparisons while retaining optimal learning rates. Distributed versions have been developed in \cite{lin2024lepskii,lin2025adaptive}, and related ideas have been used to select the stopping time of kernel gradient methods \cite{lin2016optimal,lin2020optimal,liu2026beyond}. These results show that the structure of a specific regularization path can be used to design sharper adaptive procedures.

The random feature setting, however, introduces an additional approximation layer. The estimators are determined by finite-dimensional random feature covariance operators, and the analysis must control the sampling error, regularization bias, and random feature approximation error simultaneously. Moreover, a practically useful stopping rule should be computable directly from the random feature representation rather than from the exact kernel matrix. These considerations motivate a neighboring comparison rule tailored specifically to KRR-RF.

\paragraph*{Our Contribution.} 
In this paper, we propose a novel early-stopping rule for KRR-RF that compares only neighboring RF estimators along a uniformly subdivided grid of regularization parameters. Compared with classical Lepskii-type methods, which require all-pairs comparisons, our approach substantially reduces the computational burden. Moreover, thanks to the random feature approximation, the stopping quantity can be computed directly from empirical prediction differences and coefficient norms, making the proposed procedure fully implementable in practice. In contrast to the exact KRR setting, our method exploits the structure of random features to achieve a more efficient and adaptive regularization strategy.

On the theoretical side, we first establish a high-probability comparison bound for two neighboring KRR-RF estimators. We then show that, under standard source and capacity assumptions together with suitable grid and random feature budget conditions, the estimator selected by the proposed rule attains the oracle polynomial learning rate up to logarithmic factors. The regularization rule itself does not require prior knowledge of the source and capacity exponents, and the result covers both the well-specified regime and part of the misspecified regime. The proof combines a fixed-scale KRR-RF error bound with a pathwise comparison argument and a concentration bound for the empirical random feature effective dimension. Finally, simulation and real-data experiments compare the proposed method with cross-validation and Lepskii-type parameter selection. The numerical results illustrate that neighboring comparisons can provide prediction performance comparable to the benchmark methods while reducing the number of discrepancy comparisons along the regularization path.

\section{Preliminaries}\label{sec:pre}

We begin by introducing the necessary notations and background on kernel ridge regression and its variant with random features. 

\subsection{Reproducing Kernel and Kernel Ridge Regression}

Consider a compact input space $\mathcal{X}\subset \mathbb{R}^d$ and a continuous, symmetric, positive-definite kernel function $K: \mathcal{X}\times \mathcal{X} \rightarrow \mathbb{R}$. The kernel $K$ induces a reproducing kernel Hilbert space (RKHS) $\mathcal{H}_K$ of functions on $\mathcal{X}$, equipped with the inner product $\langle \cdot, \cdot \rangle_{\mathcal{H}_K}$ and the associated norm $\|\cdot\|_{\mathcal{H}_K}$. The RKHS has the reproducing property: for any $f \in \mathcal{H}_K$ and $\bfx \in \mathcal{X}$, we have $f(\bfx) = \langle f, K(\cdot, \bfx) \rangle_{\mathcal{H}_K}$. For a standard supervised learning setting, we are given a sample $D=\{(\bfx_i,y_i)\}_{i=1}^{|D|}$ drawn independently from an unknown distribution $\rho$ on $\mathcal{X}\times \mathbb{R}$, where $|D|$ denotes the sample size. The goal is to learn a function $f:\mathcal{X}\rightarrow \mathbb{R}$ that can predict the output $y$ given the input $\bfx$. 

Kernel ridge regression (KRR), which is also known as the Tikhonov regularization in RKHS, is a widely used method for this problem \cite{wahba1990spline,murphy2012machine}. The KRR estimator is defined as the minimizer of the regularized empirical risk:
\begin{align}\label{krr_estimator}
f_{D,\lambda}=\argmin_{f\in \mathcal{H}_K}\left\{\frac{1}{|D|}\sum_{i=1}^{|D|}(f(\bfx_i)-y_i)^2+\lambda \|f\|_{\mathcal{H}_K}^2\right\},
\end{align}
where $\lambda>0$ is the regularization parameter that controls the bias-variance trade-off. Thanks to the celebrated representer theorem \cite{scholkopf2001generalized,wahba2019representer}, the KRR estimator can be expressed in closed form as 
\begin{align}\label{krr_solution}
f_{D,\lambda}(\cdot)=\sum_{i=1}^{|D|}\alpha_i K(\cdot, \bfx_i),
\end{align}
where $\mathbf{\alpha}=(\alpha_1,\ldots,\alpha_{|D|})^T$ is the coefficient vector given by $\mathbf{\alpha}=(\mathbf{K}+\lambda |D| I)^{-1}\mathbf{y}$, with $\mathbf{K}$ being the kernel matrix defined by $\mathbf{K}_{ij}=K(\bfx_i, \bfx_j)$ and $\mathbf{y}=(y_1,\ldots,y_{|D|})^T$.

Despite its simple implementation and strong theoretical properties \cite{smale2007learning,caponnetto2007optimal}, KRR suffers from computational challenges when the sample size $|D|$ is large, as it requires inverting a $|D|\times |D|$ kernel matrix. The computational cost and memory requirements of KRR scale as $O(|D|^3)$ and $O(|D|^2)$, respectively, which can be prohibitive for large datasets. To address this issue, scalable variants of KRR have been developed, including distributed learning \cite{zhang2015divide,lin2017distributed,lin2020distributed,wang2024communication}, Nystr{\"o}m sub-sampling \cite{williams2001using,rudi2015less}, random features \cite{rahimi2007random,rahimi2008weighted,rudin2017Fourier,rudi2017generalization}, stochastic gradient methods \cite{lin2016optimal,lin2020optimal}. 

\subsection{Kernel Ridge Regression with Random Features}

Among these scalable approaches, random features (RF) has gained significant attention due to its simplicity and effectiveness in approximating the exact KRR solution \cite{rahimi2007random,rudin2017Fourier,rudi2017generalization}. Assuming the kernel $K$ has an integral representation
\begin{align}\label{in_re}
K(\bfx, \bfx^{\prime})=\int_{\Omega}\phi(\bfx,\bfw)\phi(\bfx^{\prime},\bfw)d\pi(\bfw),
\end{align}
where $\phi:\mathcal{X}\times \Omega \rightarrow \mathbb{R}$ is a feature map and $\pi$ is a probability measure on $\Omega$. The random feature method constructs an explicit feature map by sampling $M$ random features $\{\bfw_j\}_{j=1}^M$ independently from $\pi$ and defining the random feature map $\phi_M(\bfx)=\frac{1}{\sqrt{M}}(\phi(\bfx,\bfw_1),\ldots,\phi(\bfx,\bfw_M))^T$. The associated random feature kernel is given by $K_M(\bfx, \bfx^{\prime})=\Phi_M(\bfx)^T\Phi_M(\bfx^{\prime})$, which approximates the original kernel $K$.

Substituting the random feature kernel into the KRR formulation, we obtain the KRR with random features (KRR-RF) estimator defined as $f_{M,D,\lambda}(\cdot)=\bfu_{M,D,\lambda}^T \bfphi_{M}(\cdot)$, where $\bfu_{M,D,\lambda}$ is the solution to the following regularized least squares problem:
\begin{align}\label{krr_rf_estimator}
\bfu_{M,D,\lambda}= \argmin_{\bfu \in \mathbb{R}^M}\frac{1}{|D|} \sum_{i=1}^{|D|}\big(y_i-\bfu^T\bfphi_{M}(\bfx_i)\big)^2+\lambda \bfu^T\bfu.
\end{align}
It is easy to derive the closed-form solution $\bfu_{M,D,\lambda}= (\Phi_{M,D}\Phi_{M,D}^T+\lambda I)^{-1}\Phi_{M,D}\mathbf{y}_{D}$, where $\Phi_{M,D}=\frac{1}{\sqrt{|D|}}\big(\bfphi_{M}(\bfx_1),
\ldots,\bfphi_{M}(\bfx_{|D|}) \big)$ and $\mathbf{y}_{D}=\frac{1}{\sqrt{|D|}}(y_1,\ldots,y_{|D|})^T$. Compared with exact KRR, KRR-RF significantly reduces the computational cost to $O(|D|M^2+M^3)$ and memory requirement to $O(|D|M+M^2)$, making it feasible for large-scale learning. 

In statistical learning theory, it has been shown that KRR-RF can achieve minimax optimal rates of convergence under mild assumptions, provided that the regularization parameter $\lambda$ is properly tuned and the number of random features $M$ is sufficiently large \cite{rudi2017generalization,li2021towards}. However, the optimal choice of $\lambda$ typically depends on unknown smoothness parameters of the target function and capacity parameters of the RKHS, which motivates the need for adaptive regularization strategies that can select $\lambda$ in a data-driven manner without prior knowledge of these parameters.

\subsection{Optimal Learning Rates for KRR-RF with Oracle Tuning}

In this section, we briefly review the optimal learning rates for KRR-RF when the regularization parameter $\lambda$ is chosen at the theoretically optimal scale. These results serve as the benchmark for evaluating the performance of our proposed adaptive procedure. The goal of KRR-RF is to learn a function $f_{M,D,\lambda}$ that approximates the regression function $f_{\rho}(\bfx)=\mathbb{E}[y|\bfx]$ based on the observed data $D$ and the random features. The performance of the estimator is typically measured by the excess risk 
\begin{align}\label{excess_risk}
\mathcal{E}(f_{M,D,\lambda})-\mathcal{E}(f_{\rho})=\|f_{M,D,\lambda}-f_{\rho}\|_{\rho}^2,
\end{align}
where $\mathcal{E}(f)=\mathbb{E}[(f(\bfx)-y)^2]$ is the expected risk and $\|f\|_{\rho}=\sqrt{\int_{\cal X} f(\bfx)^2 d\rho_{\cal X}(\bfx)}$ is the $L_{\cal X}^2$-norm with respect to the marginal distribution of $\bfx$. To derive the sharp bounds for the excess risk, we need to impose some standard notations and assumptions.

\begin{assumption}\label{ass1}
    Assume the kernel $K$ admits the integral representation in \eqref{in_re} with $\phi$ being bounded and continuous in both variables. Specifically, there exists a constant $\kappa \geq 1$ such that $|\phi (\bfx, \bfw)| \leq \kappa$ for all $\bfx \in \mathcal{X}$ and $\bfw \in \Omega$. The associated RKHS ${\cal H}_K$ is separable.
\end{assumption}

\begin{assumption}\label{ass2}
For any $\bfx \in \mathcal{X}$ and $p\geq 2$, there holds that $\mathbb E[|y|^p|\bfx]\leq \frac{1}{2}p!B^{p-2}\sigma^2.$
\end{assumption}

Assumption \ref{ass1} ensures that the kernel is well-defined and the associated random features are uniformly bounded. Assumption \ref{ass2} is a standard Bernstein-type moment condition on the response distribution, which allows for sub-Gaussian tails and is commonly used in the analysis of kernel methods \cite{caponnetto2007optimal,rudi2017generalization}. To characterize the smoothness of the target function and the capacity of the RKHS, we introduce the integral operators
\begin{align}\label{l_k}
  L_Kf=\int_{\cal X}K(\bfx,\cdot)f(\bfx)d\rho_{\cal X}.
\end{align}
The operator $L_K$ is a positive, self-adjoint, compact operator on $L^2_{\rho_{\cal X}}$, and its spectral properties play a crucial role in determining the learning rates of KRR-RF. We denote by $\{\mu_j\}_{j=1}^{\infty}$ the eigenvalues of $L_K$ in non-increasing order, and define the effective dimension as
\begin{align}\label{n_lambda}
{\cal N}(\lambda)=\text{Tr}((L_K+\lambda I)^{-1}L_K)=\sum_{j=1}^{\infty}\frac{\mu_j}{\mu_j+\lambda}.
\end{align}

 \begin{assumption}\label{ass3}
      Suppose there exists constants $R>0, r\in(0,1]$ and $h_{\rho} \in L^2_{\rho_{\cal X}}$ such that$f_{\rho}=L_K^r h_{\rho}$, where $\|h_{\rho}\|_{\rho}\leq R$ and $L_K^r$ denotes the $r$-th power of the integral operator $L_K$.
\end{assumption}

\begin{assumption}\label{ass4}
          For any $\lambda>0$, there exists constants $Q>0$ and $\gamma \in [0,1]$ such that ${\cal N}({\lambda}) \leq Q^2\lambda^{-\gamma}$.
          \end{assumption}
\begin{assumption}\label{ass5}
  For any $\lambda>0$, define the maximum dimension of random features as
$ \mathcal{N}_{\infty}(\lambda)=\sup _{\bfw \in \Omega}\left\|(L_K+\lambda I)^{-1 / 2} \phi(\cdot, \bfw)\right\|_{\rho}^2$. Assume there exists constants $\alpha \in[0,1]$ and $F>0$, such that $\mathcal{N}_{\infty}(\lambda) \leq F \lambda^{-\alpha}$. 
\end{assumption}

Assumptions \ref{ass3} and \ref{ass4} are standard source and capacity conditions in the analysis of kernel methods \cite{caponnetto2007optimal,rudi2017generalization,wang2024optimal}. The source condition in Assumption \ref{ass3} characterizes the smoothness of the target function $f_{\rho}$ in terms of the integral operator $L_K$. The parameter $r$ quantifies the degree of smoothness, with larger values of $r$ corresponding to smoother functions. The capacity condition in Assumption \ref{ass4} controls the complexity of the RKHS through the growth rate of the effective dimension ${\cal N}(\lambda)$. The parameter $\gamma$ captures the decay rate of the eigenvalues, with smaller values of $\gamma$ indicating faster decay and thus lower complexity. Assumption \ref{ass5} captures the interaction between the random features and the data-generating distribution via the integral operator $L_K$. It is used to derive refined bounds on the number of random features required to achieve optimal rates, see details in \cite{rudi2017generalization}. Under these assumptions, we have the following optimal learning rates for KRR-RF with oracle tuning of the regularization parameter.

\begin{theorem}\label{thm1}
Suppose Assumptions \ref{ass1}-\ref{ass5} hold and $2r+\gamma>1$. Let
$\lambda_{\mathrm{opt}}=|D|^{-\frac{1}{2r+\gamma}}$, for $\delta \in (0,1)$, assume the sample size satisfying that
$
|D|\geq16(\kappa^2\lambda_{\mathrm{opt}}^{-1}+1)\log(8/\delta).
$
If the number of random features $M$ satisfies $$M \geq
\begin{cases}
c_1\lambda_{\mathrm{opt}}^{-\alpha}\log\frac{96\kappa^2|D|}{\delta}, & r\in (0,\frac{1}{2});\\
c_1\lambda_{\mathrm{opt}}^{-(2r-1)(1+\gamma-\alpha)-\alpha}\log\frac{96\kappa^2|D|}{\delta}, & r\in [\frac{1}{2},1],
\end{cases}$$ 
then with probability at least $1-\delta$, there holds that
\[
\|f_{M,D,\lambda_{\mathrm{opt}}}-f_{\rho}\|_{\rho}
\leq C_1 |D|^{-\frac{r}{2r+\gamma}}\log \frac{16}{\delta},
\]
where $c_1, C_1$ are some positive constants independent of $|D|$, $\lambda_{\mathrm{opt}}$, $M$, and $\delta$.
\end{theorem}

Theorem \ref{thm1} states that if the regularization parameter $\lambda$ is chosen at the optimal scale $\lambda_{\mathrm{opt}}=|D|^{-\frac{1}{2r+\gamma}}$ and the number of random features $M$ is sufficiently large, then the KRR-RF estimator achieves the minimax optimal learning rate of $|D|^{-\frac{r}{2r+\gamma}}$ up to logarithmic factors. Different from the results in \cite{rudi2017generalization} and \cite{li2021towards}, the optimality in Theorem \ref{thm1} allows for $r\in (0,1)$, which covers part of the misspecified case where the regression function does not belong to the RKHS. However, it is worth noting that the optimal choice of $\lambda$ depends on the unknown parameters $r$ and $\gamma$, which motivates the need for adaptive regularization strategies that can select $\lambda$ in a data-driven manner without prior knowledge of these parameters.

\section{Method}\label{sec:method}

In this Section, we present our main method for adaptive regularization in KRR-RF. We first state a key comparison estimate that bounds the difference between two KRR-RF estimators with different regularization parameters. This comparison estimate is crucial for the analysis of adaptive procedures based on comparisons of estimators. We then
introduce our neighboring early-stopping rule, which is the main contribution of this paper. The neighboring early-stopping rule is designed to be computationally efficient and fully implementable in practice, while still achieving optimal adaptive guarantees. 

\subsection{Motivation and Key Comparison Estimate}
In the literature of random feature methods \cite{rahimi2007random,rahimi2008weighted,rudi2017generalization,li2021towards}, the most common approach for selecting the regularization parameter is cross-validation or grid search. The basic idea is to evaluate the performance of the KRR-RF estimator on a validation set for different values of $\lambda$ and select the one that minimizes the validation error. However, this approach can be computationally expensive, as it requires refitting the model multiple times for different $\lambda$ values. Moreover, it may not guarantee optimal performance if the search grid is not sufficiently fine or does not cover the optimal parameter.

To address these issues, we seek a more efficient and adaptive strategy for selecting $\lambda$ that does not require data splitting or exhaustive grid search. Inspired by the principle in \cite{lepskii1991problem} and \cite{lin2024adaptive}, which selects the regularization parameter by comparing estimators across different scales, we aim to design a comparison-based adaptive rule for KRR-RF. The key idea is to compare the estimators corresponding to different regularization parameters and select the one that balances the bias, variance and random feature approximation error. Before introducing our methods, we present a key proposition that provides a comparison estimate between two KRR-RF estimators with different regularization parameters. This comparison estimate is crucial for the analysis of the proposed neighboring early-stopping rule. 

We first introduce some notations, let $C_{M,D}=\Phi_{M,D}\Phi_{M,D}^T\in\mathbb R^{M\times M}$ be the empirical covariance operator associated with the random features, and define the empirical random feature effective dimension as 
\begin{align}\label{N_d}
{\cal N}_{M,D}(\lambda)
=\operatorname{Tr}\!\left[C_{M,D}(C_{M,D}+\lambda I_M)^{-1}\right], \quad \text{for any} \ \lambda>0.
\end{align}
Equivalently, if
$\mathbf K_{M,D}=(K_M(\bfx_i,\bfx_j))_{i,j=1}^{|D|}$ is the random feature Gram matrix, the shared nonzero eigenvalues of $C_{M,D}$ and $\mathbf K_{M,D}/|D|$ give
\[
{\cal N}_{M,D}(\lambda)
=\operatorname{Tr}\!\left[\mathbf K_{M,D}(\mathbf K_{M,D}+\lambda|D|I)^{-1}\right].
\]
The first representation in \eqref{N_d} is used computationally and does not form either an exact-kernel or random feature $|D|\times|D|$ Gram matrix.
We also define the following two quantities that will appear in our proposition and stopping thresholds:
\begin{align}\label{W_d}
{\cal W}_{M,D,\lambda}=\frac{1}{\sqrt{\lambda}|D|}+\left(1+\frac{1}{\sqrt{\lambda|D|}}\right)\sqrt{\frac{\max\{{\cal N}_{M,D}(\lambda),1\}}{|D|}}, 
\end{align}
and for any $\delta \in (0,1)$,
\begin{align}\label{U_d}
{\cal U}_{D,\lambda,\delta}=\frac{2(\kappa^2\lambda^{-1}+1)\log (8/\delta)}{|D|}+\sqrt{\frac{2\kappa^2\log (8/\delta)}{\lambda|D|}}.
\end{align}
\begin{proposition}\label{arf_prop}
Suppose Assumptions \ref{ass1}--\ref{ass5} hold. Let $0<\widetilde\lambda\leq\lambda\leq1$ satisfy
\begin{align}\label{arf_cond1}
\tilde{\lambda} \geq 4\kappa^2/|D| \ \text{and} \ \max\{{\cal U}_{D,\lambda,\delta},{\cal U}_{D,\tilde{\lambda},\delta} \}\leq 1/2,
\end{align} 
If the number of random features $M$ satisfies
\begin{align}\label{arf_cond2}
M\gtrsim 
\begin{cases}
c_{\mathrm{RF}}\widetilde\lambda^{-\alpha}\log\frac{384\kappa^3|D|}{\delta}, & 0<r<\frac12,\\
c_{\mathrm{RF}} \widetilde\lambda^{-[(2r-1)(1+\gamma-\alpha)+\alpha]}\log\frac{384\kappa^3|D|}{\delta}, & \frac12\leq r\leq1,
\end{cases}
\end{align}
then with probability at least $1-\delta$, there holds that
\begin{align}\label{arf_bound}
\|(C_{M,D}+\lambda I)^{1/2}({\bfu}_{M,D,\lambda}-{\bfu}_{M,D,\widetilde\lambda})\|_2
\leq C_2\frac{\lambda-\widetilde\lambda}{\widetilde\lambda}
\left({\cal W}_{M,D,\lambda}+\widetilde\lambda^r\right)\log^2\frac{16}{\delta},
\end{align}
where $c_{\mathrm{RF}}, C_2$ are some positive constants independent of $M$, $\lambda$, $\tilde{\lambda}$, $|D|$ and $\delta$. 
\end{proposition}

Proposition \ref{arf_prop} provides a high-probability bound on the difference between two KRR-RF estimators with different regularization parameters $\lambda$ and $\tilde{\lambda}$. The bound depends on the difference between the regularization parameters, the empirical RF effective dimension, and the smoothness of the target function. This comparison estimate is crucial for analyzing the performance of adaptive procedures that select $\lambda$ based on comparisons of estimators. In particular, it allows us to control the bias-variance trade-off and the random feature approximation error when comparing different estimators, which is essential for establishing optimal adaptive rates. The conditions in \eqref{arf_cond1} and \eqref{arf_cond2} ensure that the regularization parameters are not too small and that the number of random features is sufficiently large to guarantee the stability of the estimators and the validity of the comparison bound.

\subsection{Adaptive Early-Stopping Rule for KRR-RF}\label{sec:adaptive}

In this section, we propose a neighboring early-stopping rule for selecting the regularization parameter in KRR-RF. In Proposition \ref{arf_prop}, a key quantity is $\frac{\lambda-\tilde{\lambda}}{\tilde{\lambda}}$, which is the relative difference between the two regularization parameters. This suggests that when comparing two estimators, it is more efficient to compare those corresponding to neighboring and subdivided regularization parameters, as this can lead to a tighter bound and a more efficient adaptive procedure. Specifically, let $\lambda_k=\frac{1}{hk}$ for some $h>0$ and $k \in \mathbb{N}$, then we have 
$$
\frac{|\lambda_{k+1}-\lambda_k|}{\lambda_{k+1}}=\frac{1}{k}=h\lambda_k,
$$
which is proportional to $\lambda_k$. We set $\lambda=\lambda_{k-1}$ and $\tilde{\lambda}=\lambda_{k}$ in Proposition \ref{arf_prop}, then the bound in \eqref{arf_bound} can be rewritten as
\begin{align}\label{arf_bound_es}
\|(C_{M,D}+\lambda_{k-1} I)^{1/2}({\bfu}_{M,D,\lambda_{k-1}}-{\bfu}_{M,D,\lambda_{k}})\| \leq C_2 h\lambda_{k-1} \left({\cal W}_{M,D,\lambda_{k-1}}+\lambda_k^{r}\right)\log^2 \frac{16}{\delta}.
\end{align}
Now we are ready to introduce our neighboring early-stopping rule for KRR-RF. For any $\delta>0$, we set $K_{\max}=\left\lfloor\frac{|D|}{4\kappa^2h}\right\rfloor$ and $ \delta_{D}=\frac{\delta}{5K_{\max}}$, and assume that $K_{\max} \geq 2$. Denote by $K_{ES}$ the largest index such that $\lambda_k$ satisfies the conditions in \eqref{arf_cond1}, i.e.,
\begin{align}\label{es_max}
K_{ES}=\max\left\{k\in \{1,\ldots,K_{\max}\}:{\cal U}_{D,\lambda_k,\delta_{D}}\leq \frac{1}{2}  \right\}.
\end{align}
and we define the candidate set of regularization parameters as
\begin{align}\label{es_set}
\Lambda_{ES}=\left\{\lambda_k=\frac{1}{hk}:k=K_{ES},\ldots,1\right\}.
\end{align} 
We then select the regularization parameter $\lambda_{ES}$ by comparing neighboring estimators along the grid $\Lambda_{ES}$. Let $C_{ES}=2hC_2$, we define
\begin{equation}
\begin{aligned}\label{es_estimator}
{\cal I}=\Big\{k\in\{2,\ldots,K_{ES}\}:\;&\|(C_{M,D}+\lambda_{k-1}I)^{1/2}(\bfu_{M,D,\lambda_{k-1}}-\bfu_{M,D,\lambda_k})\|_2\\
&\geq C_{ES}\lambda_{k-1}{\cal W}_{M,D,\lambda_{k-1}}\log^2\frac{16}{\delta_{D}}\Big\}
\end{aligned}
\end{equation}
and
\begin{equation}
\begin{aligned}\label{es_estimator_1}
\widehat k=\begin{cases}
\max {\cal I},&{\cal I}\neq\varnothing,\\
1,&{\cal I}=\varnothing,
\end{cases}
\qquad \lambda_{ES}=\lambda_{\widehat k}.
\end{aligned}
\end{equation}
Thus the scan is from $k=K_{ES}$ down to $2$.  If there is no $\lambda_k$ satisfying the above conditions, we set $\lambda_{ES}=\lambda_{1}$. We define $\hat{k}$ as the index of $\lambda_{ES}$ in $\Lambda_{ES}$, i.e., $\lambda_{ES}=\lambda_{\hat{k}}$. Consequently, it implies that 
\begin{align}\label{es_cond}
\|(C_{M,D}+\lambda_{k} I)^{1/2}(\bfu_{M,D,\lambda_{k}}-\bfu_{M,D,\lambda_{k+1}})\|\leq C_{ES}\lambda_{k} {\cal W}_{M,D,\lambda_{k}}\log^2 \frac{16}{\delta_{D}},\ \text{for any} \ k=\hat{k},\ldots,K_{ES}-1.
\end{align}
The next proposition shows that the stopping quantity can be evaluated directly from empirical prediction differences and coefficient norms, so the proposed rule is fully implementable in the random feature coordinates.

\begin{proposition}\label{explicit_prop}
For two KRR-RF estimators $f_{M,D,\lambda}(\bfx)=\boldsymbol{u}_{M,D,\lambda}^T\phi_M(\bfx)$ and $f_{M,D,\tilde{\lambda}}(\bfx)=\boldsymbol{u}_{M,D,\tilde{\lambda}}^T\phi_M(\bfx)$ with $\lambda, \tilde{\lambda} > 0$, there holds the following explicit equality:
$$\|(C_{M,D} + \lambda I)^{1/2} (\boldsymbol{u}_{M,D,\lambda} - \boldsymbol{u}_{M,D,\tilde{\lambda}})\|_2^2 =  \|f_{M,D,\lambda} - f_{M,D,\tilde{\lambda}}\|_D^2 + \lambda \|\boldsymbol{u}_{M,D,\lambda} - \boldsymbol{u}_{M,D,\tilde{\lambda}}\|_2^2 ,$$
where we denote the empirical norm of $f$ as $\|f\|_D^2 = \frac{1}{|D|} \sum_{i=1}^{|D|} f(\bfx_i)^2$.
\end{proposition}

We summarize the complete procedure for the proposed neighboring early-stopping rule in Algorithm \ref{alg1}.
\begin{algorithm}
\caption{Neighboring Early-Stopping Rule for KRR-RF (NESR-KRR-RF)}
\label{alg1}
\begin{algorithmic}[1]
\Require Sample $D$, number of random features $M$, confidence level
$\delta\in(0,1)$, step size $h>0$, kernel bound $\kappa$, and
threshold constant $C_{ES}$.

\State Set
\[
K_{\max}
=
\left\lfloor
\frac{|D|}{4\kappa^2h}
\right\rfloor,
\qquad
\delta_D
=
\frac{\delta}{5K_{\max}},
\qquad
\lambda_k=\frac{1}{hk}.
\]

\State Determine the admissible endpoint
\[
K_{ES}
=
\max
\left\{
k\in\{1,\ldots,K_{\max}\}:
{\cal U}_{D,\lambda_k,\delta_D}\leq\frac12
\right\}.
\]
If the above set is empty, set $K_{ES}=1$.

\State Construct
\[
\Lambda_{ES}
=
\{\lambda_k:k=1,\ldots,K_{ES}\}.
\]

\State Generate random features $\{\bfw_j\}_{j=1}^M$,
construct the random feature map $\bfphi_M$, and compute
\[
C_{M,D}
=
\Phi_{M,D}\Phi_{M,D}^{\top}.
\]

\State Compute the KRR-RF estimator
$\bfu_{M,D,\lambda_{K_{ES}}}$.

\If{$K_{ES}=1$}
    \State \Return
    $f_{M,D,\lambda_1}
    =S_M\bfu_{M,D,\lambda_1}$.
\EndIf

\For{$k=K_{ES},K_{ES}-1,\ldots,2$}
    \State Compute $\bfu_{M,D,\lambda_{k-1}}$ and
    ${\cal N}_{M,D}(\lambda_{k-1})$, and then evaluate
    ${\cal W}_{M,D,\lambda_{k-1}}$ according to \eqref{W_d}.

    \State Compute the neighboring discrepancy
    \[
    \Delta_k
    =
    \left\|
    (C_{M,D}+\lambda_{k-1}I)^{1/2}
    \left(
    \bfu_{M,D,\lambda_{k-1}}
    -
    \bfu_{M,D,\lambda_k}
    \right)
    \right\|_2.
    \]

    \If{
    $\displaystyle
    \Delta_k
    \geq
    C_{ES}\lambda_{k-1}
    {\cal W}_{M,D,\lambda_{k-1}}
    \log^2\frac{16}{\delta_D}
    $
    }
        \State Set $\widehat{k}=k$ and
        $\lambda_{ES}=\lambda_{\widehat{k}}$.
        \State \Return
        $f_{M,D,\lambda_{ES}}
        =S_M\bfu_{M,D,\lambda_{ES}}$.
    \EndIf
\EndFor

\State No threshold crossing occurs. Set
$\widehat{k}=1$ and $\lambda_{ES}=\lambda_1$.
\State \Return
$f_{M,D,\lambda_{ES}}
=S_M\bfu_{M,D,\lambda_{ES}}$.

\end{algorithmic}
\end{algorithm}

\subsection{Comparison with Lepskii-Type Parameter Selection}

The proposed neighboring early-stopping rule is closely related to the
Lepskii principle, which selects a regularization scale by comparing
estimators computed at different levels of regularization. In this subsection,
we first compare NESR with a standard Lepskii-type rule implemented in the
random feature space, and then discuss its relation to the adaptive selection
with uniform subdivision proposed by \cite{lin2024adaptive} for exact KRR.

\paragraph*{Comparison with the standard Lepskii-type rule.}
A standard Lepskii-type procedure considers a geometric grid of regularization
parameters and selects the largest regularization level for which the
corresponding estimator remains sufficiently close to all estimators at finer
scales. For $q\in(0,1)$, define the geometric grid
\[
\lambda_k=q^k, \qquad k=0,1,2,\ldots,
\]
and let
\begin{align}\label{lk_max}
K_{LP}
=
\max\left\{
k\in\mathbb N_0:\ {\cal U}_{D,\lambda_k,\delta_{D}}\le \frac12,
\ \ 0\le k \le \left\lfloor\log_q \frac{4\kappa^2}{|D|}\right\rfloor
\right\}.
\end{align}
The candidate set is then given by
\begin{align}\label{lk_set}
\Lambda_{LP}
=
\{\lambda_k=q^k:\ k=0,1,\ldots,K_{LP}\}.
\end{align}

Based on this grid, the standard Lepskii-type rule selects
\begin{equation}
\begin{aligned}\label{lp_estimator}
\lambda_{LP}
=
\max\Bigg\{
\lambda_k\in \Lambda_{LP}:
\|(C_{M,D}+\lambda_k I)^{1/2}(\bfu_{M,D,\lambda_{k^{\prime}}}-\bfu_{M,D,\lambda_k})\|_2 \\
\leq C_{LP} {\cal W}_{M,D,\lambda_k}\log^2 \frac{16}{\delta_{D}},
\quad
k^{\prime}=k+1,\ldots,K_{LP}
\Bigg\}.
\end{aligned}
\end{equation}
In other words, $\lambda_{LP}$ is chosen as the largest regularization level such that the estimator at scale $\lambda_k$ remains stable when compared with all estimators on the candidate grid.

The main drawback of \eqref{lp_estimator} is that it requires all-pairs comparisons over $\Lambda_{LP}$, which leads to a substantially higher computational cost than the proposed neighboring rule. More importantly, the geometric grid does not fully exploit the local behavior of the regularization path. Indeed, if $\lambda_k=q^k$, then
\[
\frac{|\lambda_{k+1}-\lambda_k|}{\lambda_{k+1}}
=
\frac{1-q}{q}.
\]
Applying Proposition~\ref{arf_prop} with $\lambda=\lambda_k$ and $\tilde{\lambda}=\lambda_{k+1}$ yields
\begin{align}\label{LP_bound}
\|(C_{M,D}+\lambda_k I)^{1/2}({\bfu}_{M,D,\lambda_k}-{\bfu}_{M,D,\lambda_{k+1}})\|_2
\leq
C_2 (1-q)q^{-1}
\left({\cal W}_{M,D,\lambda_k}+\lambda_{k+1}^{r}\right)
\log^2 \frac{16}{\delta_{D}}.
\end{align}
The factor $(1-q)q^{-1}$ is determined solely by the geometric spacing of the grid and does not decrease with $k$. Consequently, the comparison bound in the standard Lepskii rule cannot capture the finer scale variation of neighboring estimators when the regularization path is locally smooth.

By contrast, our neighboring early-stopping rule is constructed on a uniformly subdivided grid, so that adjacent parameters are much closer and the corresponding comparison quantity is more sensitive to local changes of the estimator. This refined construction not only reduces the computational burden by replacing all-pairs comparisons with neighboring comparisons, but also provides a sharper characterization of the regularization path in the random feature setting. In this sense, the proposed rule can be viewed as a localized and computationally more efficient alternative to the standard Lepskii-type method.

\paragraph*{Comparison with the adaptive rule of \cite{lin2024adaptive}.}
The adaptive selection with uniform subdivision (ASUS) developed by
\cite{lin2024adaptive} is more closely related to NESR than the classical
Lepskii rule. For exact KRR, \cite{lin2024adaptive} exploits the special
spectral structure of the KRR estimator to relate the difference between
successive estimators to an empirical complexity quantity. This leads to an
early-stopping-type implementation of the Lepskii principle that avoids the
recurrent pairwise comparisons required by classical LP. In this respect,
both the method of \cite{lin2024adaptive} and NESR use local comparisons
along a suitably subdivided regularization path.

The main distinction lies in the statistical and computational representation
in which the comparison is carried out. The procedure of
\cite{lin2024adaptive} is developed for exact KRR and is formulated through
quantities associated with the exact kernel estimator. In the random feature
setting, replacing the kernel by a finite-dimensional randomized
approximation introduces an additional source of error. Consequently, an
adaptive rule for KRR-RF must simultaneously account for sample fluctuations,
regularization bias, and random feature approximation. To address this issue, Proposition~\ref{arf_prop} establishes a comparison
bound directly between two KRR-RF estimators. The stochastic part of the
bound is measured by ${\cal W}_{M,D,\lambda}$,
which depends on the empirical random feature effective dimension ${\cal N}_{M,D}(\lambda)
    =
    \operatorname{Tr}
    \left[
        C_{M,D}
        (C_{M,D}+\lambda I)^{-1}
    \right].
$
This quantity is computed entirely from the empirical random feature
covariance matrix $C_{M,D}\in\mathbb R^{M\times M}$ and therefore does not
require the exact kernel Gram matrix.
Moreover, Proposition \ref{explicit_prop} shows that the neighboring
discrepancy admits an explicit representation.
Thus, once the two KRR-RF estimators have been computed, their discrepancy can
be evaluated directly from their fitted values and coefficient vectors,
without explicitly computing a matrix square root or forming the exact
kernel Gram matrix.

\begin{remark}[Computational Comparison]
Suppose that the candidate regularization path contains $K$
values. We compare the computational costs of the proposed NESR
method with the classical Lepskii principle (LP), the adaptive
selection with uniform subdivision for exact KRR proposed by
\cite{lin2024adaptive}, and the direct random feature implementation
of LP (LP-RF). We consider a direct implementation in which the
linear system corresponding to each candidate regularization
parameter is solved separately.

For exact KRR, both LP and the method of \cite{lin2024adaptive}
require the $|D|\times |D|$ kernel Gram matrix. For each
regularization parameter, computing the corresponding KRR
estimator by a standard dense linear solver requires
$\mathcal{O}(|D|^3)$ operations. Hence, computing the estimators
along a path containing $K$ candidate values requires
$\mathcal{O}(K|D|^3)$ operations. Once the candidate estimators
have been obtained, a discrepancy between two estimators can be
evaluated using the kernel Gram matrix in
$\mathcal{O}(|D|^2)$ operations. Therefore, the all-pairs
comparisons in LP require $\mathcal{O}(K^2|D|^2)$ additional
operations, whereas the neighboring comparisons in
\cite{lin2024adaptive} require only
$\mathcal{O}(K|D|^2)$ operations.

For KRR-RF, let
$C_{M,D}\in\mathbb{R}^{M\times M}$ be the empirical random feature
covariance matrix. Constructing $C_{M,D}$ requires
$\mathcal{O}(|D|M^2)$ operations. For each regularization
parameter $\lambda$, the coefficient vector $u_{M,D,\lambda}$ is
obtained by solving
$
(C_{M,D}+\lambda I)u_{M,D,\lambda}
=
\Phi_{M,D}y_D,
$
which requires $\mathcal{O}(M^3)$ operations using a standard
dense linear solver. Thus, computing the KRR-RF estimators over
$K$ candidate values requires
$
\mathcal{O}(|D|M^2+KM^3)
$
operations. The main difference between LP-RF and NESR lies in the comparison
stage. For two KRR-RF estimators, let
$
\Delta u
=
u_{M,D,\lambda}
-
u_{M,D,\widetilde{\lambda}}.
$
By Proposition \ref{explicit_prop},
$
\left\|
(C_{M,D}+\lambda I)^{1/2}\Delta u
\right\|_2^2
=
\Delta u^\top C_{M,D}\Delta u
+
\lambda\|\Delta u\|_2^2.
$
Since $C_{M,D}$ has already been constructed, each discrepancy
can be evaluated in $\mathcal{O}(M^2)$ operations. Consequently,
the all-pairs comparisons in LP-RF require
$\mathcal{O}(K^2M^2)$ operations, whereas NESR requires only
$\mathcal{O}(KM^2)$ operations for the neighboring comparisons.
The corresponding overall costs are summarized in
Table~\ref{tab:complexity_comparison}.

\begin{table}[htbp]
    \centering
    \caption{Computational complexity of different adaptive
    regularization methods.}
    \label{tab:complexity_comparison}
    \begin{tabular}{lcc}
        \hline
        Method
        & Main computational cost
        & Memory cost
        \\
        \hline
        LP
        & $\mathcal{O}(K|D|^3+K^2|D|^2)$
        & $\mathcal{O}(|D|^2)$
        \\
        ASUS
        & $\mathcal{O}(K|D|^3+K|D|^2)$
        & $\mathcal{O}(|D|^2)$
        \\
        LP-RF
        & $\mathcal{O}(|D|M^2+KM^3+K^2M^2)$
        & $\mathcal{O}(|D|M+M^2)$
        \\
        NESR-KRR-RF
        & $\mathcal{O}(|D|M^2+KM^3+KM^2)$
        & $\mathcal{O}(|D|M+M^2)$
        \\
        \hline
    \end{tabular}
\end{table}

Here the cost of generating the random features is not included,
since it is common to LP-RF and NESR-KRR-RF. The estimator
computation along the regularization path has the same order for
LP-RF and NESR-KRR-RF. Their computational difference arises
from the comparison stage: LP-RF performs a quadratic number of
cross-scale comparisons, whereas NESR uses only a linear number
of neighboring comparisons. Compared with the exact-KRR method
of \cite{lin2024adaptive}, NESR performs the entire adaptive
selection in the $M$-dimensional random feature space rather than
the $|D|$-dimensional sample space. When $M\ll |D|$, this leads
to a substantially lower computational and memory cost.
\end{remark}

\section{Theory}\label{sec:theory}

In this section, we present the main theoretical result of this paper, which establishes that the KRR-RF estimator with the regularization parameter selected by the neighboring early-stopping rule achieves the optimal adaptive learning rate up to logarithmic factors. 

\begin{theorem}\label{thm2}
Suppose Assumptions \ref{ass1}-\ref{ass5} hold, $h\geq1$, $2r+\gamma>1$, and $K_{ES}\geq2$.  We also assume the candidate path covers the oracle parameter
$$
\lambda_{K_{ES}}\leq\lambda_{\mathrm{opt}}\leq\lambda_1.
$$
If the number of random features $M$ satisfies 
$$
M \geq
\begin{cases}
c_{\mathrm{RF}}\lambda_{K_{ES}}^{-\alpha}\log\frac{\kappa^2|D|K_{\max}}{\delta}, & r\in (0,\frac{1}{2}),\\[4pt]
c_{\mathrm{RF}}\lambda_{K_{ES}}^{-(2r-1)(1+\gamma-\alpha)-\alpha}\log\frac{\kappa^2|D|K_{\max}}{\delta}, & r\in [\frac{1}{2},1],
\end{cases}
$$
then the estimator selected with $\lambda_{ES}=\lambda_{\hat{k}}$ defined in \eqref{es_estimator_1} satisfies, with probability at least $1-\delta$,
\[
\|f_{M,D,\lambda_{ES}}-f_{\rho}\|_{\rho}
\leq C_3 |D|^{-\frac{r}{2r+\gamma}}\log^4 \frac{16}{\delta_{D}},
\]
where $c_{\mathrm{RF}}, C_3$ are some positive constants independent of $|D|$, $\lambda$, $M$, and $\delta$.
\end{theorem}

Theorem \ref{thm2} states that the KRR-RF estimator with the regularization parameter selected by the neighboring early-stopping rule achieves the same convergence rate as the oracle choice, up to logarithmic factors. This result demonstrates that the proposed adaptive procedure can effectively select the regularization parameter without prior knowledge of the smoothness and capacity parameters, while still attaining optimal statistical performance. 

\begin{remark}
The lower bound on the number of random features $M$ in Theorem \ref{thm2} depends on
the unknown parameters $r$, $\alpha$, and $\gamma$. The role of $M$, however,
is different from that of the regularization parameter $\lambda$. While
$\lambda$ determines the bias--variance trade-off and is selected adaptively
by NESR, $M$ controls the accuracy of the random feature approximation. Accordingly, we treat $M$ as a prespecified computational budget rather than
an additional tuning parameter of the proposed procedure. Theorem \ref{thm2} requires
this budget to be sufficiently large for the stated oracle-rate guarantee. In
the numerical experiments, we either fix $M$ or vary it over a prescribed
range to examine the sensitivity of the method to the number of random
features. Adaptive selection of $M$ is not considered in this work. A
parameter-uniform sufficient feature bound is given in the following
corollary.
\end{remark}

The following corollary provides a conservative feature bound that is uniform over the unknown exponents $r$, $\alpha$, and $\gamma$.

\begin{corollary}\label{cor1}
Under the conditions of Theorem~\ref{thm2}, Define
\[
    \beta
    :=
    \alpha + (2r-1)_{+}(1+\gamma-\alpha),
\]
where $(x)_{+}:=\max\{x,0\}$, then we have
$
    0\le \beta\le 2.
$
Hence, a sufficient condition for the number of random features is
\[
    M
    \ge
    \bar c_{\mathrm{RF}}
    \lambda_{K_{\mathrm{ES}}}^{-2}
    \log\left(
        \frac{
        \kappa^2 |D|K_{\max}
        }{\delta}
    \right),
\]
where $\bar c_{\mathrm{RF}}$ is a sufficiently large constant. This bound is
valid for all $r\in(0,1]$ and $\alpha,\gamma\in[0,1]$ covered by
Theorem~\ref{thm2}.
\end{corollary}

\paragraph*{Proof sketch of Theorem \ref{thm2}.}
Let
$
    k^*
    :=
    \max\{k\in\{1,\ldots,K_{ES}\}:\lambda_k\ge \lambda_{opt}\}.
$
Under the grid-coverage condition and the inverse-regularization grid,
$\lambda_{k^*}$ is within a constant factor of the oracle scale,
$
    \lambda_{opt}
    \le
    \lambda_{k^*}
    \le
    2\lambda_{opt}.
$
We first construct a single high-probability event on which the fixed-scale
error bound, the empirical RF effective-dimension estimates, the covariance
stability bounds, and Proposition~1 hold simultaneously for all candidate
scales and all neighboring pairs. This is obtained by applying the
corresponding results with failure probability $\delta_D$ and taking a union
bound over the candidate path. The feature condition imposed at
$\lambda_{K_{ES}}$ is sufficient for all larger grid values. We then distinguish two cases according to the position of the selected scale
relative to $\lambda_{k^*}$.

\emph{Case 1: $\lambda_{ES}\le \lambda_{k^*}$.}
Except for the boundary case $\hat{k}=k^*=1$, the selected index is generated
by a threshold crossing. Combining this crossing inequality with
Proposition~1 shows that the empirical stochastic scale at the selected point
is controlled by the bias scale,
$
    {\cal W}_{M,D,\lambda_{ES}}
    \lesssim
    \lambda_{ES}^{r}.
$
The fixed-scale oracle inequality then gives
$
    \|f_{M,D,\lambda_{ES}}-f_\rho\|_\rho
    \lesssim
    \lambda_{ES}^{r}
$
up to logarithmic factors. Since
$\lambda_{ES}\le\lambda_{k^*}\le2\lambda_{opt}$, this yields the desired
oracle rate.

\emph{Case 2: $\lambda_{ES}>\lambda_{k^*}$.}
In this case, $\hat{k}<k^*$, and the neighboring discrepancies from
$\hat{k}$ to $k^*-1$ are all below their stopping thresholds. This remains
true in the no-crossing case, where the fallback index is $\hat{k}=1$.
By summing the differences between consecutive estimators along the
regularization path, we obtain a bound for
$
    \|f_{M,D,\lambda_{ES}}-f_{M,D,\lambda_{k^*}}\|_\rho
$. The empirical RF
effective-dimension bounds and the capacity condition control this sum by
$
    |D|^{-r/(2r+\gamma)}
$
up to logarithmic factors. Combining this estimate with the fixed-scale bound
at $\lambda_{k^*}$ gives the same rate for the selected estimator.

\section{Simulation}

\subsection{Simulation Setup}

In this section, we present simulation results to evaluate the performance of the proposed neighboring early-stopping rule (NESR) for KRR-RF. Following the simulation setting in \cite{rudi2017generalization}, we consider the periodic spline kernel on the unit interval $\mathcal{X}=[0,1]$. For $q>1$, define
$$
    \Lambda_q(x,x')
    =
    1
    +
    2\sum_{k=1}^{\infty}
    k^{-q}
    \cos\bigl(2\pi k(x-x')\bigr).
$$
The periodic spline kernels satisfy the convolution identity
$$
    \int_0^1
    \Lambda_q(x,z)
    \Lambda_{q'}(x',z)
    \,dz
    =
    \Lambda_{q+q'}(x,x'),
$$
whenever the corresponding series are well defined. This identity provides a convenient way to construct kernels and regression functions with different smoothness levels.
For $r\in(0,1]$ and $\gamma\in(0,1]$, we set
$K(x,x')
    =
    \Lambda_{1/\gamma}(x,x')$. Motivated by the convolution identity above, we use the random feature representation $\phi(x,w)
    =
    \Lambda_{1/(2\gamma)}(x,w)$, where $w\sim U(0,1)$,
so that formally
$    K(x,x')
    =
    \int_0^1
    \phi(x,w)\phi(x',w)
    \,dw.$
The samples are generated according to
$$
    y
    =
    f_\rho(x)+\epsilon,
$$
where
$$
    f_\rho(x)
    =
    0.1\,
    \Lambda_{r/\gamma+1/2}(x,0),
    \qquad
    x\sim U(0,1),
    \qquad
    \epsilon\sim N(0,0.1^2).
$$
For this construction, the random feature complexity exponent is given by $\alpha=\gamma$ \cite{rudi2017generalization}.
We consider four parameter configurations,
$$
    (r,\gamma)
    \in
    \left\{
    (0.8,0.2),
    (0.6,0.2),
    (0.5,0.45),
    (0.4,0.1)
    \right\},
$$
which represent different combinations of target smoothness and kernel capacity. The first three settings satisfy $2r+\gamma>1$ and therefore fall within the regime covered by Theorem~\ref{thm2}. For the last setting, $(r,\gamma)=(0.4,0.1)$, we have $2r+\gamma<1$. This case is included as an additional numerical experiment outside the theoretical regime of Theorem~\ref{thm2}.

To assess the performance of NESR, we compare it with three benchmark
parameter-selection strategies for KRR-RF: the standard Lepskii-type rule
(LP), five-fold cross-validation (CV), and an oracle selector. In the synthetic
experiments, the regularization parameter is parameterized as $\lambda
    =
    C|D|^{-1/(2r+\gamma)},
$
where $r$ and $\gamma$ are known from the data-generating mechanism. The
oracle selector chooses $C$ by minimizing the test error over a logarithmically
spaced grid of 100 values between $10^{-2}$ and $10^{1}$. For cross-validation, we use standard five-fold CV over $\mathcal{C}_{\mathrm{cv}}
    =
    \{0.01,\,0.05,\,0.1,\,0.5,\,1,\,2,\,4,\,8,\,10\},
$
and select the value of $C$ that minimizes the average validation error across
the five folds. The final KRR-RF estimator is then refitted on the full 
training sample using the selected regularization parameter. We note that the
parameterization $\lambda=C|D|^{-1/(2r+\gamma)}$ uses the known values of
$r$ and $\gamma$ in the simulation design and is therefore used here only for
controlled numerical comparison.

The values used in each experiment are reported in Table~\ref{tab:sim_parameter_settings}.
For LP, we use the geometric sequence
$
    \lambda_k=q^k
$
with $q=0.5$.
The range of indices is specified separately for each experiment. For the
prediction experiments, LP is evaluated over the full candidate grid
$
    \Lambda_{LP}^{\mathrm{pred}}
    =
    \{q^k:q^k\ge 10^{-10},\; k=0,1,\ldots\}.
$
For the runtime experiments, we use restricted subgrids of the same
Geometric sequence, as specified below.
The confidence level is denoted by $\delta$, while $C_{ES}$ and $C_{LP}$
are the threshold constants defined in \eqref{es_cond} and \eqref{lp_estimator}. For NESR,
$h$ determines the spacing of the inverse-regularization grid and
$K_{ES}$ specifies the endpoint of the candidate set $\Lambda_{ES}$.
The quantities $|D|$ and $|D'|$ denote the numbers of training and test
samples, respectively. Prediction performance is evaluated on the independent test set using
either mean squared error (MSE) or root mean squared error (RMSE),
depending on the experiment
$$
\text{MSE}=\frac{1}{n_{te}}\sum_{i=1}^{n_{te}}\left(\widehat{f}(\bfx_i)-f_{\rho}(\bfx_i)\right)^2, \quad \text{RMSE}=\sqrt{\frac{1}{n_{te}}\sum_{i=1}^{n_{te}}\left(\widehat{f}(\bfx_i)-f_{\rho}(\bfx_i)\right)^2}
$$
All simulation results are averaged over
50 independent repetitions.

\begin{longtable}{@{}lccccccccc@{}}
    \caption{Parameter settings for NESR and LP.}
    \label{tab:sim_parameter_settings} \\ 
    
    % First Head
    \toprule

    & $q$ & $\delta$ & $C_{LP}$ & $C_{ES}$ & $h$ & $K_{ES}$ & $M$ & $|D|$ & $|D'|$ \\
    \midrule
    \endfirsthead
    
    % Head
    \caption[]{Parameter settings for NESR and LP. (continued)} \\
    \toprule

    & $q$ & $\delta$ & $C_{LP}$ & $C_{ES}$ & $h$ & $K_{ES}$ & $M$ & $|D|$ & $|D'|$ \\
    \midrule
    \endhead
    
    \bottomrule
    \endfoot
    
    \bottomrule
    \endlastfoot

    Fig.~\ref{sim:fig1}:(a) & 0.5 & 0.01 & 0.1    & 0.005  & 50  & 20  & 100 & 5000 & 1000 \\
    Fig.~\ref{sim:fig1}:(b) & 0.5 & 0.01 & 0.05   & 0.003  & 50  & 20  & 100 & 5000 & 1000 \\
    Fig.~\ref{sim:fig1}:(c) & 0.5 & 0.01 & 0.0007 & 0.01   & 50  & 20  & 100 & 5000 & 1000 \\
    Fig.~\ref{sim:fig1}:(d) & 0.5 & 0.01 & 0.002  & 0.002  & 200 & 100 & 100 & 5000 & 1000 \\
    \addlinespace
    Fig.~\ref{sim:fig2}:(a) & 0.5 & 0.01 & 0.1    & 0.005  & 50  & 20  & 100 & -    & 1000 \\
    Fig.~\ref{sim:fig2}:(b) & 0.5 & 0.01 & 0.05   & 0.003  & 50  & 20  & 100 & -    & 1000 \\
    Fig.~\ref{sim:fig2}:(c) & 0.5 & 0.01 & 0.0007 & 0.01   & 50  & 20  & 100 & -    & 1000 \\
    Fig.~\ref{sim:fig2}:(d) & 0.5 & 0.01 & 0.002  & 0.002  & 200 & 100 & 100 & -    & 1000 \\
    \addlinespace
    Fig.~\ref{sim:fig3}:(a) & 0.5 & 0.01 & 0.0001 & 0.0025 & 50  & 20  & 100 & -    & 1000 \\
    Fig.~\ref{sim:fig3}:(b) & 0.5 & 0.01 & 0.0001 & 0.002  & 50  & 20  & 100 & -    & 1000 \\
    Fig.~\ref{sim:fig3}:(c) & 0.5 & 0.01 & 0.0002 & 0.006  & 50  & 20  & 100 & -    & 1000 \\
    Fig.~\ref{sim:fig3}:(d) & 0.5 & 0.01 & 0.0001 & 0.0005 & 200 & 100 & 100 & -    & 1000 \\
    \addlinespace
    Fig.~\ref{sim:fig4}:(a) & 0.5 & 0.01 & 0.1    & 0.005  & 50  & 20  & -   & 5000 & 1000 \\
    Fig.~\ref{sim:fig4}:(b) & 0.5 & 0.01 & 0.05   & 0.003  & 50  & 20  & -   & 5000 & 1000 \\
    Fig.~\ref{sim:fig4}:(c) & 0.5 & 0.01 & 0.0007 & 0.01   & 50  & 20  & -   & 5000 & 1000 \\
    Fig.~\ref{sim:fig4}:(d) & 0.5 & 0.01 & 0.002  & 0.002  & 200 & 100 & -   & 5000 & 1000 \\
    \addlinespace
    Fig.~\ref{sim:fig5}:(a) & 0.5 & 0.01 & 0.0001 & 0.0025 & 50  & 20  & -   & 5000 & 1000 \\
    Fig.~\ref{sim:fig5}:(b) & 0.5 & 0.01 & 0.0001 & 0.002  & 50  & 20  & -   & 5000 & 1000 \\
    Fig.~\ref{sim:fig5}:(c) & 0.5 & 0.01 & 0.0002 & 0.006  & 50  & 20  & -   & 5000 & 1000 \\
    Fig.~\ref{sim:fig5}:(d) & 0.5 & 0.01 & 0.0001 & 0.0005 & 200 & 100 & -   & 5000 & 1000 \\
    \addlinespace
    Fig.~\ref{sim:fig6}:(a) & 0.5 & 0.01 & -      & 0.0045 & -   & 20  & -   & 5000 & 1000 \\
    Fig.~\ref{sim:fig6}:(b) & 0.5 & 0.01 & -      & 0.003  & -   & 20  & -   & 5000 & 1000 \\
    Fig.~\ref{sim:fig6}:(c) & 0.5 & 0.01 & -      & 0.01   & -   & 20  & -   & 5000 & 1000 \\
    Fig.~\ref{sim:fig6}:(d) & 0.5 & 0.01 & -      & 0.002  & -   & 100 & -   & 5000 & 1000 \\
\end{longtable}

\subsection{Curve Fitting Results}

\begin{figure}[htbp!]
    \centering
    \begin{subfigure}[b]{0.45\linewidth}
        \centering
        \includegraphics[width=\linewidth]{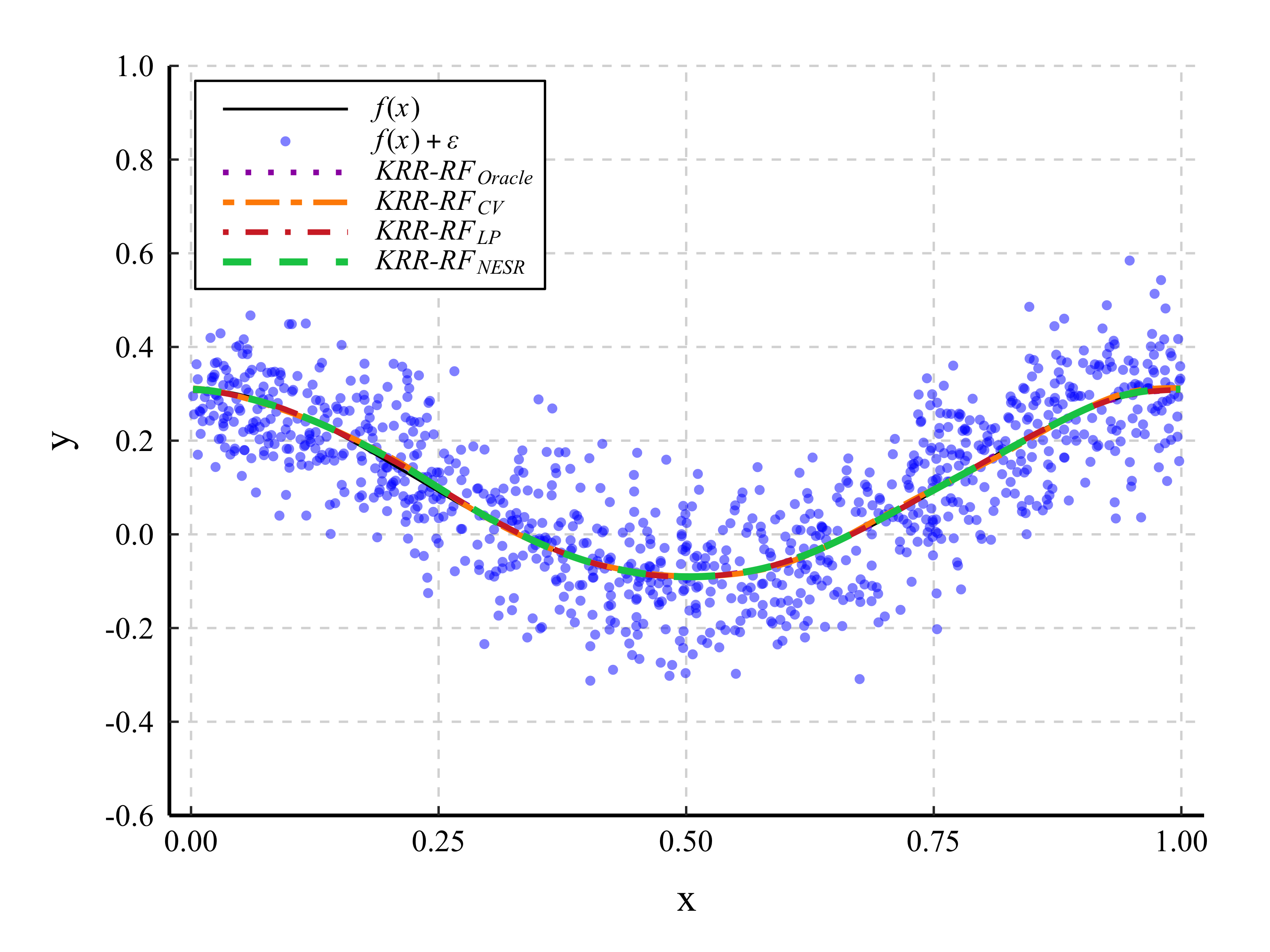}
        \caption{$r=0.8, \gamma=0.2$}
        \label{sim:fig1_a}
    \end{subfigure}
    \hfill
    \begin{subfigure}[b]{0.45\linewidth}
        \centering
        \includegraphics[width=\linewidth]{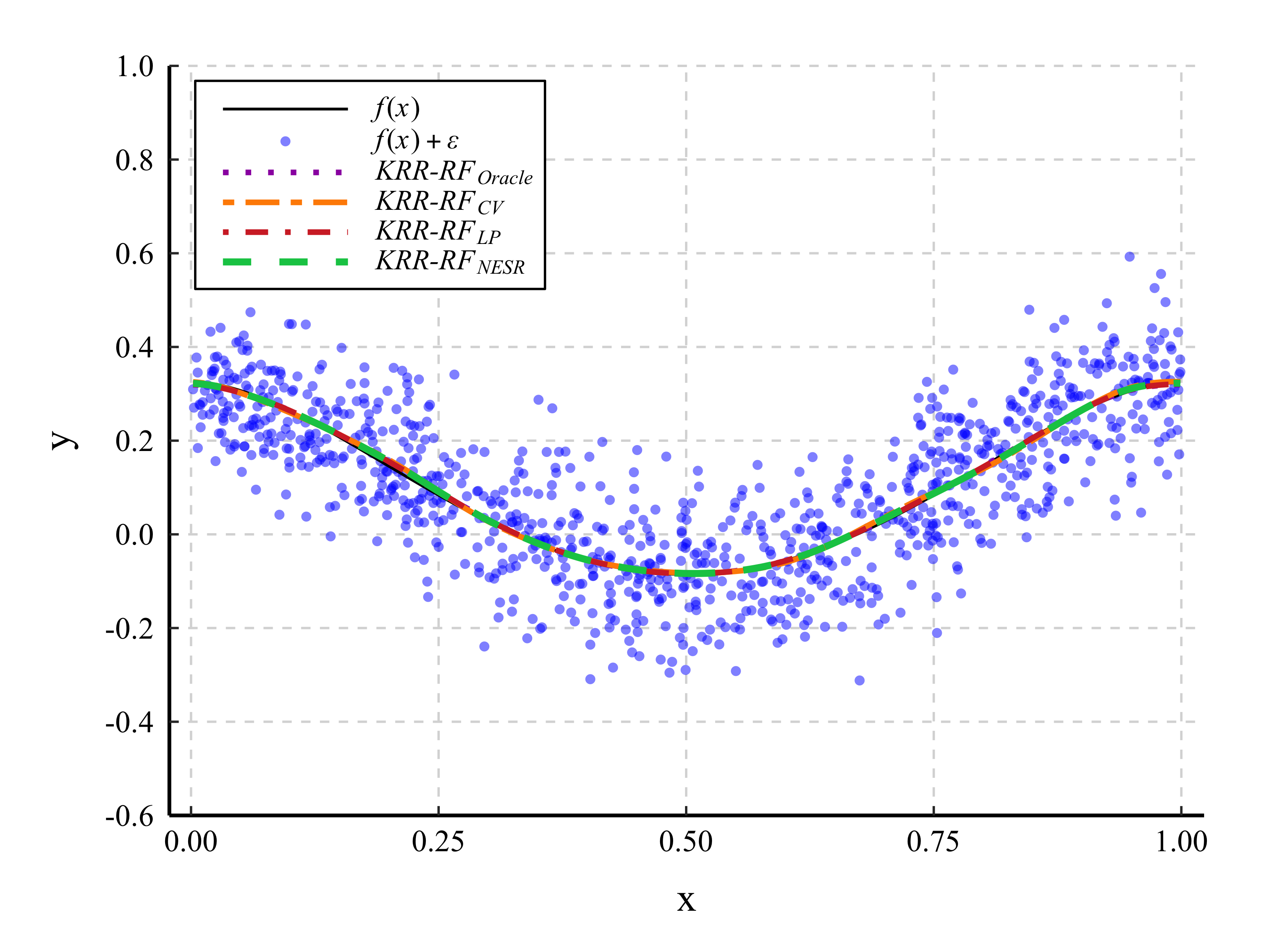}
        \caption{$r=0.6, \gamma=0.2$}
        \label{sim:fig1_b}
    \end{subfigure}
    
    \vspace{0.4cm}
    
    \begin{subfigure}[b]{0.45\linewidth}
        \centering
        \includegraphics[width=\linewidth]{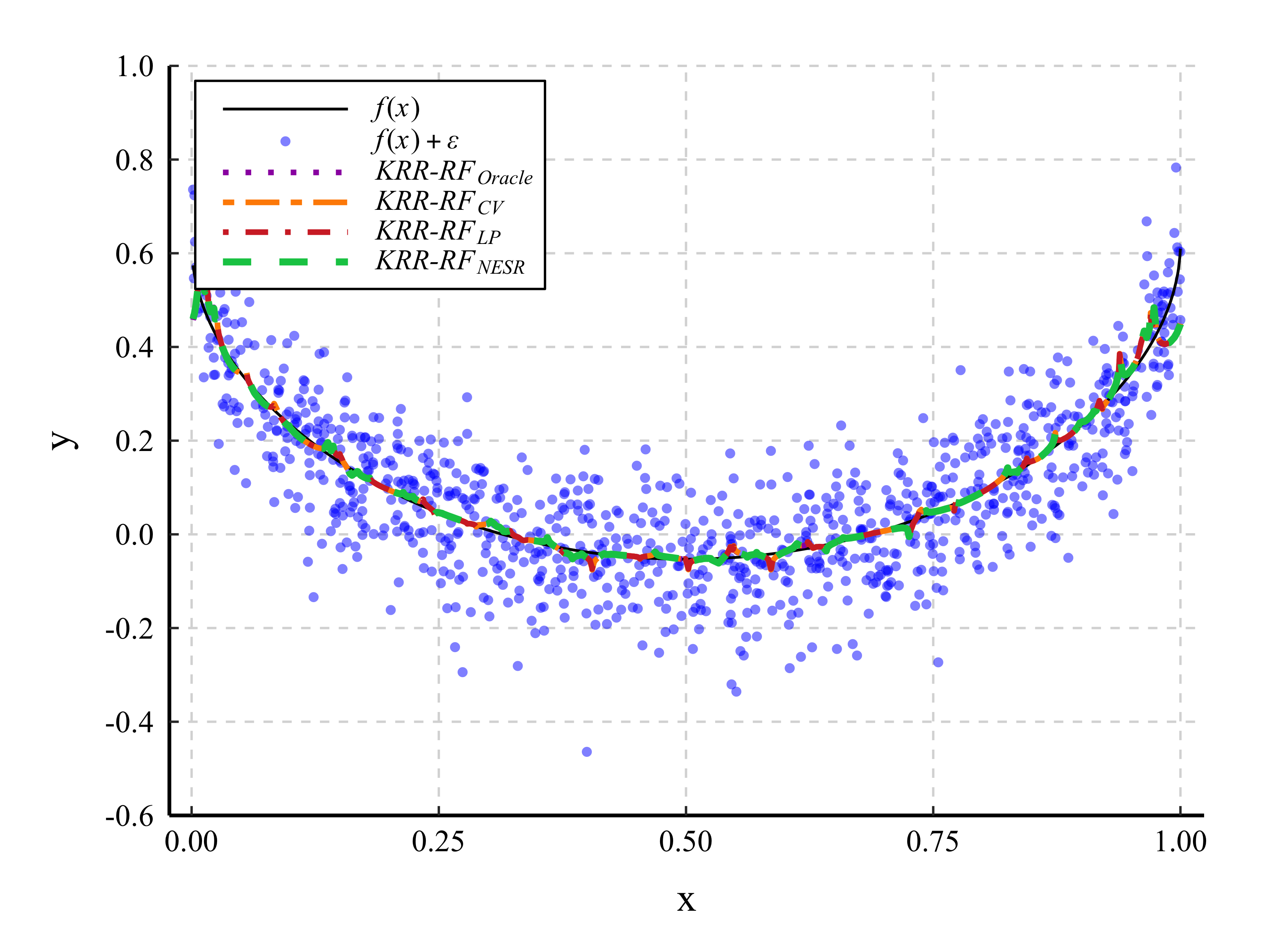}
        \caption{$r=0.5, \gamma=0.45$}
        \label{sim:fig1_c}
    \end{subfigure}
    \hfill
    \begin{subfigure}[b]{0.45\linewidth}
        \centering
        \includegraphics[width=\linewidth]{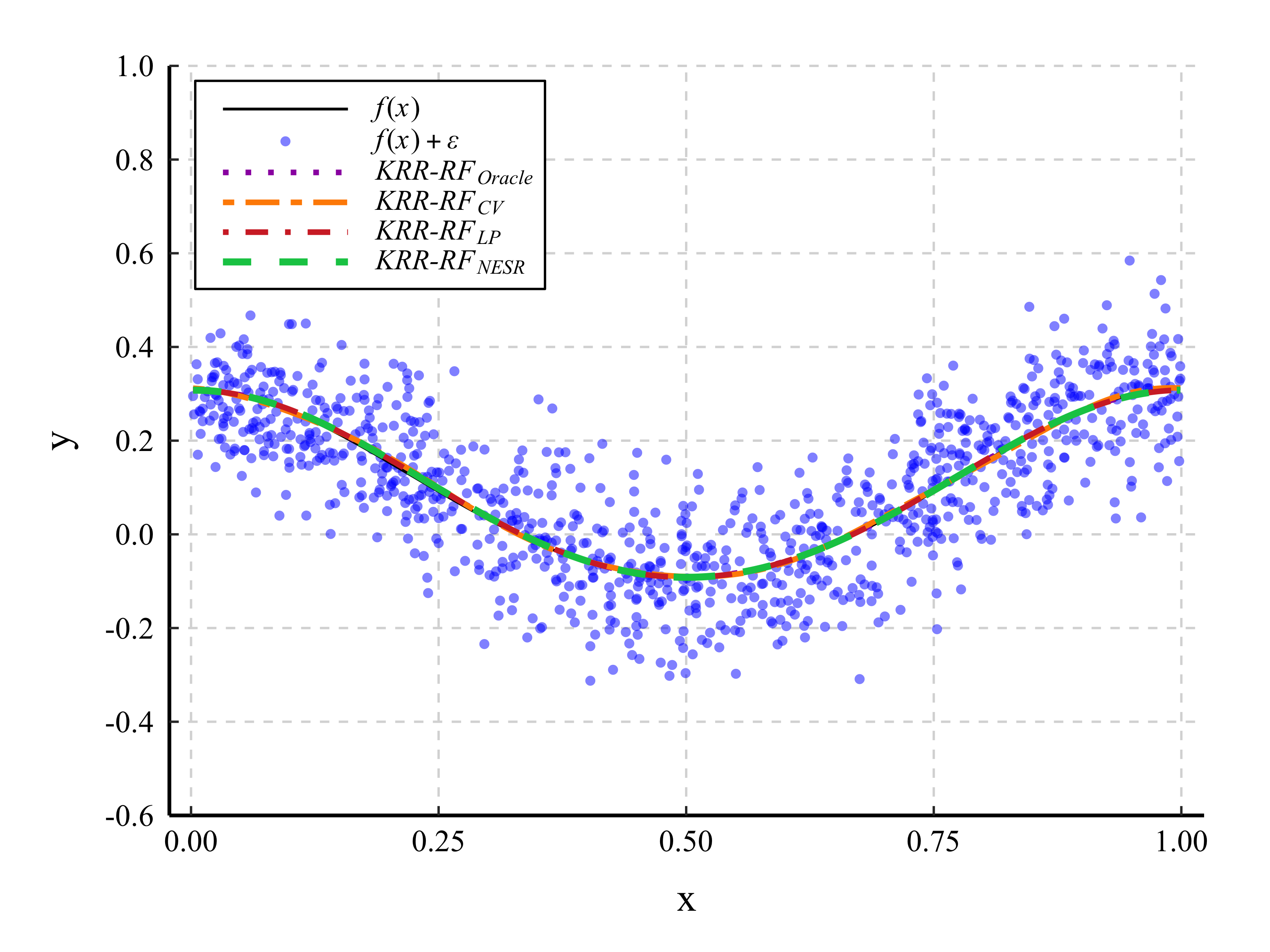}
        \caption{$r=0.4, \gamma=0.1$}
        \label{sim:fig1_d}
    \end{subfigure}
    
    \caption{Curve fitting performance with different $r$ and $\gamma$. In each panel, the solid black line represents the true function $f(x)$, and the blue scatter points show the noisy observations $f(x) + \varepsilon$. Among the four compared estimators, the purple dotted line indicates the Oracle benchmark, and the orange two-dash line corresponds to Cross-Validation (CV). Additionally, the red dot-dash line denotes Lepskii Principle (LP), while the green dashed line illustrates our proposed NESR.}
    \label{sim:fig1}
\end{figure}

To provide a visual comparison, we generate one dataset under each of the four
$(r,\gamma)$ configurations and plot the fitted curves obtained by NESR, LP,
CV, and the Oracle method. As shown in Figure~\ref{sim:fig1}, the four methods
produce broadly similar fits to the underlying regression function in most
settings. In particular, the curves obtained by NESR and LP closely track the
true function over most of the input domain. For the setting
$(r,\gamma)=(0.5,0.45)$, larger discrepancies are observed near the boundaries
of the interval, although the overall fitting patterns of the four methods
remain similar.

\subsection{Effect of Training Sample Size}

\begin{figure}[htbp!]
    \centering
    \begin{subfigure}[b]{0.45\linewidth}
        \centering
        \includegraphics[width=\linewidth]{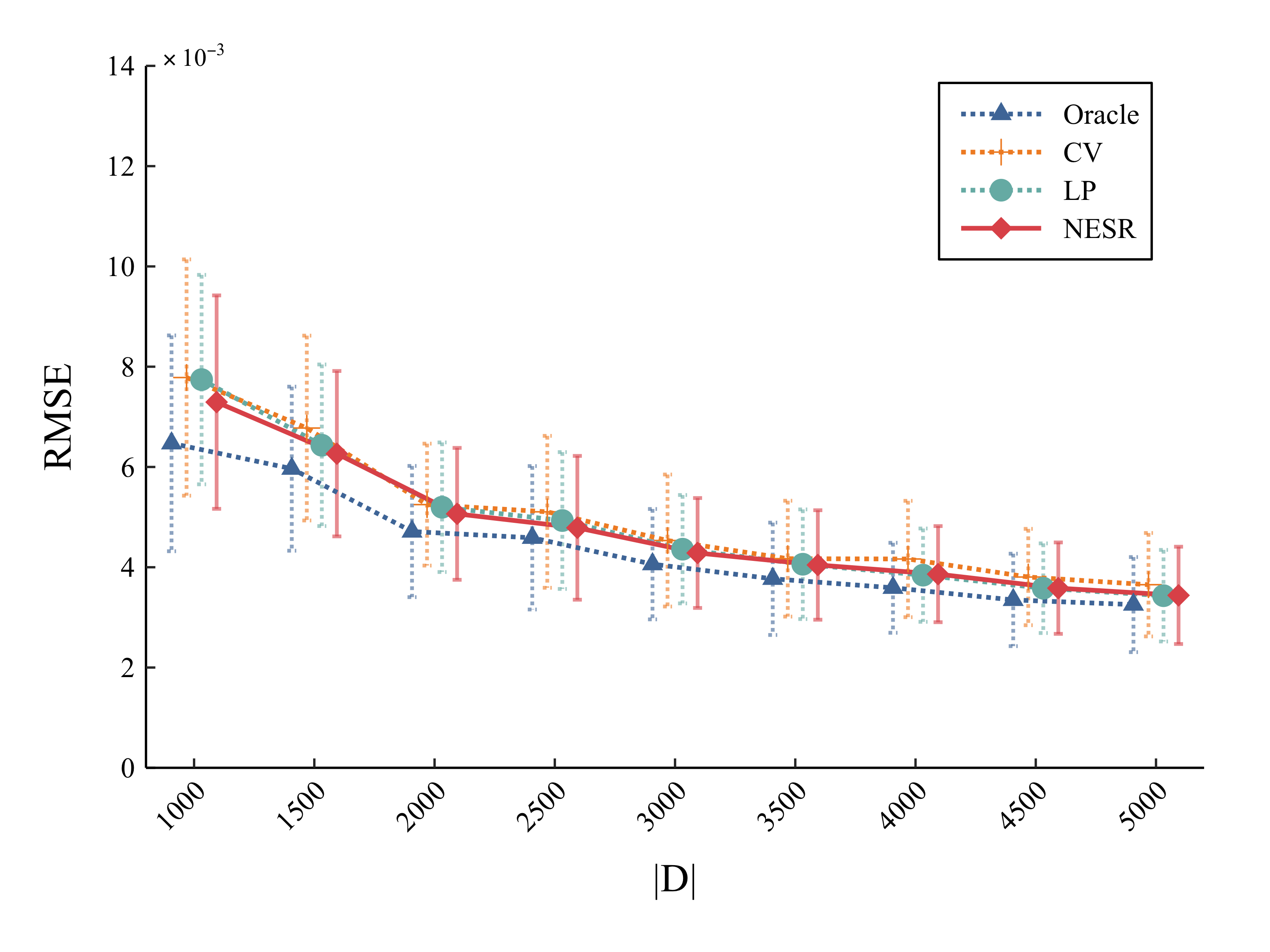}
        \caption{$r=0.8, \gamma=0.2$}
        \label{sim:fig2_a}
    \end{subfigure}
    \hfill
    \begin{subfigure}[b]{0.45\linewidth}
        \centering
        \includegraphics[width=\linewidth]{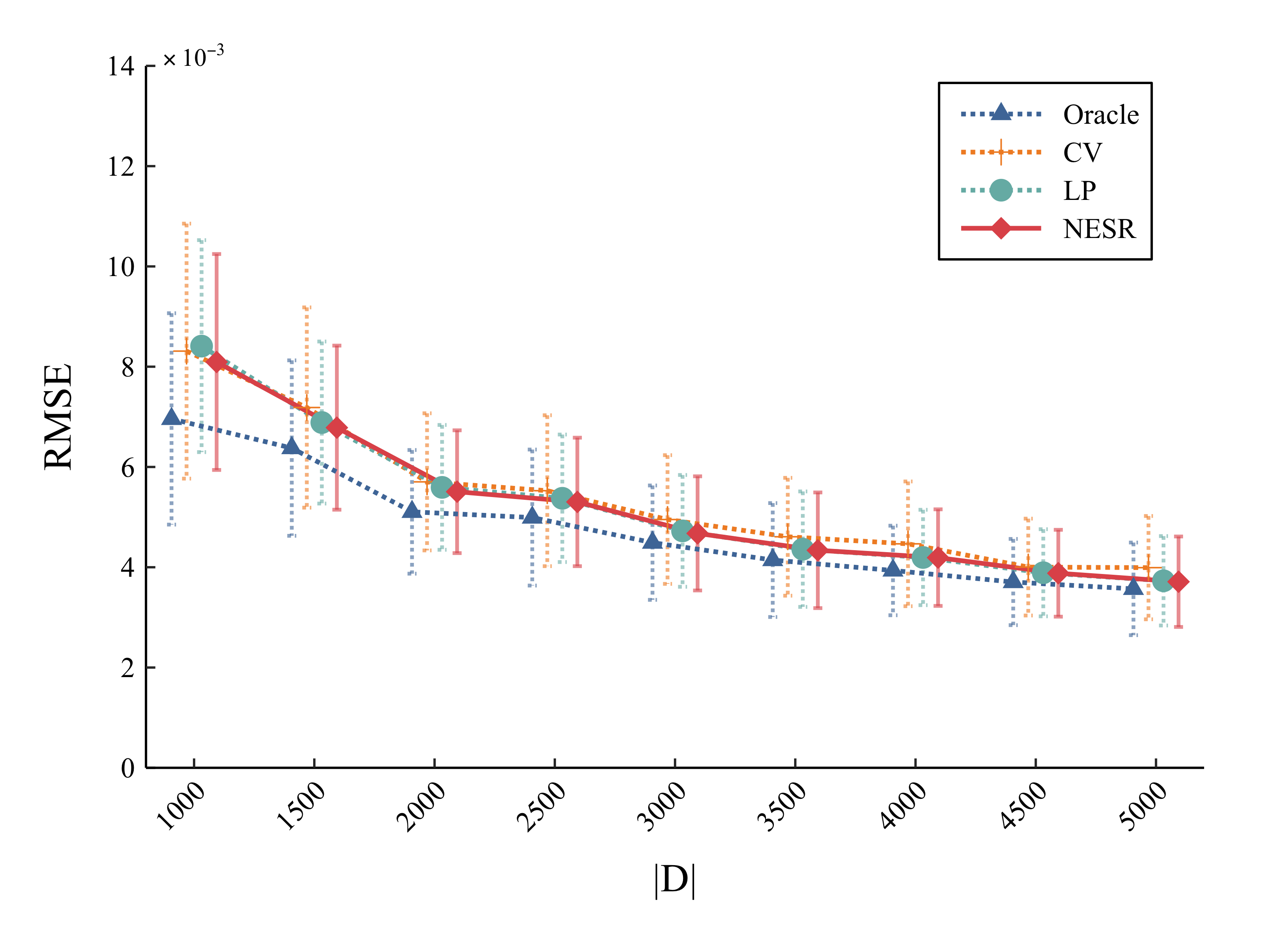}
        \caption{$r=0.6, \gamma=0.2$}
        \label{sim:fig2_b}
    \end{subfigure}
    
    \vspace{0.15cm}
    
    \begin{subfigure}[b]{0.45\linewidth}
        \centering
        \includegraphics[width=\linewidth]{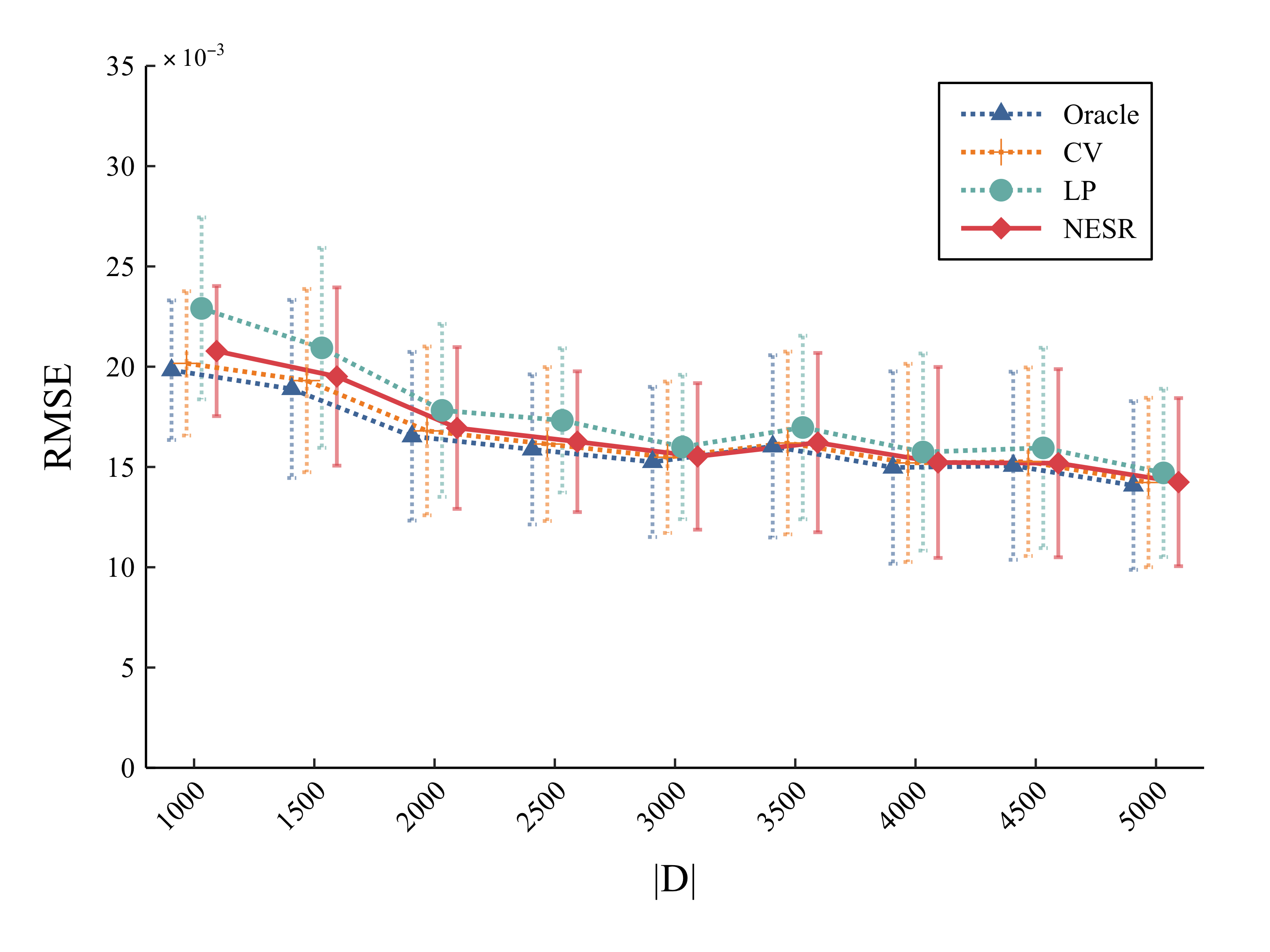}
        \caption{$r=0.5, \gamma=0.45$}
        \label{sim:fig2_c}
    \end{subfigure}
    \hfill
    \begin{subfigure}[b]{0.45\linewidth}
        \centering
        \includegraphics[width=\linewidth]{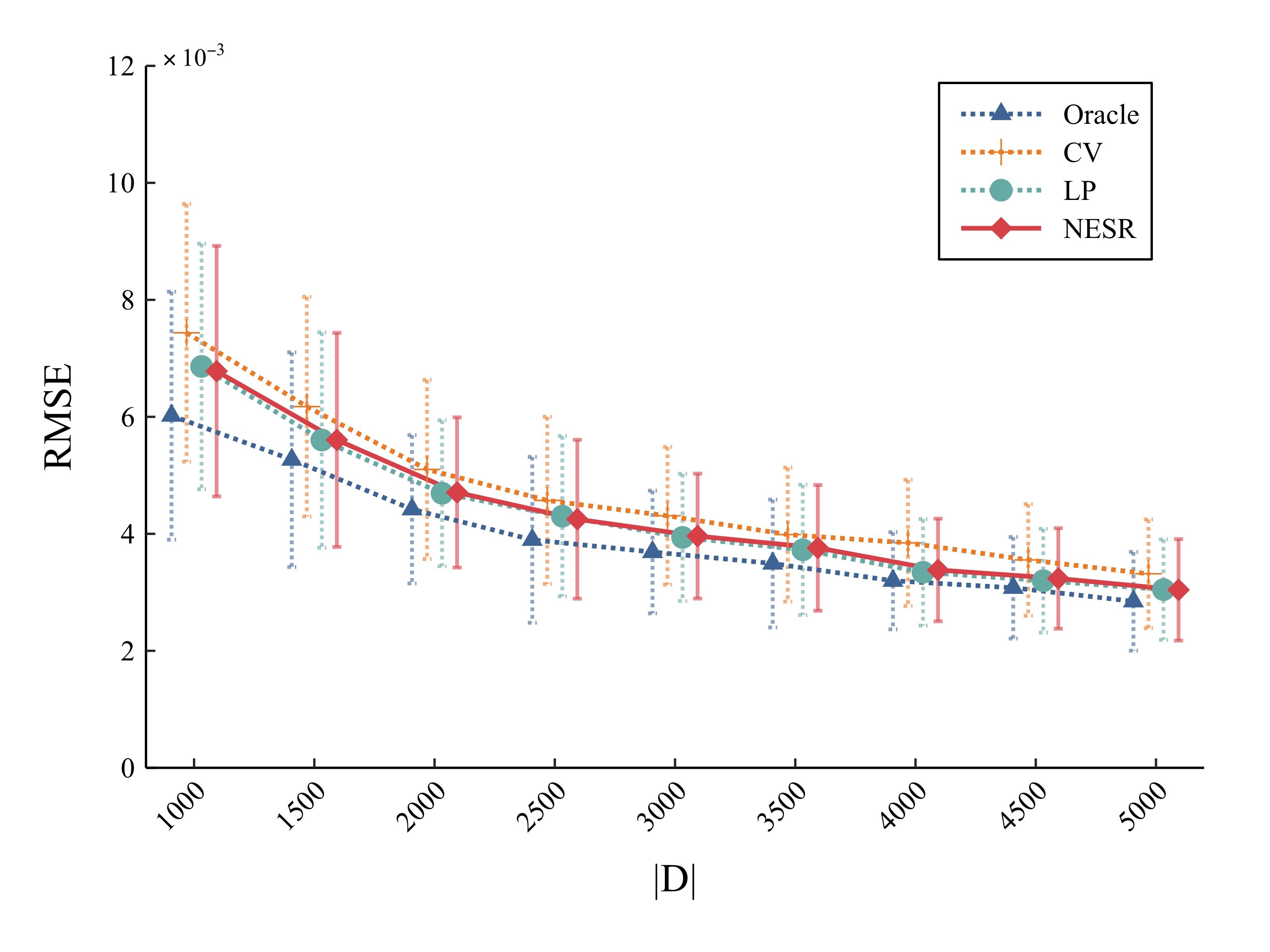}
        \caption{$r=0.4, \gamma=0.1$}
        \label{sim:fig2_d}
    \end{subfigure}
    
    \caption{RMSE vs number of training data $|D|$ for four approaches with different $r$ and $\gamma$.}
    \label{sim:fig2}
\end{figure}

To examine the effect of the training sample size on prediction performance,
we compare NESR, LP, and CV with the Oracle benchmark. In this experiment, the
test sample size is fixed at $|D'|=1000$ and the number of random features is
fixed at $M=100$, while the training sample size varies over
$
    |D| \in \{1000,1500,\ldots,5000\}.
$
Figure~\ref{sim:fig2} reports the average RMSE and the corresponding error bars
for the four methods. Overall, the RMSE tends to decrease as the training
sample size increases across the four $(r,\gamma)$ configurations.

For $(r,\gamma)=(0.8,0.2)$ and $(0.6,0.2)$, NESR, LP, and CV achieve broadly
comparable RMSE values over the range of sample sizes considered. The error
bars of NESR and LP are generally narrower than those of CV, and the
difference between NESR or LP and the Oracle benchmark decreases as $|D|$
increases. NESR also gives slightly smaller average RMSE than LP at several
sample sizes. For $(r,\gamma)=(0.4,0.1)$, both NESR and LP attain lower RMSE than CV over
most of the sample-size range, with NESR and LP showing similar overall
behavior. In the setting $(r,\gamma)=(0.5,0.45)$, NESR gives the lowest RMSE
among the three feasible selection methods for most of the reported sample
sizes and remains close to the Oracle benchmark. The error bars of the four
methods are of similar magnitude in this setting.

% Figure 2: RMSE vs N

\begin{figure}[htbp!]
    \begin{subfigure}[b]{0.45\linewidth}
        \centering
        \includegraphics[width=\linewidth]{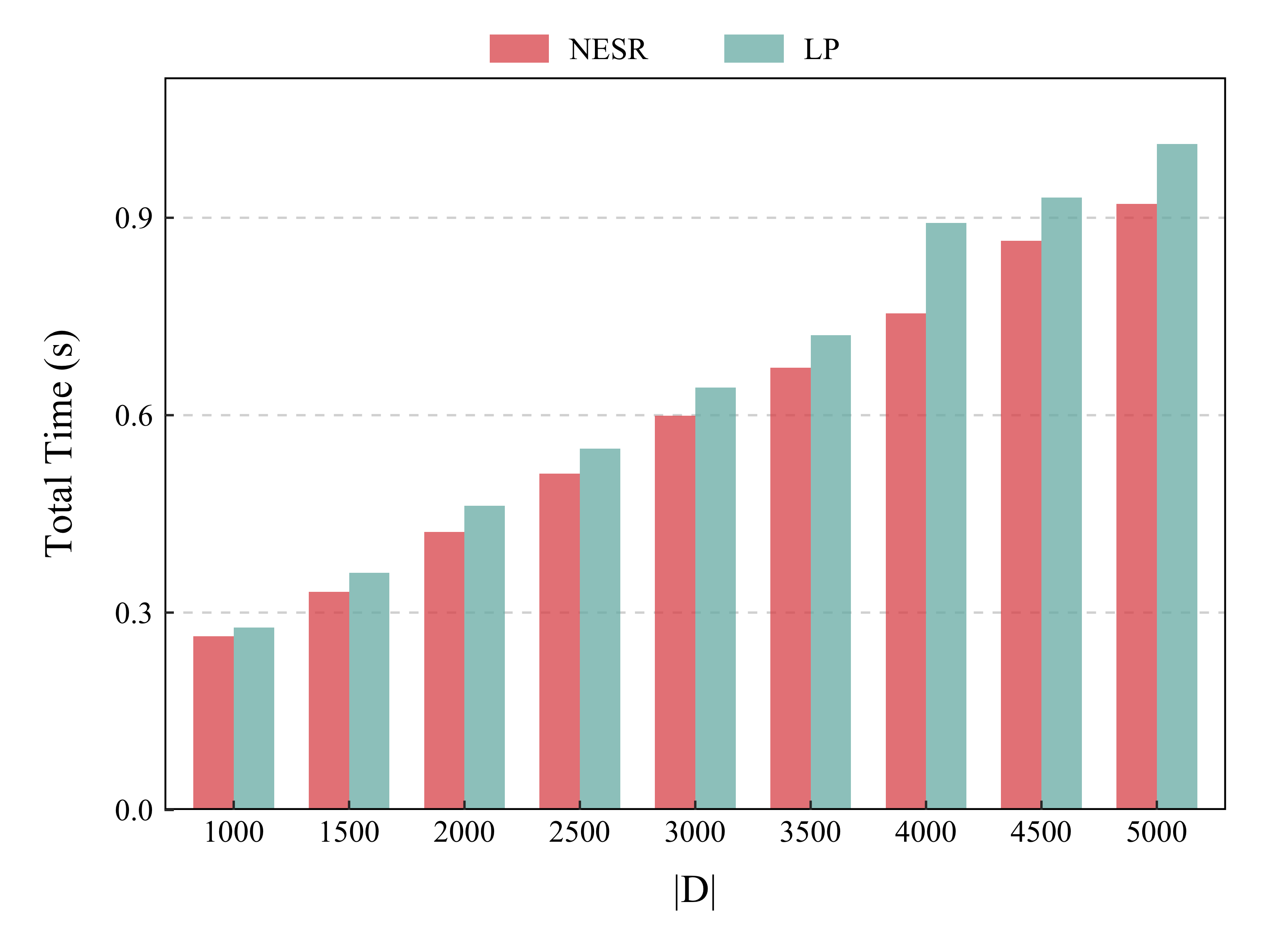}
        \caption{$r=0.8, \gamma=0.2$}
        \label{sim:fig3_a}
    \end{subfigure}
    \hfill
    \begin{subfigure}[b]{0.45\linewidth}
        \centering
        \includegraphics[width=\linewidth]{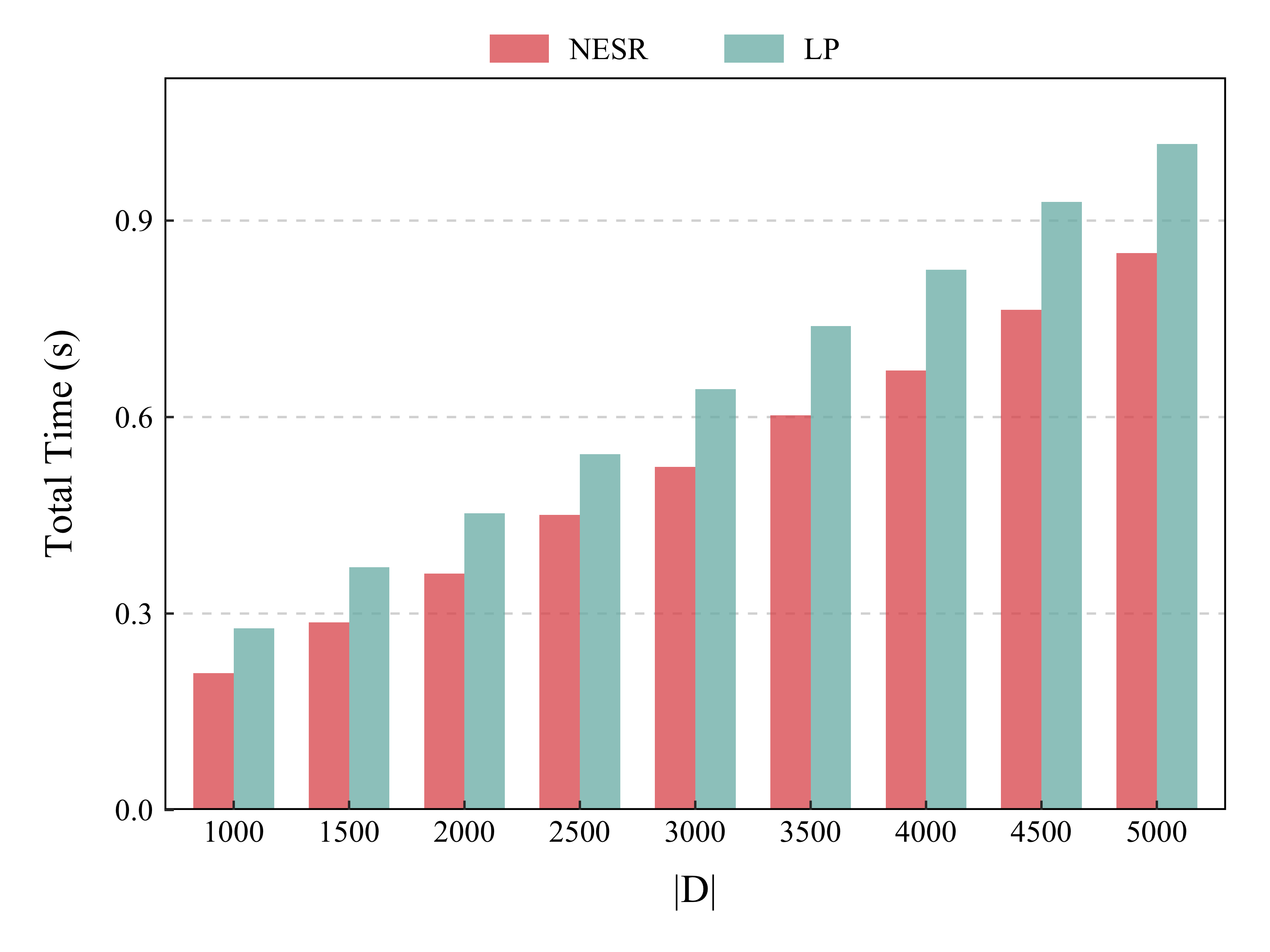}
        \caption{$r=0.6, \gamma=0.2$}
        \label{sim:fig3_b}
    \end{subfigure}
    
    \vspace{0.15cm}
    
    \begin{subfigure}[b]{0.45\linewidth}
        \centering
        \includegraphics[width=\linewidth]{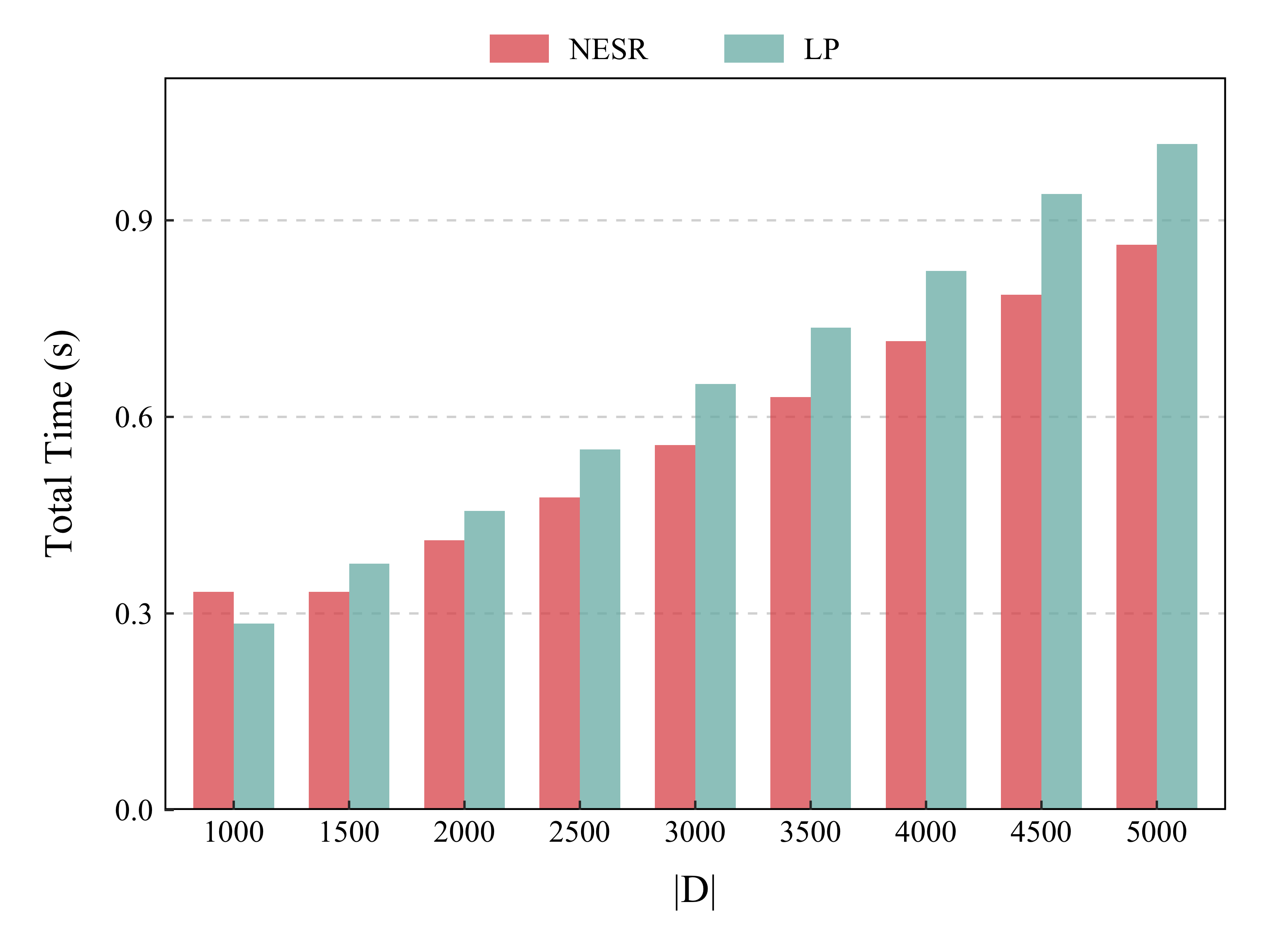}
        \caption{$r=0.5, \gamma=0.45$}
        \label{sim:fig3_c}
    \end{subfigure}
    \hfill
    \begin{subfigure}[b]{0.45\linewidth}
        \centering
        \includegraphics[width=\linewidth]{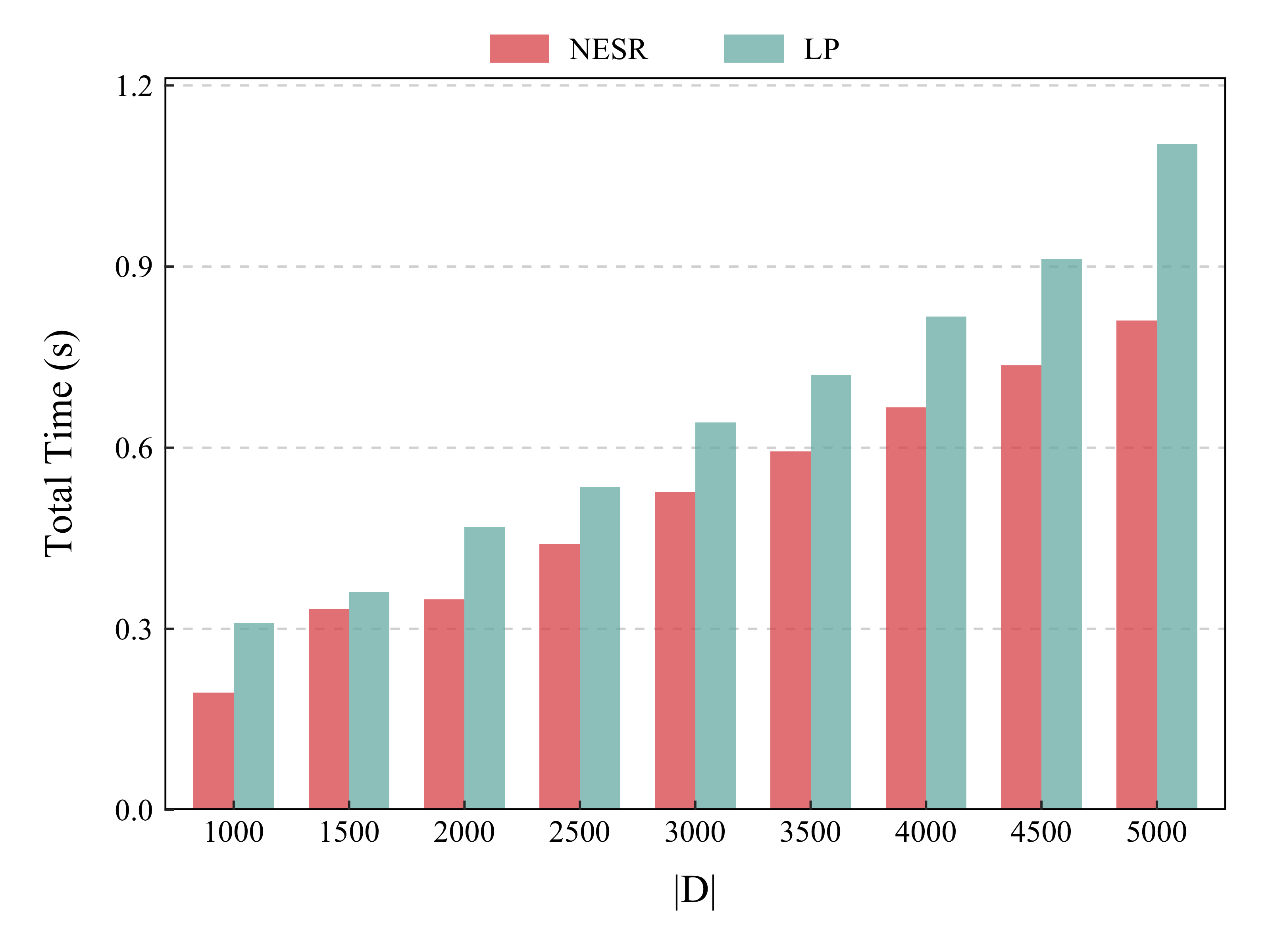}
        \caption{$r=0.4, \gamma=0.1$}
        \label{sim:fig3_d}
    \end{subfigure}
    
    \caption{Run time vs number of training data $|D|$ for NESR and LP with different $r$ and $\gamma$.}
    \label{sim:fig3}
\end{figure}

Beyond prediction accuracy, we also examine the computational cost of NESR and LP as the
training sample size increases. Figure~\ref{sim:fig3} reports the total running time over
50 independent trials. For this experiment, we use restricted candidate ranges centered
around the regions in which the Oracle solutions are observed. Specifically, for
$(r,\gamma)\in\{(0.8,0.2),(0.6,0.2),(0.5,0.45)\}$, we set $\Lambda_{LP} := \{q^k : k = 4, 5, \dots, 12\}$ for LP, and $\Lambda_{ES} \subseteq [10^{-3}, 10^{-2}]$ with $h = 50$ and $K_{ES} = 20$ for NESR. For the case $(r, \gamma) = \{0.4, 0.1\}$, we set $\Lambda_{LP} := \{q^k : k = 8, 9, \dots, 16\}$ for LP, and $\Lambda_{ES} \subseteq [5 \times 10^{-5}, 2.5 \times 10^{-3}]$ with $h = 200$ and $K_{ES} = 100$ for NESR.

The settings used in Figure~\ref{sim:fig3} differ from those used in
Figure~\ref{sim:fig2}. In Figure~\ref{sim:fig2}, the candidate ranges are chosen to
compare the prediction performance of the different selection rules over a broader
regularization path. In Figure~\ref{sim:fig3}, narrower candidate ranges are used to
focus on the computational cost of the parameter-selection procedures in regions
relevant to the fitted models. Consequently, the prediction errors reported in the two
figures are not directly comparable. In particular, restricting the LP candidate set may
exclude some values considered in Figure~\ref{sim:fig2} and can therefore lead to a
larger RMSE. 

Under these restricted candidate ranges, NESR generally requires less running time than
LP in the experiments shown in Figure~\ref{sim:fig3}, while maintaining comparable
prediction accuracy. These results illustrate the computational benefit of replacing
all-pairs comparisons with neighboring comparisons along the regularization path. Since
the candidate ranges in this experiment are informed by the Oracle solutions, the timing
results should be interpreted as a comparison under the specified candidate paths rather
than as a fully data-driven end-to-end benchmark.

% Figure 3: Run Time vs N

\subsection{Effect of Random Features}

\begin{figure}[htbp!]
    \centering
    \begin{subfigure}[b]{0.45\linewidth}
        \centering
        \includegraphics[width=\linewidth]{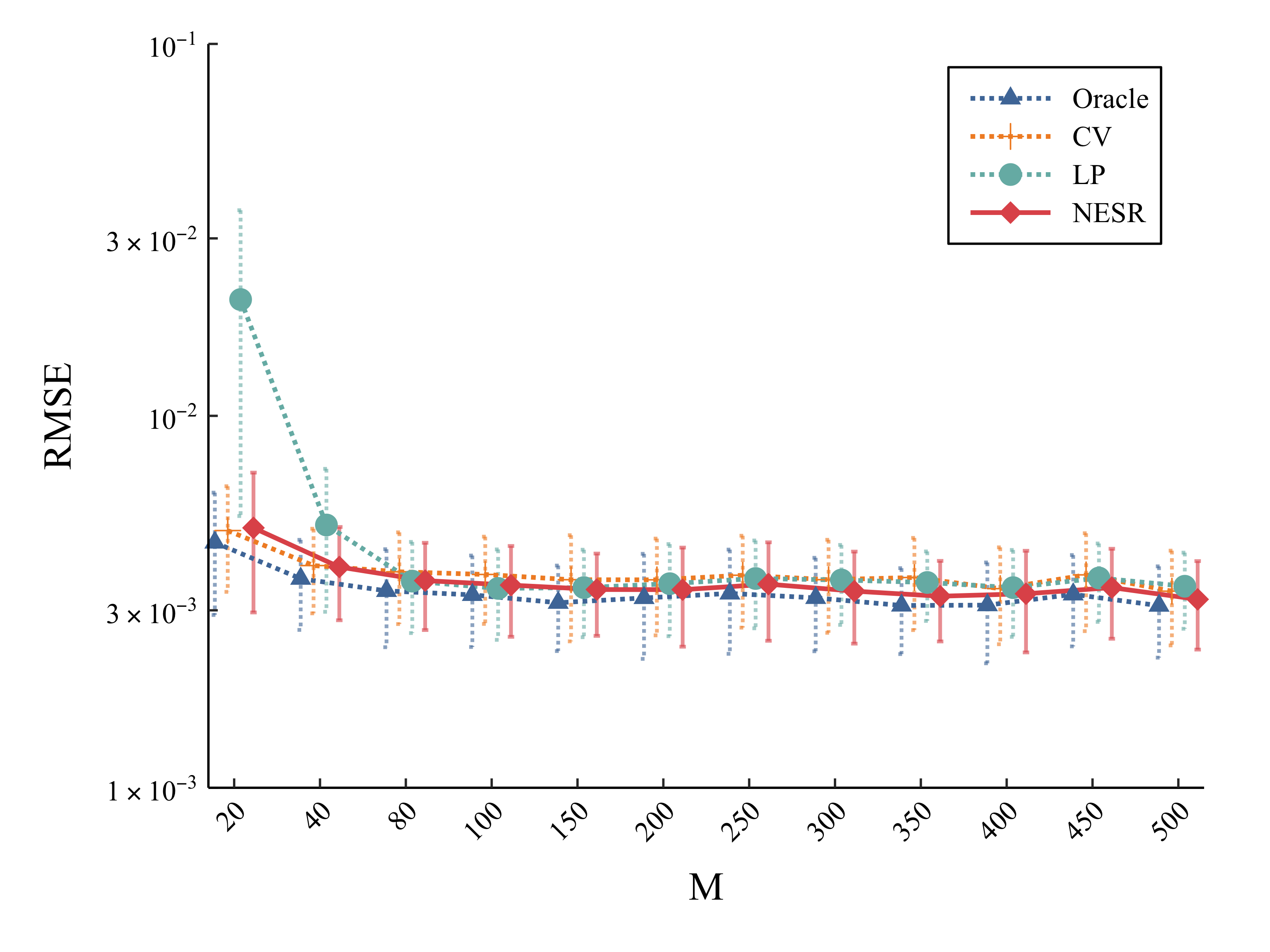}
        \caption{$r=0.8, \gamma=0.2$}
        \label{sim:fig4_a}
    \end{subfigure}
    \hfill
    \begin{subfigure}[b]{0.45\linewidth}
        \centering
        \includegraphics[width=\linewidth]{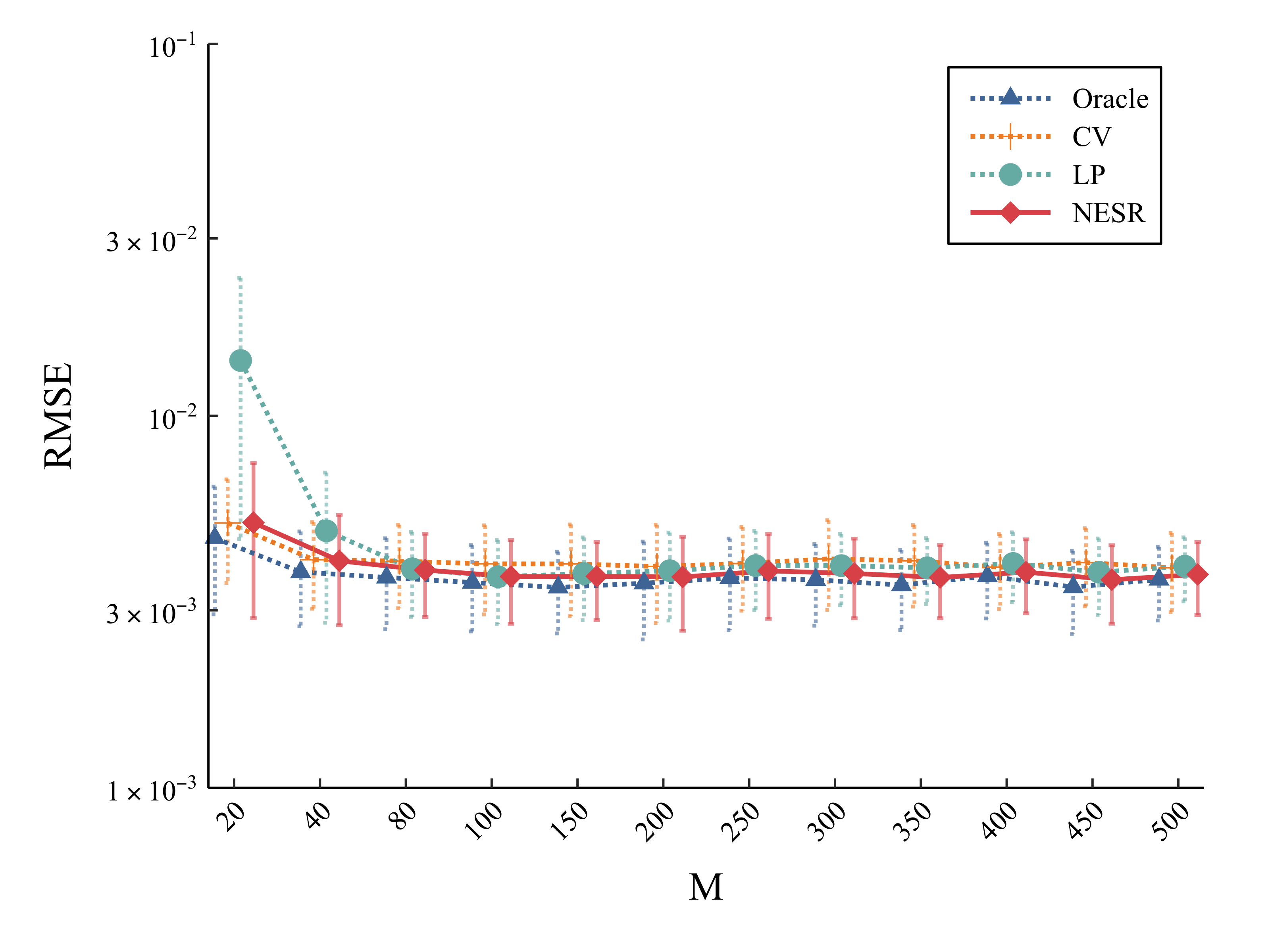}
        \caption{$r=0.6, \gamma=0.2$}
        \label{sim:fig4_b}
    \end{subfigure}
    
    \vspace{0.15cm}
    
    \begin{subfigure}[b]{0.45\linewidth}
        \centering
        \includegraphics[width=\linewidth]{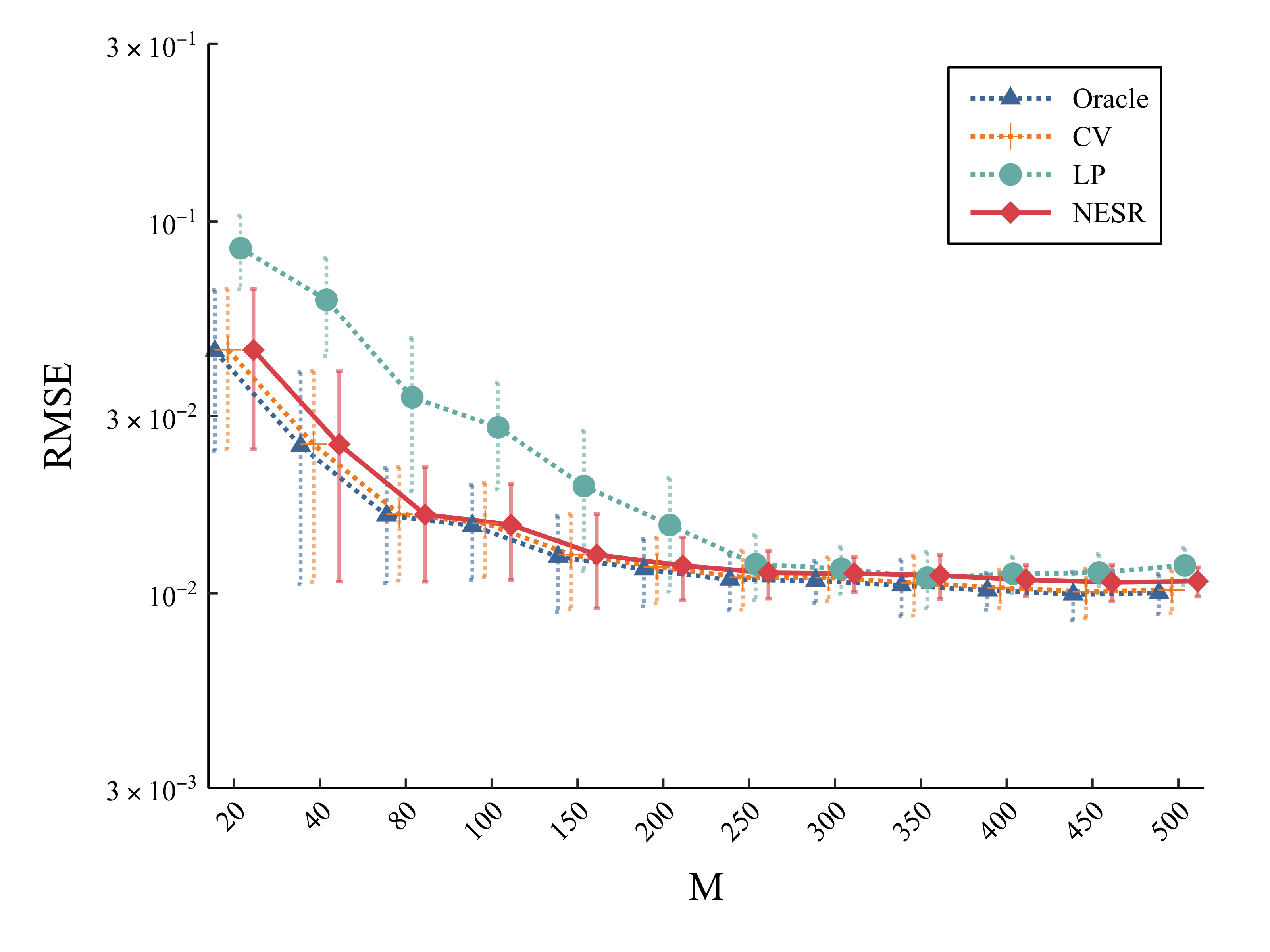}
        \caption{$r=0.5, \gamma=0.45$}
        \label{sim:fig4_c}
    \end{subfigure}
    \hfill
    \begin{subfigure}[b]{0.45\linewidth}
        \centering
        \includegraphics[width=\linewidth]{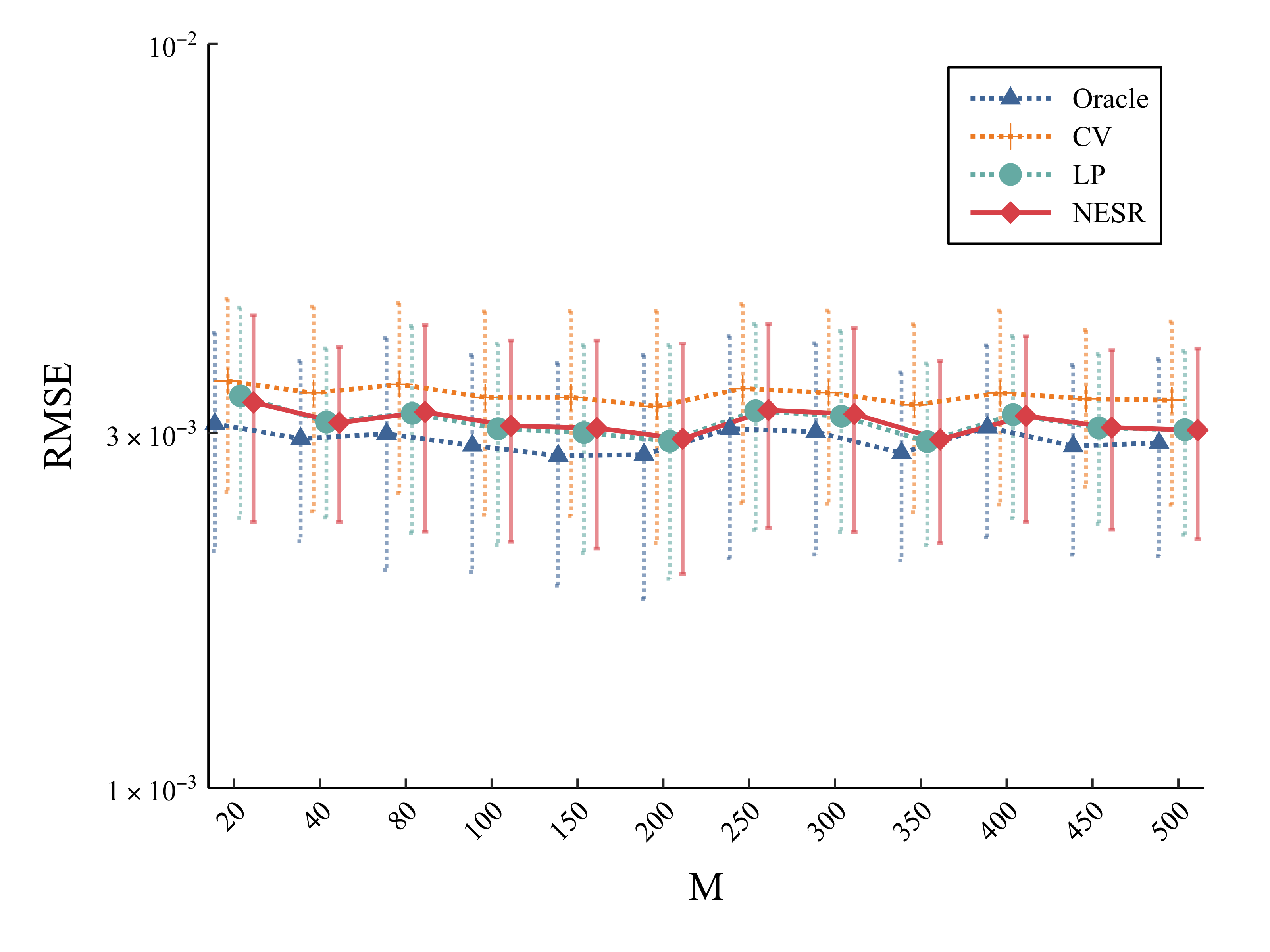}
        \caption{$r=0.4, \gamma=0.1$}
        \label{sim:fig4_d}
    \end{subfigure}
    
    \caption{RMSE vs number of random features $M$ for four approaches with different $r$ and $\gamma$.}
    \label{sim:fig4}
\end{figure}

In addition to varying the training sample size, we examine the effect of the
number of random features $M$ on prediction performance. We fix the training
sample size at $|D|=5000$ and the test sample size at $|D'|=1000$, and vary $M \in \{20, 40, 80, 100, 150, 200, \dots, 500\}$. Figure~\ref{sim:fig4} reports the corresponding RMSE results. Overall, the
prediction errors tend to decrease or level off as $M$ increases, with relatively
small changes observed for moderate to large values of $M$ in most settings.

For $(r,\gamma)=(0.8,0.2)$ and $(0.6,0.2)$, LP gives a larger RMSE than NESR
and CV when the number of random features is small. As $M$ increases, the
difference becomes smaller, while NESR generally maintains a slightly lower
RMSE over the range considered. For $(r,\gamma)=(0.4,0.1)$, NESR and LP both
give lower RMSE than CV for most values of $M$. In the setting
$(r,\gamma)=(0.5,0.45)$, LP shows a noticeably larger RMSE for small and
moderate values of $M$, and the difference gradually decreases as more random
features are used. Across the four settings, the error bars of the three
selection methods are generally of comparable magnitude.

% Figure 4: RMSE vs M

% Figure 5: Run Time vs M
\begin{figure}[htbp!]
    \begin{subfigure}[b]{0.45\linewidth}
        \centering
        \includegraphics[width=\linewidth]{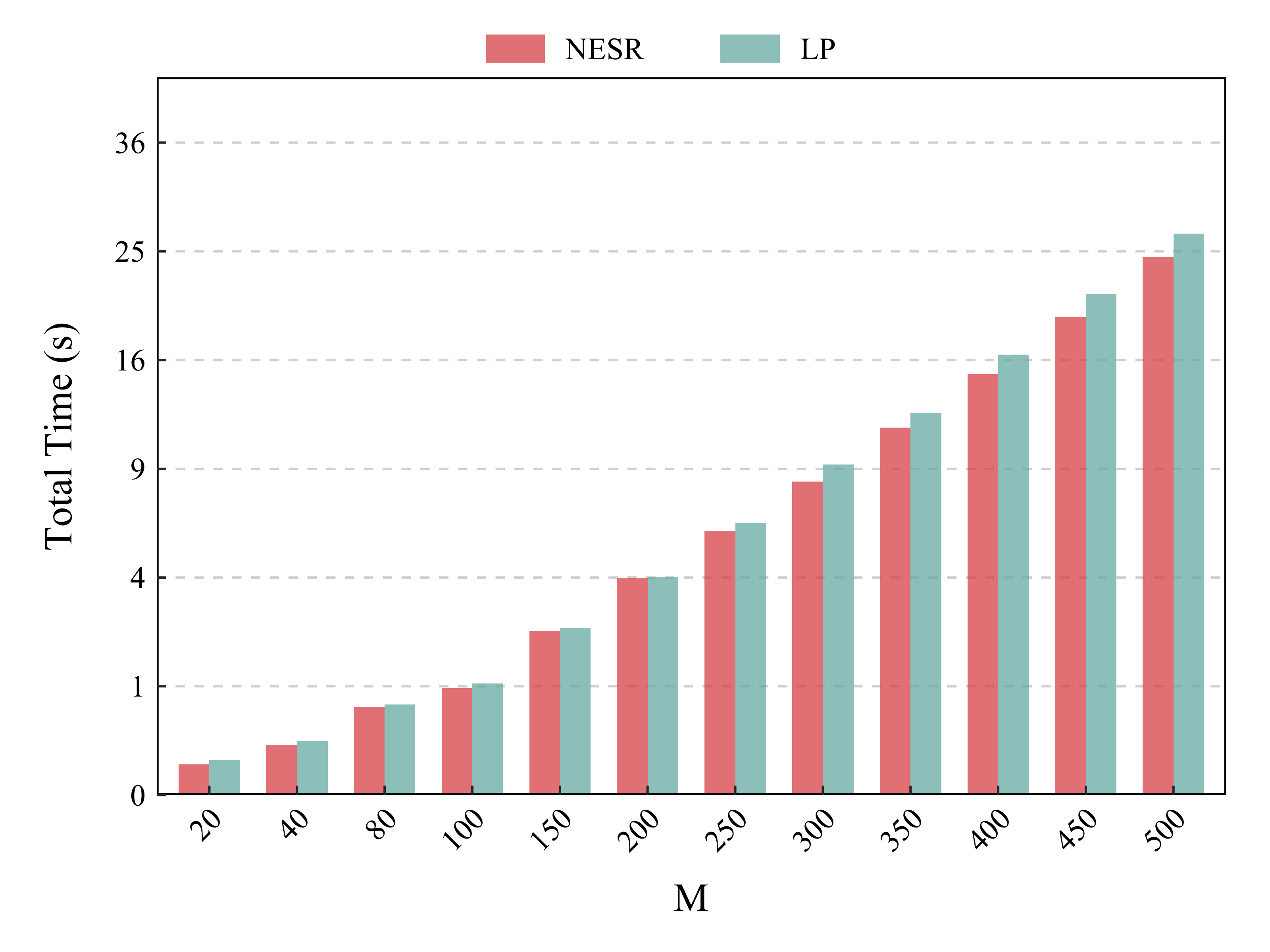}
        \caption{$r=0.8, \gamma=0.2$}
        \label{sim:fig5_a}
    \end{subfigure}
    \hfill
    \begin{subfigure}[b]{0.45\linewidth}
        \centering
        \includegraphics[width=\linewidth]{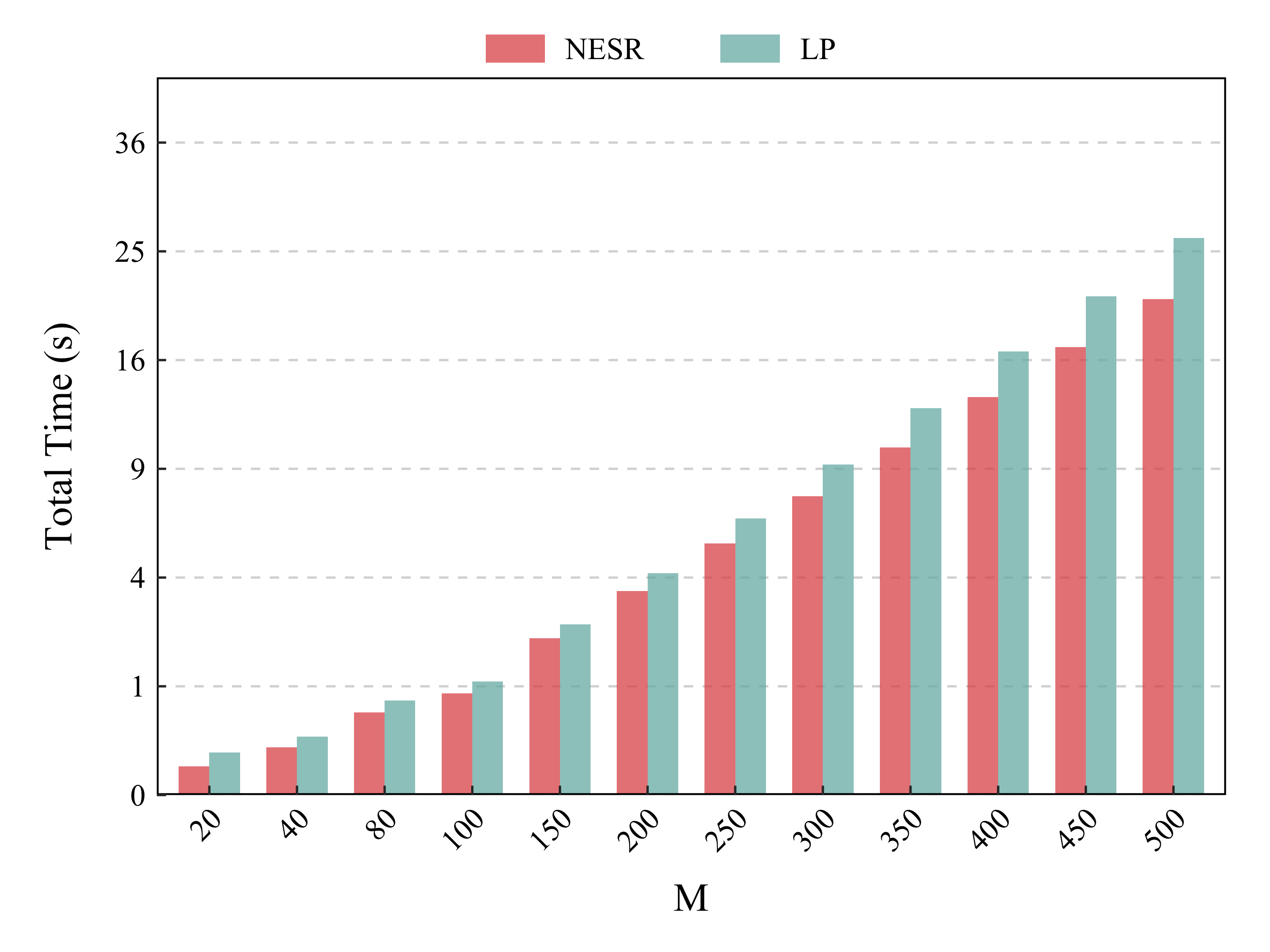}
        \caption{$r=0.6, \gamma=0.2$}
        \label{sim:fig5_b}
    \end{subfigure}
    
    \vspace{0.15cm}
    
    \begin{subfigure}[b]{0.45\linewidth}
        \centering
        \includegraphics[width=\linewidth]{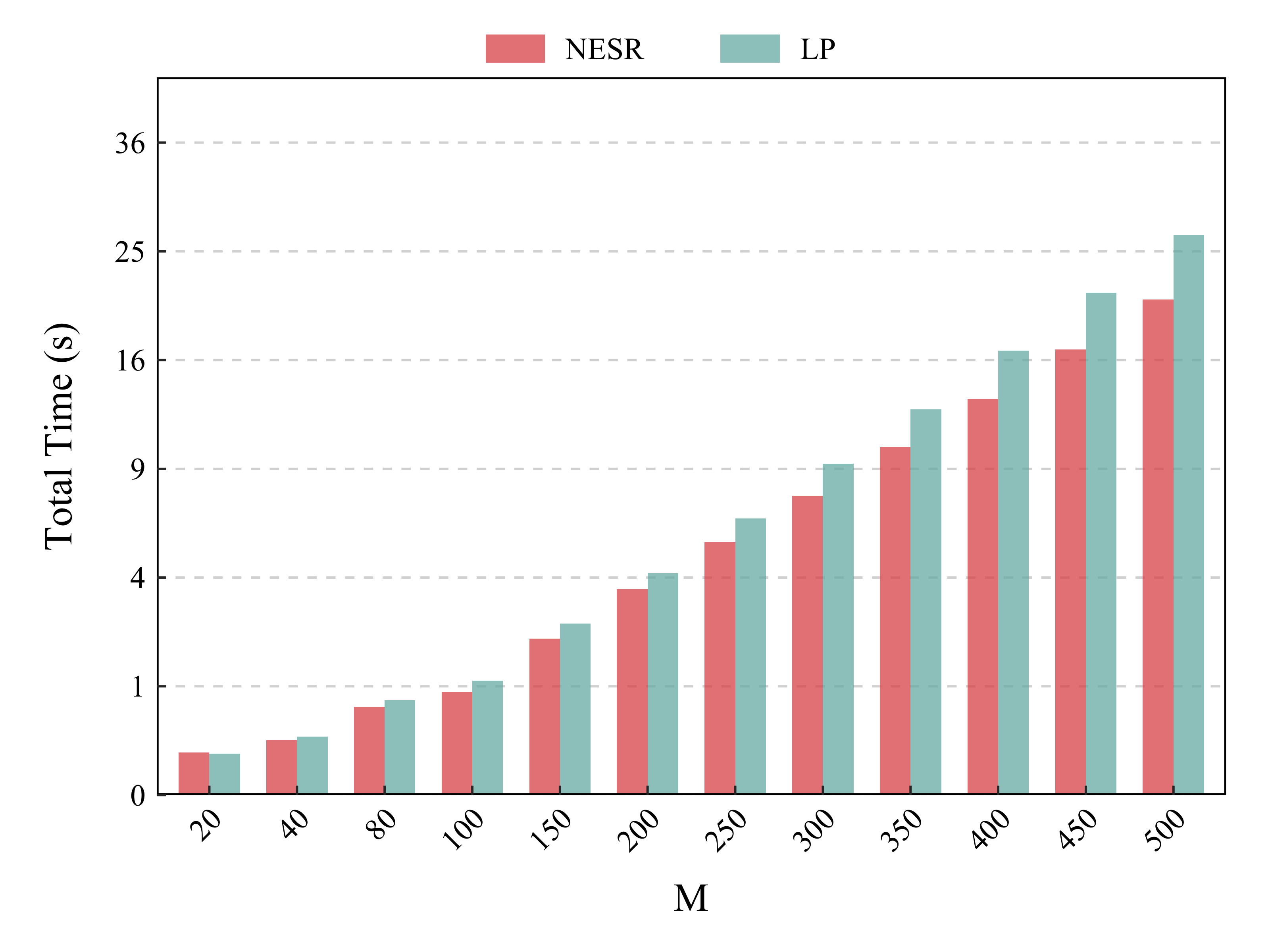}
        \caption{$r=0.5, \gamma=0.45$}
        \label{sim:fig5_c}
    \end{subfigure}
    \hfill
    \begin{subfigure}[b]{0.45\linewidth}
        \centering
        \includegraphics[width=\linewidth]{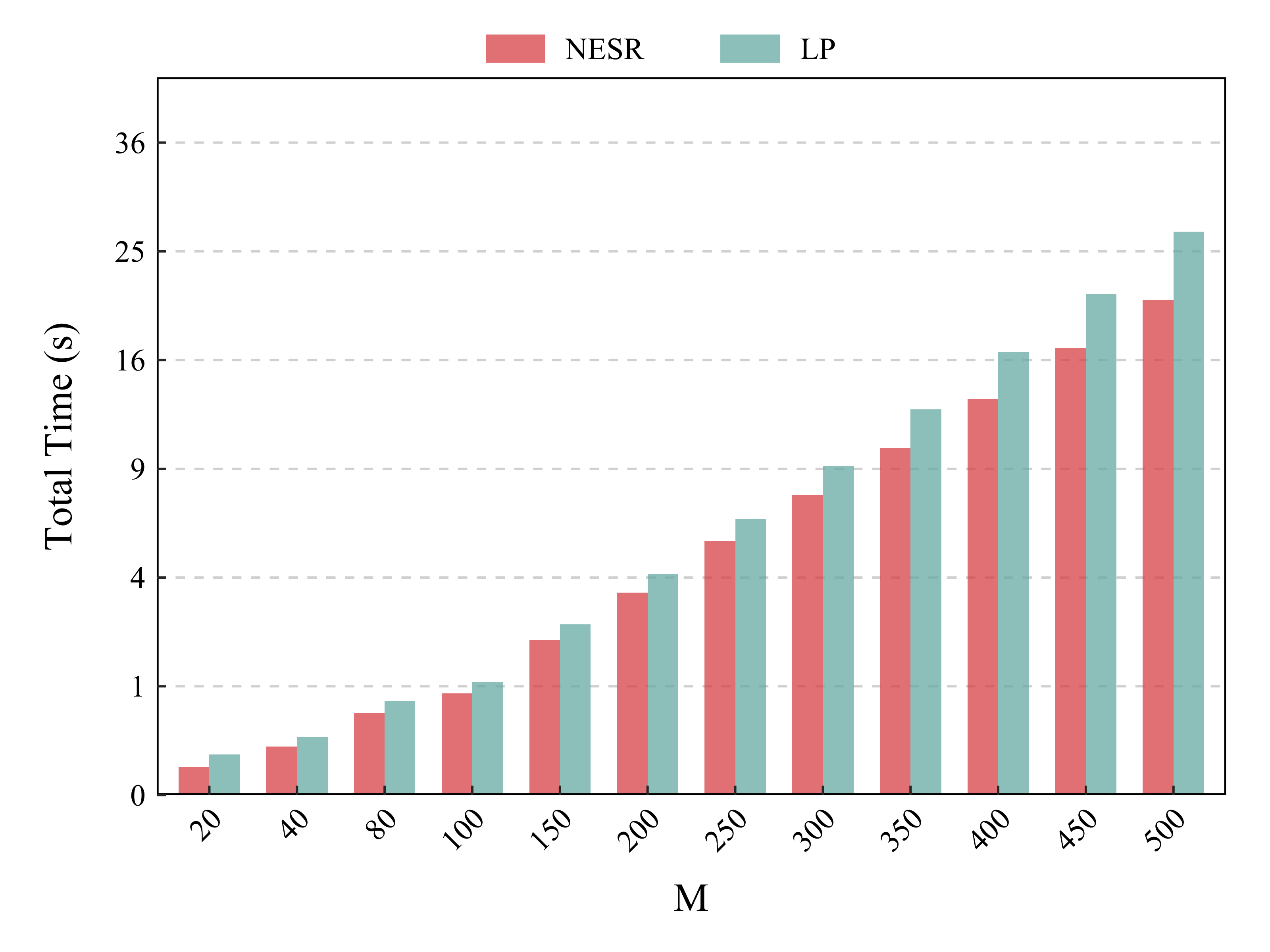}
        \caption{$r=0.4, \gamma=0.1$}
        \label{sim:fig5_d}
    \end{subfigure}
    
    \caption{Run time vs number of random features $M$ for NESR and LP with different $r$ and $\gamma$.}
    \label{sim:fig5}
\end{figure}

Furthermore, we compare the computational cost of NESR and LP as the number
of random features $M$ varies. Using the same candidate-range settings as in
the preceding runtime experiment, Figure~\ref{sim:fig5} reports the total
running time over 50 independent trials. In the reported experiments, NESR
requires less running time than LP across most values of $M$, while the
corresponding RMSE remains comparable to that of LP. The observed runtime
difference is consistent with the smaller number of discrepancy comparisons
required by the neighboring rule.

\subsection{Effect of Subdivision Factor}

Here, the subdivision factor $h$ determines the spacing of the
inverse-regularization grid. Since
$
    \lambda_k=\frac{1}{hk}$, $ \frac{1}{\lambda_{k+1}}-\frac{1}{\lambda_k}=h,
$
the candidate values are equally spaced on the $1/\lambda$ scale. For a fixed
$K_{ES}$, the candidate set ranges from
$
    \lambda_{K_{ES}}=\frac{1}{hK_{ES}}$ to $
    \lambda_1=\frac{1}{h}.
$
Thus, $h$ controls both the location and the spacing of the regularization
path. A smaller $h$ shifts the candidate values toward larger regularization
parameters, whereas a larger $h$ shifts the entire path toward smaller
regularization parameters. Consequently, if $h$ is chosen too small or too
large, the candidate set may fail to cover a region that gives good prediction
performance.

% Figure 6: RMSE vs h
\begin{figure}[htbp!]
    \centering
    \begin{subfigure}[b]{0.45\linewidth}
        \centering
        \includegraphics[width=\linewidth]{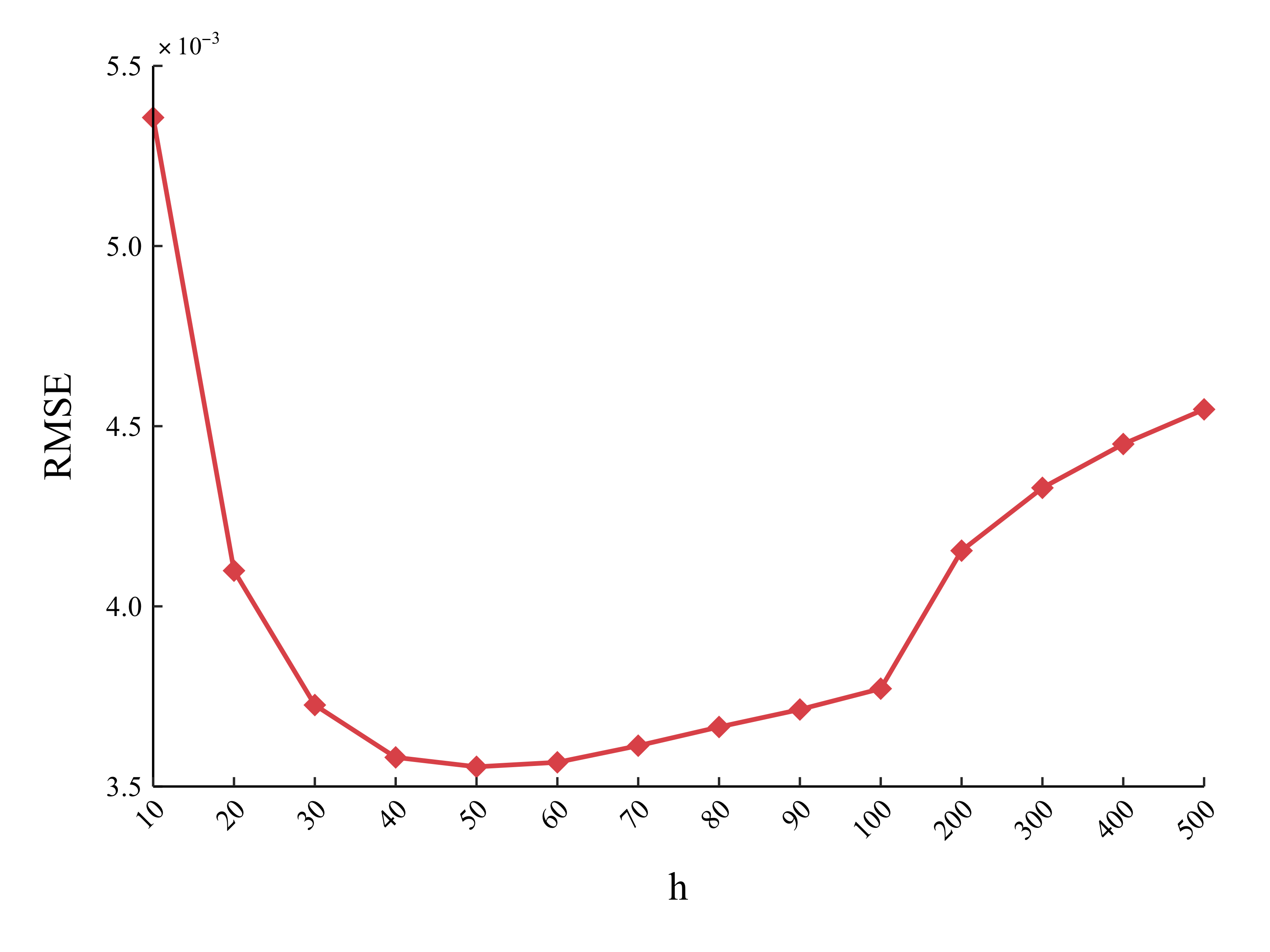}
        \caption{$\{r=0.8, \gamma=0.2\}$}
        \label{sim:fig6_a}
    \end{subfigure}
    \hfill
    \begin{subfigure}[b]{0.45\linewidth}
        \centering
        \includegraphics[width=\linewidth]{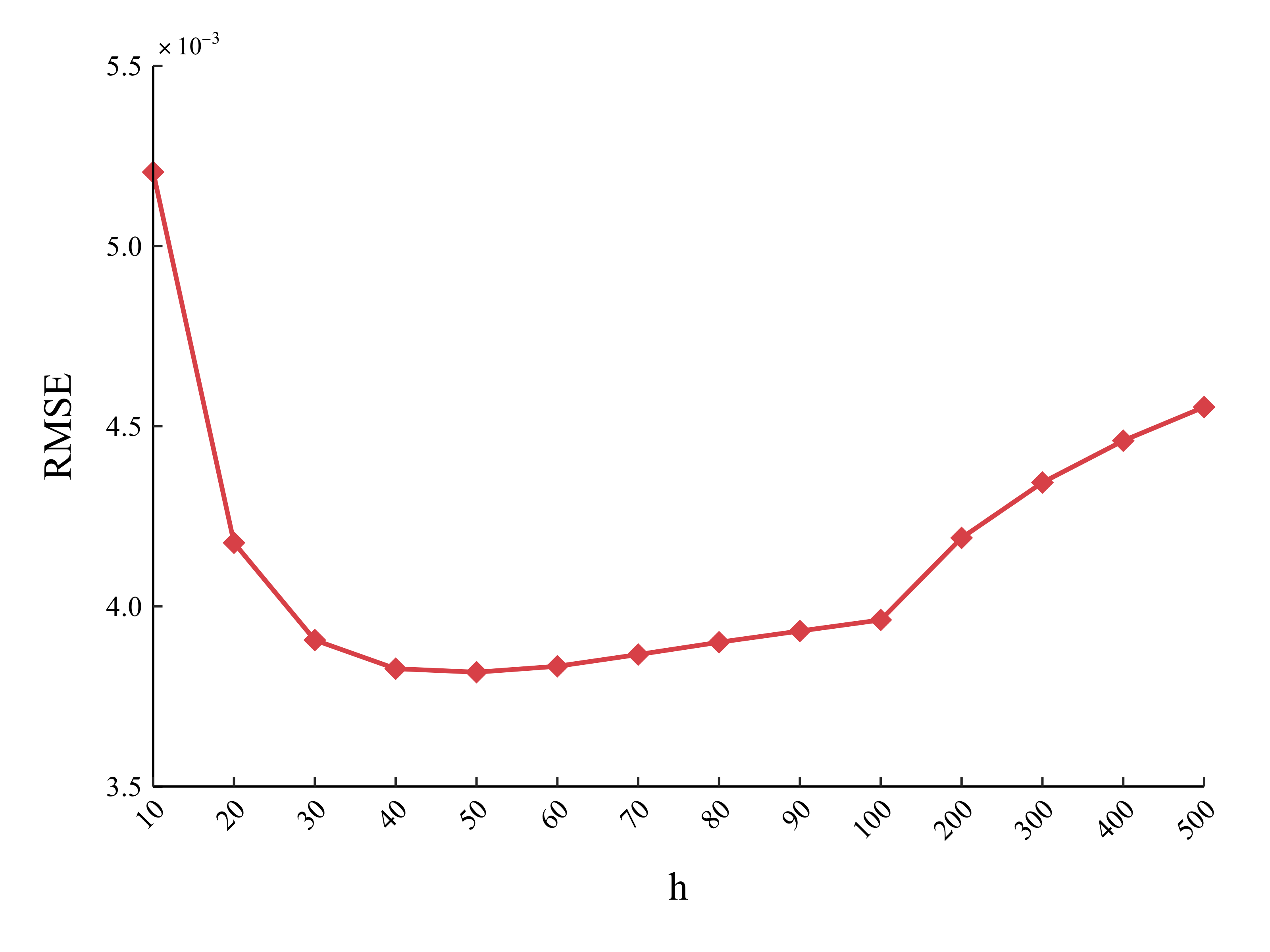}
        \caption{$\{r=0.6, \gamma=0.2\}$}
        \label{sim:fig6_b}
    \end{subfigure}
    
    \vspace{0.4cm} 
    
    \begin{subfigure}[b]{0.45\linewidth}
        \centering
        \includegraphics[width=\linewidth]{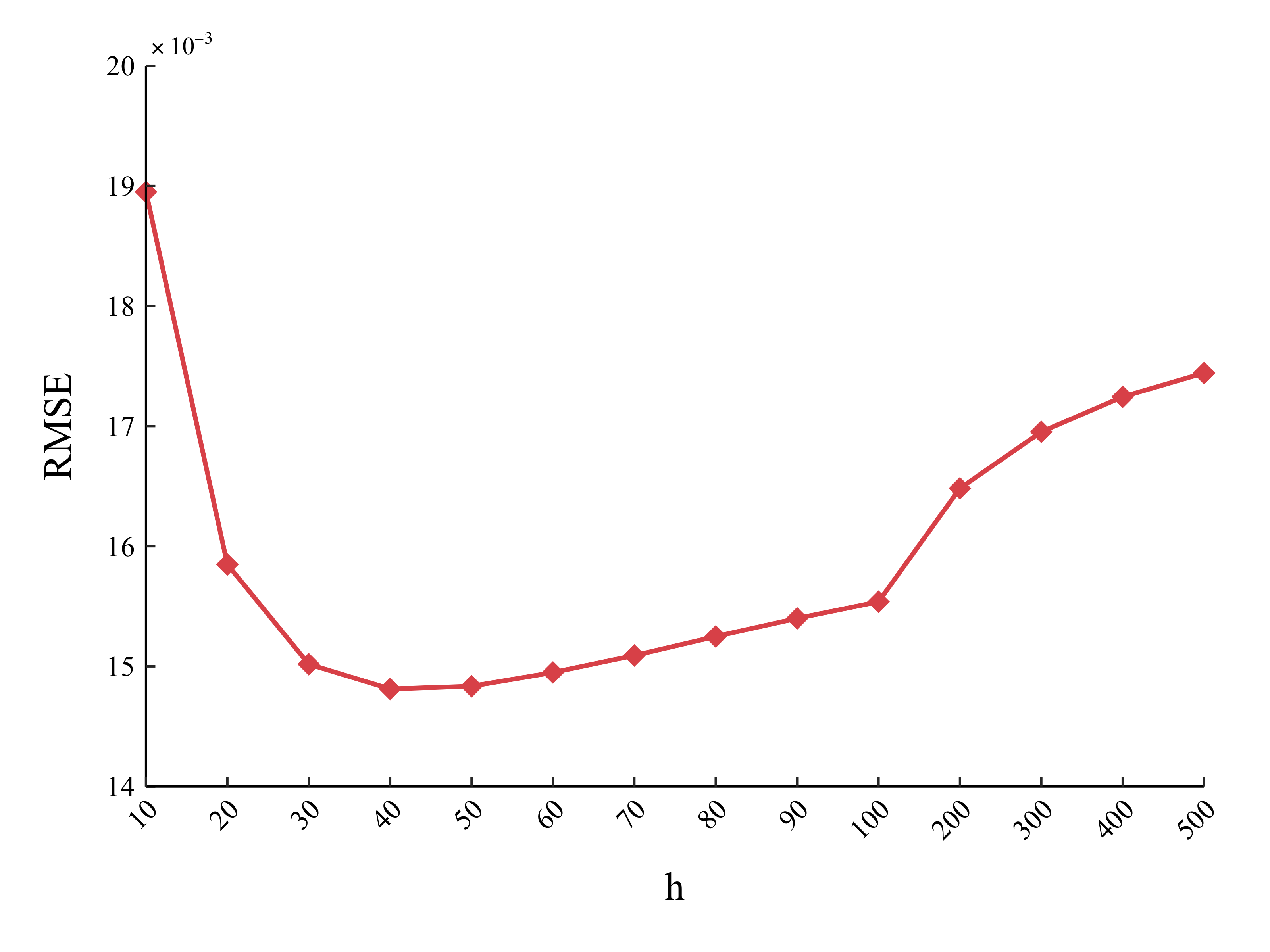}
        \caption{$\{r=0.5, \gamma=0.45\}$}
        \label{sim:fig6_c}
    \end{subfigure}
    \hfill
    \begin{subfigure}[b]{0.45\linewidth}
        \centering
        \includegraphics[width=\linewidth]{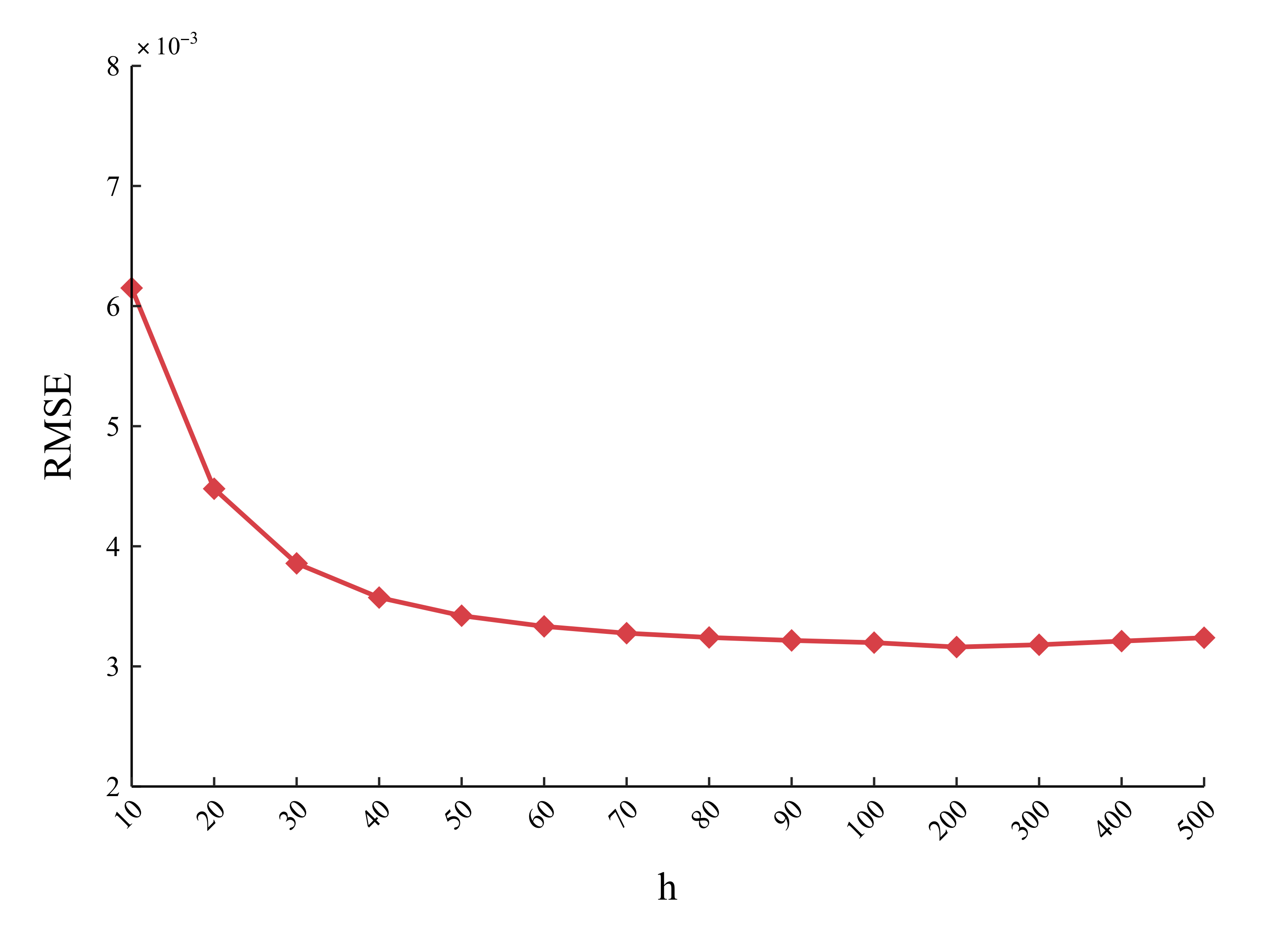}
        \caption{$\{r=0.4, \gamma=0.1\}$}
        \label{sim:fig6_d}
    \end{subfigure}
    
    \caption{RMSE vs subdivision factor $h$ for NESR with different $r$ and $\gamma$.}
    \label{sim:fig6}
\end{figure}

Figure~\ref{sim:fig6} shows the RMSE of NESR for different choices of $h$ under
the four $(r,\gamma)$ configurations. We fix $|D|=5000$ and $|D'|=1000$ in
all experiments. For
$(r,\gamma)\in
    \{(0.8,0.2),(0.6,0.2),(0.5,0.45)\}$,
we set $K_{ES}=20$, while for $(r,\gamma)=(0.4,0.1)$ we set $K_{ES}=100$.
For the first three settings, the smallest observed RMSE values are obtained
for $h$ roughly between $30$ and $60$. For $(r,\gamma)=(0.4,0.1)$, the RMSE
becomes relatively stable for $h>100$, with a small value observed around
$h=200$. Based on these results, we use $h=50$ for the first three settings
and $h=200$ for the last setting in the remaining simulation experiments.
The corresponding parameter choices are reported in
Table~\ref{tab:sim_parameter_settings}.

\subsection{Validation of the Learning Rates}

\begin{figure}[htbp!]
    \centering
    \begin{subfigure}[b]{0.45\linewidth}
        \centering
        \includegraphics[width=\linewidth]{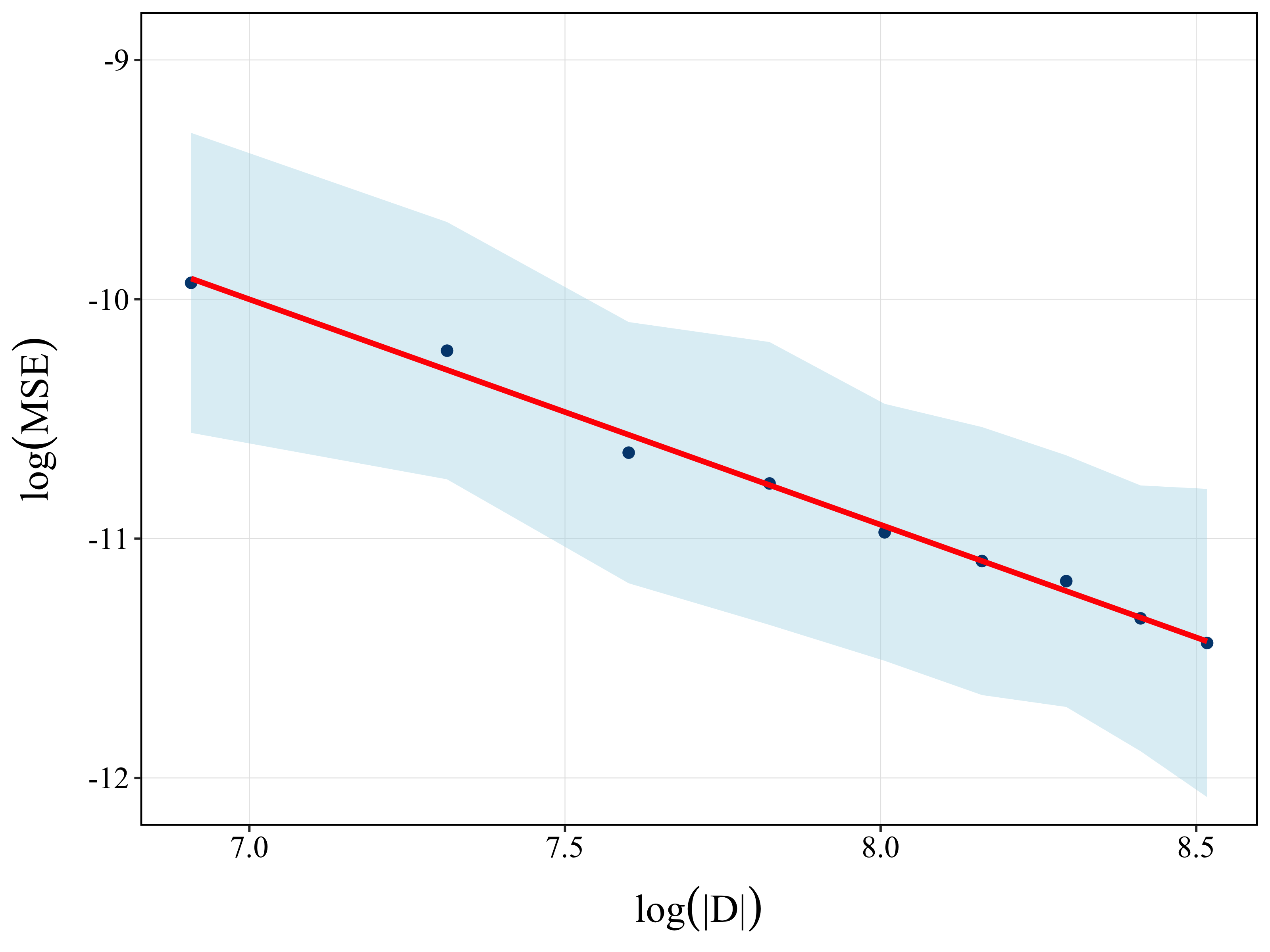}
        \caption{$\{r=0.8, \gamma=0.2\}$}
        \label{sim:fig7_a}
    \end{subfigure}
    \hfill
    \begin{subfigure}[b]{0.45\linewidth}
        \centering
        \includegraphics[width=\linewidth]{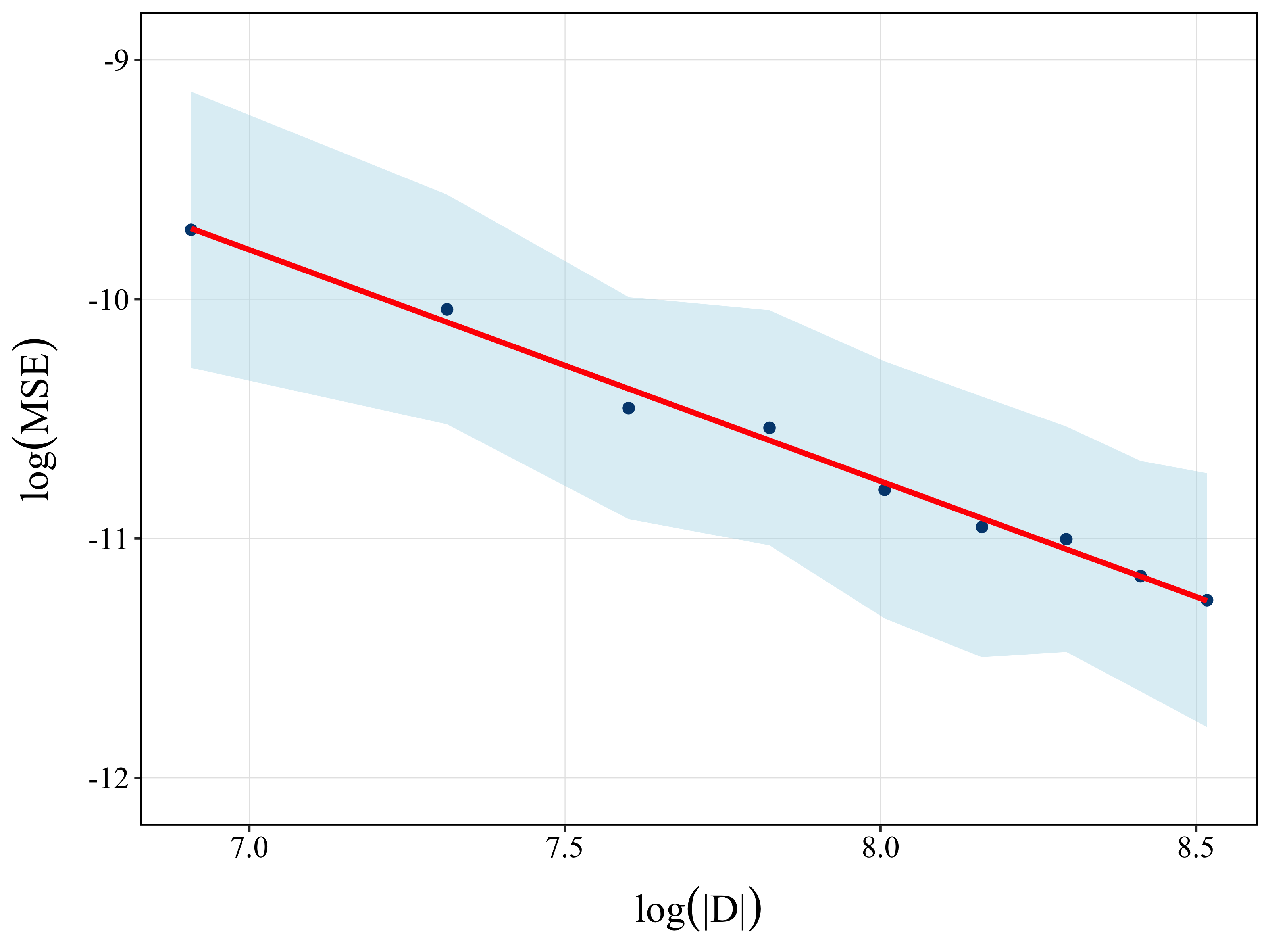}
        \caption{$\{r=0.6, \gamma=0.2\}$}
        \label{sim:fig7_b}
    \end{subfigure}
    
    \vspace{0.4cm}
    
    \begin{subfigure}[b]{0.45\linewidth}
        \centering
        \includegraphics[width=\linewidth]{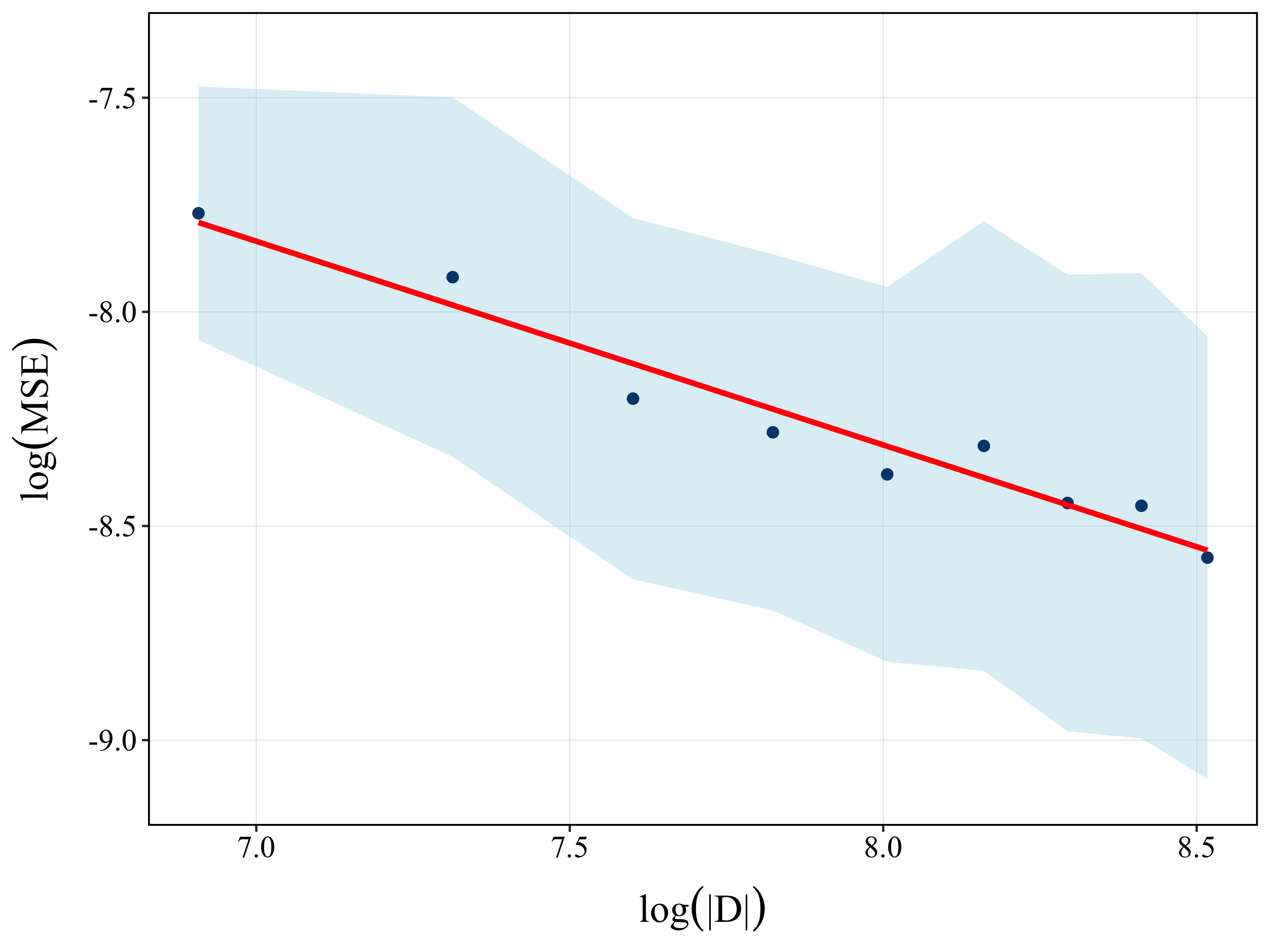}
        \caption{$\{r=0.5, \gamma=0.45\}$}
        \label{sim:fig7_c}
    \end{subfigure}
    \hfill
    \begin{subfigure}[b]{0.45\linewidth}
        \centering
        \includegraphics[width=\linewidth]{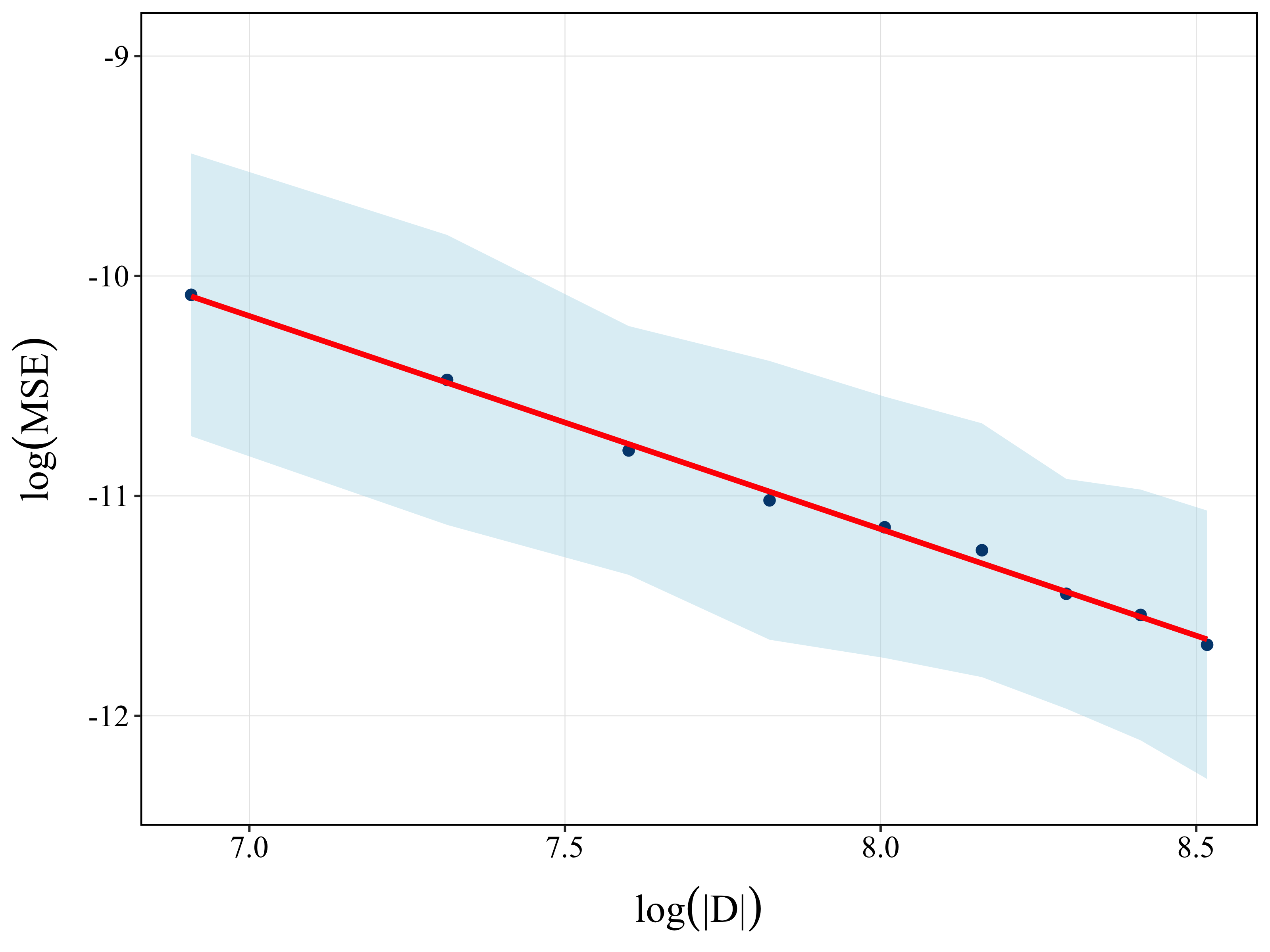}
        \caption{$\{r=0.4, \gamma=0.1\}$}
        \label{sim:fig7_d}
    \end{subfigure}
    
    \caption{log MSE vs log number of training data $|D|$ for NESR with different $r$ and $\gamma$.}
    \label{sim:fig7}
\end{figure}

To empirically validate the theoretical learning rates established in Theorem \ref{thm2}, we plot the logarithm of the MSE against the logarithm of the training sample size, for $|D| \in \{1000, 1500, \dots, 5000\}$. Figure \ref{sim:fig7} presents these plots for NESR under the four distinct parameter configurations.

According to our theoretical analysis, the expected MSE decays at a polynomial rate of $\mathcal{O}(|D|^{-\frac{2r}{2r+\gamma}})$. As illustrated in Figure \ref{sim:fig7}, For the settings
$
    (r,\gamma)\in
    \{(0.8,0.2),(0.6,0.2),(0.5,0.45)\},
$
which satisfy $2r+\gamma>1$, the log--log plots show an approximately linear
decreasing trend with slope $-\frac{2r}{2r+\gamma}$ as the training sample size increases. This behavior is 
consistent with a polynomial decay of the prediction error. The setting
$(r,\gamma)=(0.4,0.1)$ satisfies $2r+\gamma=0.9<1$ and therefore lies outside
the regime covered by Theorem~\ref{thm2}, we include it only as an additional numerical
comparison.

\section{Real Data Analysis}

To examine the empirical performance of NESR on real data, we consider six
benchmark datasets from the UCI Machine Learning Repository%
\footnote{\url{https://archive.ics.uci.edu/}}
and OpenML%
\footnote{\url{https://www.openml.org/}}.
Among them, KIN8NM, SPCD, and PPOPTS are regression datasets with continuous responses,
whereas SUSY, HTRU2, and MGT are binary classification datasets. We use prediction error for the regression tasks and classification error for the 
classification tasks. The input variables in all six datasets are continuous.
We compare NESR with LP in terms of both prediction performance and
computational cost. In this section, we use the Gaussian kernel $K(x, x') = \exp(-\Vert x - x' \Vert^2 / 2\tau^2)$. The corresponding random Fourier is $\sqrt{2}\cos(w^T\bfx+b)$ where $w$ and $b$ are sampled from $ w\sim\mathcal{N}\left(0,\tau^{-2}I_d\right)$ and $b \sim U(0,2\pi)$.where $d$ denotes the input dimension. For each dataset, the bandwidth
parameter $\tau$ is selected by five-fold cross-validation over the grid $\{2^{-5}, 2^{-4.5}, \dots, 2^5\}$. 

We consider two experimental settings. In the first, the number of random
features $M$ is fixed while the training sample size $|D|$ is varied. In the
second, $|D|$ is fixed while $M$ is varied. Preliminary experiments indicated
that restricting the candidate ranges to the regions used in the runtime
comparison produced prediction errors close to those obtained from the broader
search ranges. We therefore use the same runtime-oriented settings when
reporting both prediction error and computational cost in this section. More
specifically, for LP we take $\Lambda_{LP} := \{q^k : k = 10, 11, \dots, 20\}$, and $\Lambda_{ES} \subseteq [10^{-6}, 5 \times 10^{-4}]$ with $h = 1000$ and $K_{ES} = 1000$ for NESR. The main characteristics of the datasets and the parameter settings used in
the experiments are summarized in Table~\ref{tab:real_data_settings},
including the input dimension $d$, the test sample size $|D'|$, the selected
bandwidth $\sigma$, and the fixed value of either $M$ or $|D|$. Because SUSY
contains substantially more observations than the other datasets, we use a
random subsample for this dataset. The remaining five datasets are used without
additional subsampling.

\begin{table}
    \centering
    \caption{Parameter settings and dataset characteristics.}
    \label{tab:real_data_settings}
    \begin{tabular}{@{}lcccccccc@{}}
        \toprule
        & datasets & $d$ & $|D'|$ & $|D|$ & $\tau$ & $M$ & $C_{ES}$ & $C_{LP}$ \\
        \midrule
        Fig.~\ref{sim:fig8}:(a) & KIN8NM & 8  & 1192 & -     & 2         & 1000 & 0.0250 & 0.0001 \\
        Fig.~\ref{sim:fig8}:(b) & SPCD   & 81 & 3263 & -     & 8         & 1000 & 0.0400 & 0.0003 \\
        Fig.~\ref{sim:fig8}:(c) & PPOPTS & 9  & 5730 & -     & 1         & 1000 & 0.0800 & 0.0010 \\
        Fig.~\ref{sim:fig8}:(d) & SUSY   & 18 & 2000 & -     & $2^{2.5}$ & 1000 & 0.0750 & 0.0001 \\
        Fig.~\ref{sim:fig8}:(e) & HTRU2  & 8  & 3808 & -     & 2         & 1000 & 0.0200 & 0.0010 \\
        Fig.~\ref{sim:fig8}:(f) & MGT    & 10 & 3020 & -     & $2^{1.5}$ & 1000 & 0.0500 & 0.0030 \\
        \addlinespace
        Fig.~\ref{sim:fig9}:(a) & KIN8NM & 8  & 1192 & 7000  & 2         & -    & 0.0008 & 0.0001 \\
        Fig.~\ref{sim:fig9}:(b) & SPCD   & 81 & 3263 & 18000 & 8         & -    & 0.0009 & 0.0002 \\
        Fig.~\ref{sim:fig9}:(c) & PPOPTS & 9  & 5730 & 40000 & 1         & -    & 0.0015 & 0.0012 \\
        Fig.~\ref{sim:fig9}:(d) & SUSY   & 18 & 2000 & 14000 & $2^{2.5}$ & -    & 0.0040 & 0.0003 \\
        Fig.~\ref{sim:fig9}:(e) & HTRU2  & 8  & 3808 & 14000 & 2         & -    & 0.0100 & 0.0002 \\
        Fig.~\ref{sim:fig9}:(f) & MGT    & 10 & 3020 & 12000 & $2^{1.5}$ & -    & 0.0030 & 0.0100 \\
        \bottomrule
    \end{tabular}
\end{table}

\begin{figure}[htbp!]
    \centering
    % Varying N
    % ---- KIN8NM, SPCD, PPOPTS ----
    \begin{subfigure}[b]{0.32\linewidth}
        \centering
        \includegraphics[width=\linewidth]{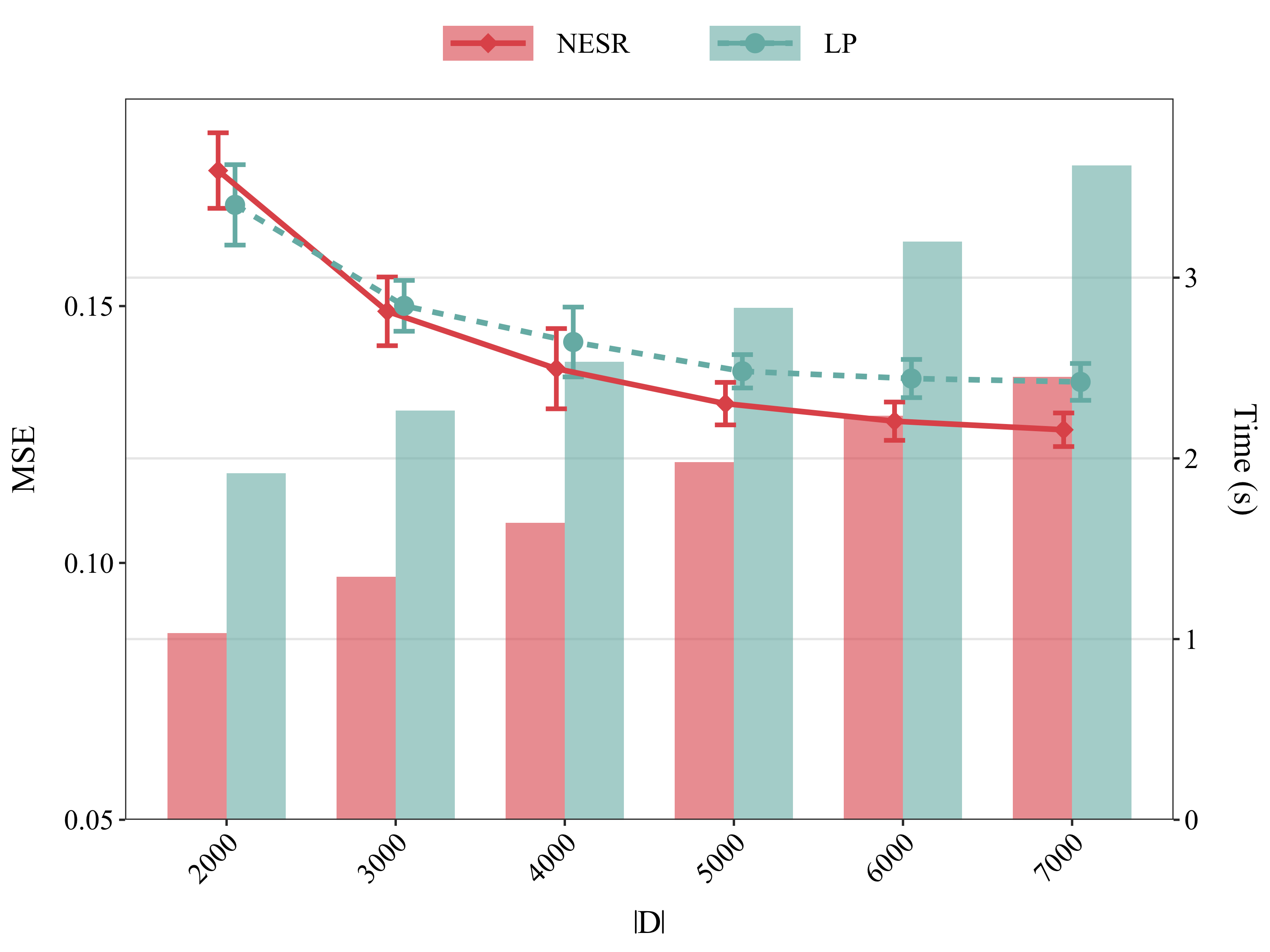}
        \caption{KIN8NM}
        \label{sim:fig8_a}
    \end{subfigure}
    \hfill
    \begin{subfigure}[b]{0.32\linewidth}
        \centering
        \includegraphics[width=\linewidth]{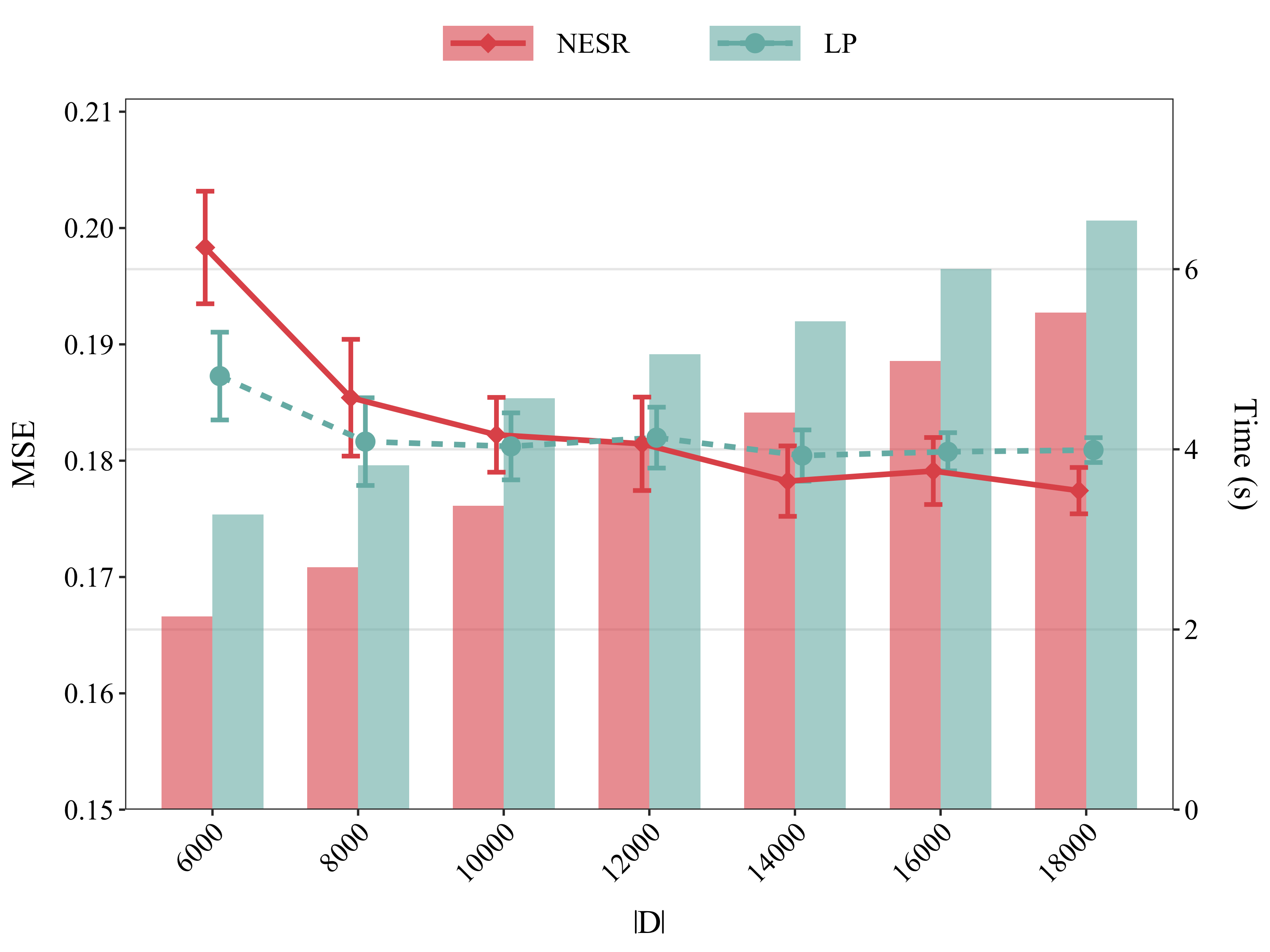}
        \caption{SPCD}
        \label{sim:fig8_b}
    \end{subfigure}
    \hfill
    \begin{subfigure}[b]{0.32\linewidth}
        \centering
        \includegraphics[width=\linewidth]{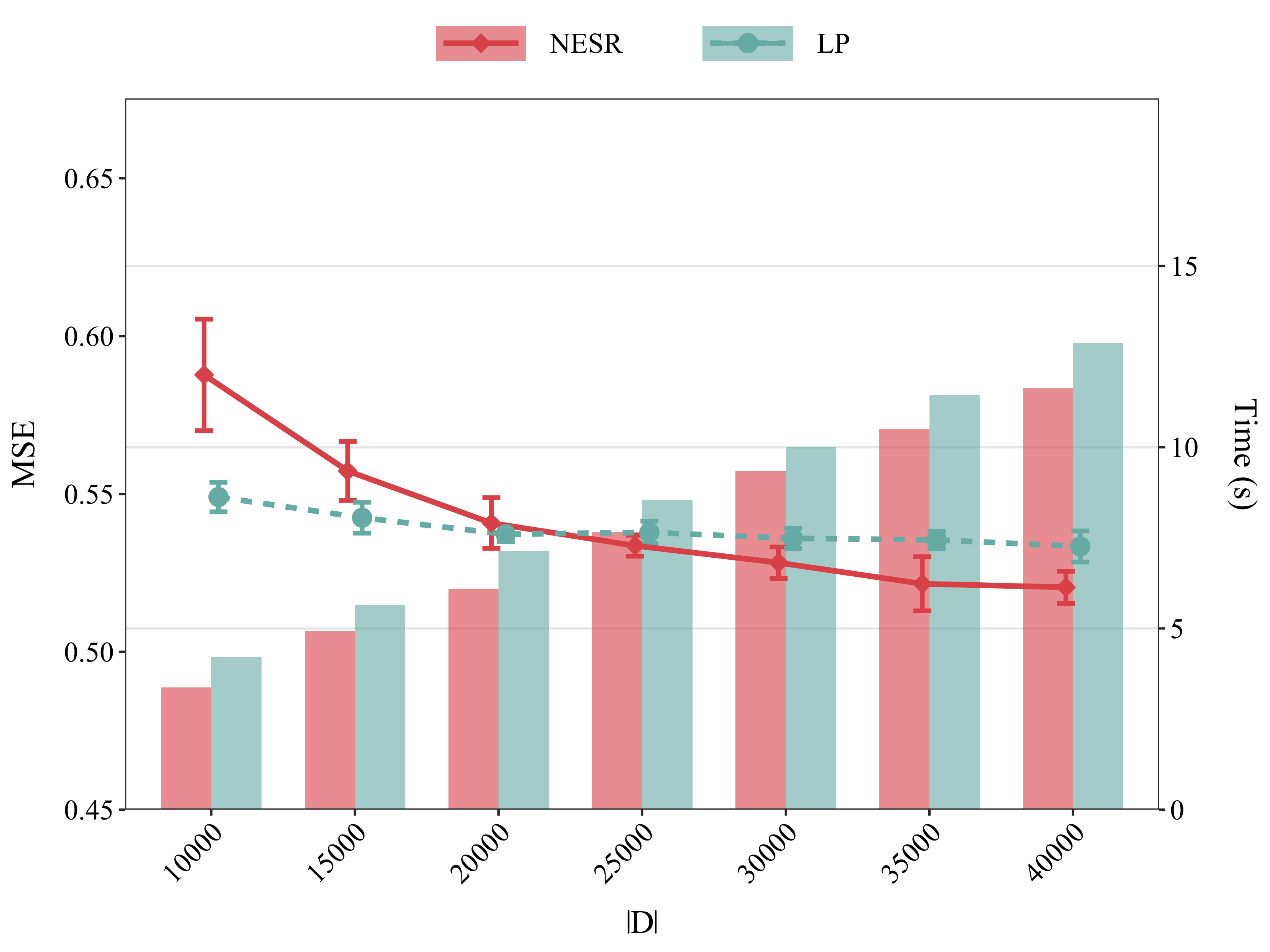}
        \caption{PPOPTS}
        \label{sim:fig8_c}
    \end{subfigure}
    
    \vspace{0.4cm} % 行间距
    
    % ----  SUSY, HTRU2, MGT ----
    \begin{subfigure}[b]{0.32\linewidth}
        \centering
        \includegraphics[width=\linewidth]{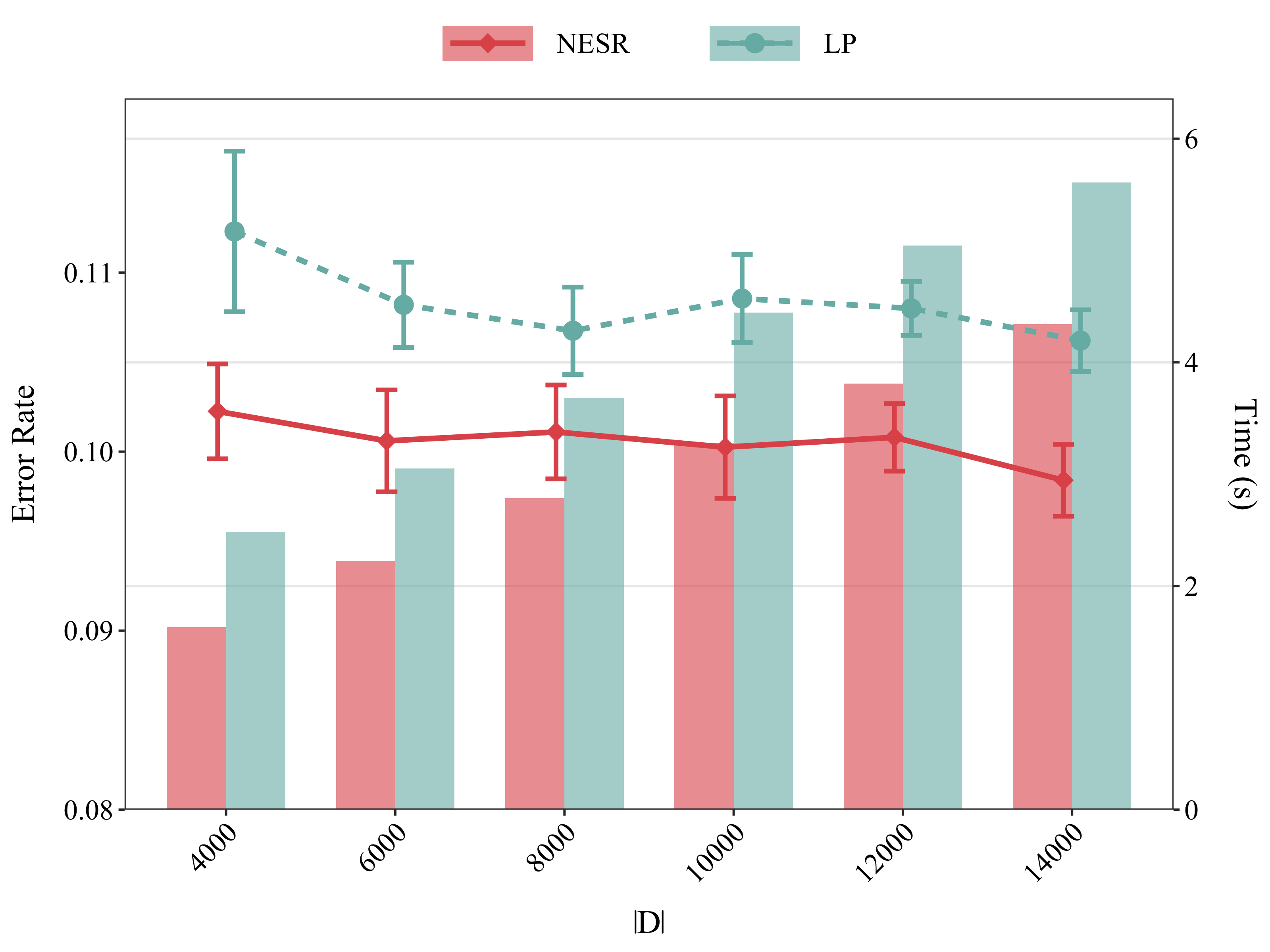}
        \caption{SUSY}
        \label{sim:fig8_d}
    \end{subfigure}
    \hfill
    \begin{subfigure}[b]{0.32\linewidth}
        \centering
        \includegraphics[width=\linewidth]{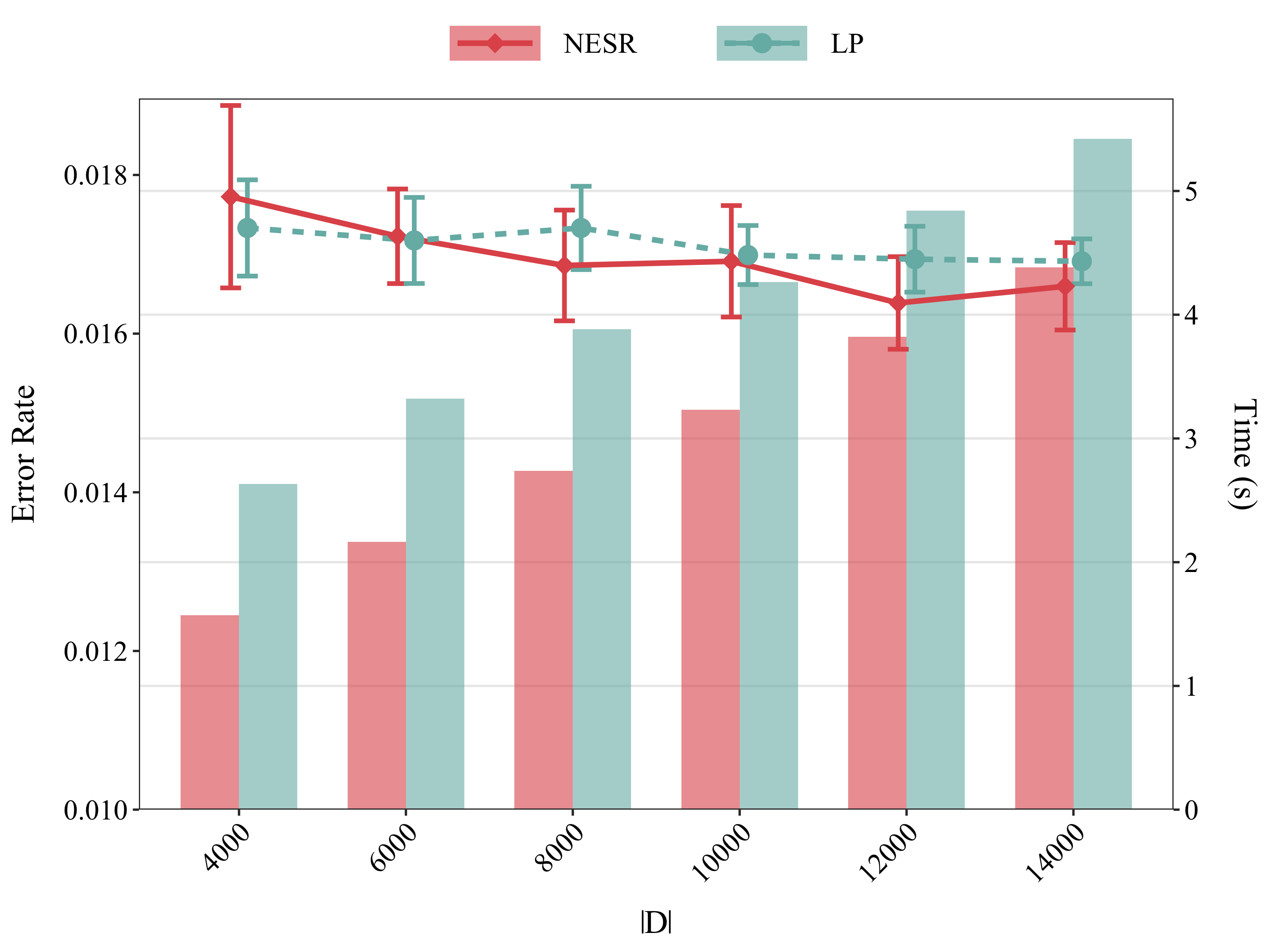}
        \caption{HTRU2}
        \label{sim:fig8_e}
    \end{subfigure}
    \hfill
    \begin{subfigure}[b]{0.32\linewidth}
        \centering
        \includegraphics[width=\linewidth]{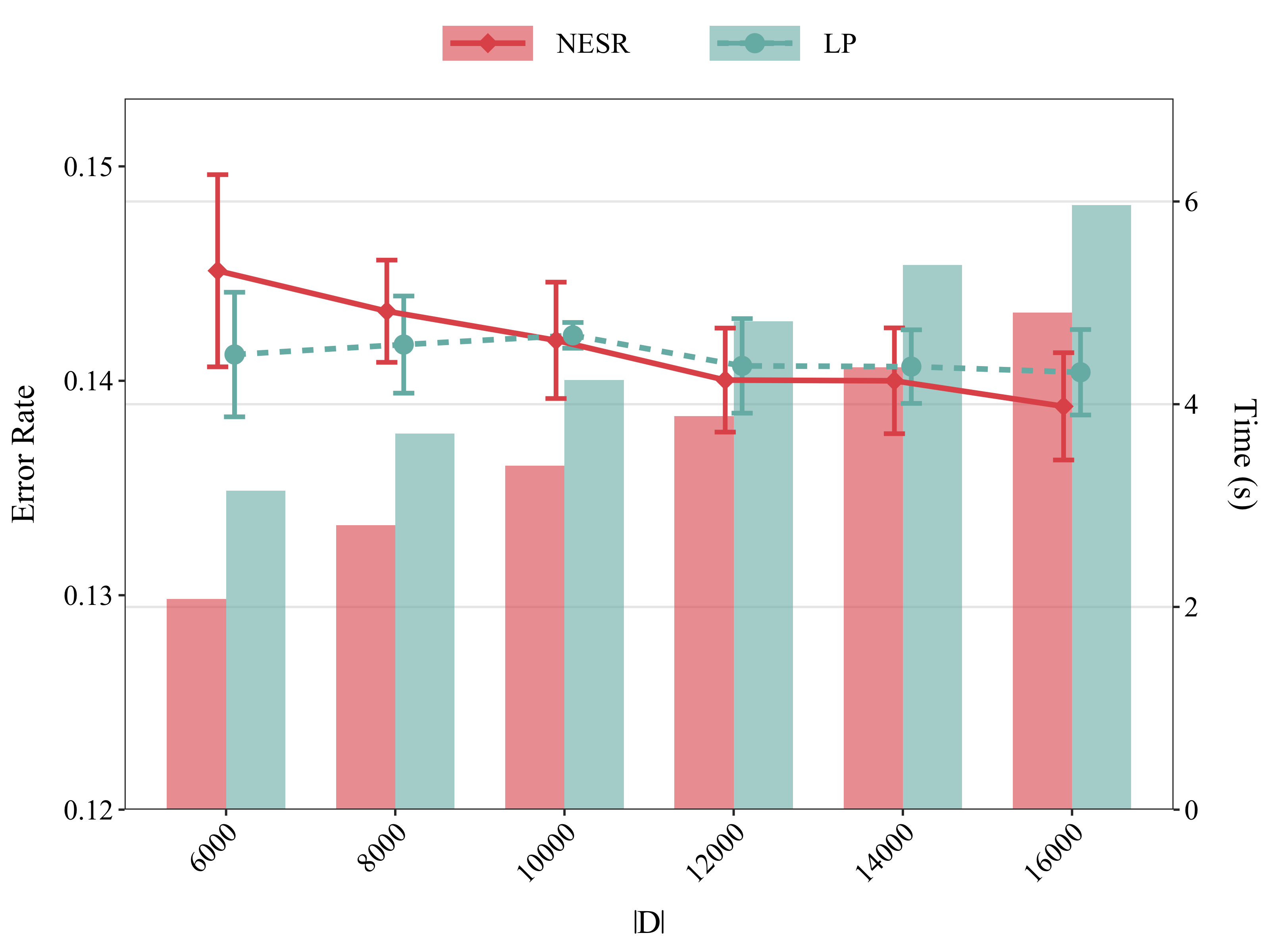}
        \caption{MGT}
        \label{sim:fig8_f}
    \end{subfigure}
    
    \caption{Prediction accuracy and runtime vs the training sample size $|D|$ on real datasets. Top panel: Regression tasks, where the line plots represent the MSE and the bar plots represent the average runtime over 10 independent trials. Bottom panel: Classification tasks, where the line plots represent the classification error rate and the bar plots represent the average runtime over 10 independent trials.}
    \label{sim:fig8}
\end{figure}

\begin{figure}[htbp!]
    % Varying M 
    \centering
    % ---- KIN8NM, SPCD, PPOPTS ----
    \begin{subfigure}[b]{0.32\linewidth}
        \centering
        \includegraphics[width=\linewidth]{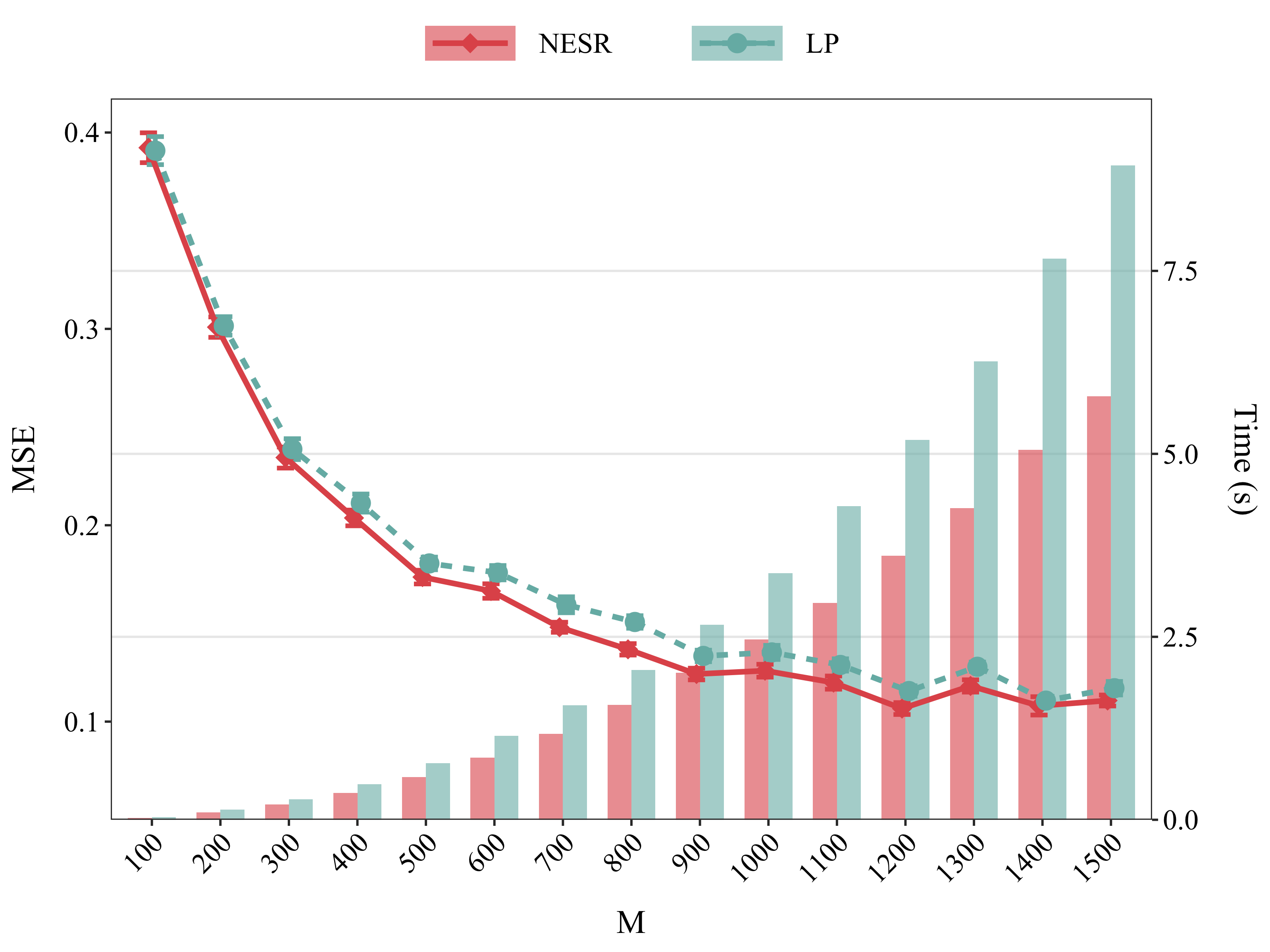}
        \caption{KIN8NM}
        \label{sim:fig9_a}
    \end{subfigure}
    \hfill
    \begin{subfigure}[b]{0.32\linewidth}
        \centering
        \includegraphics[width=\linewidth]{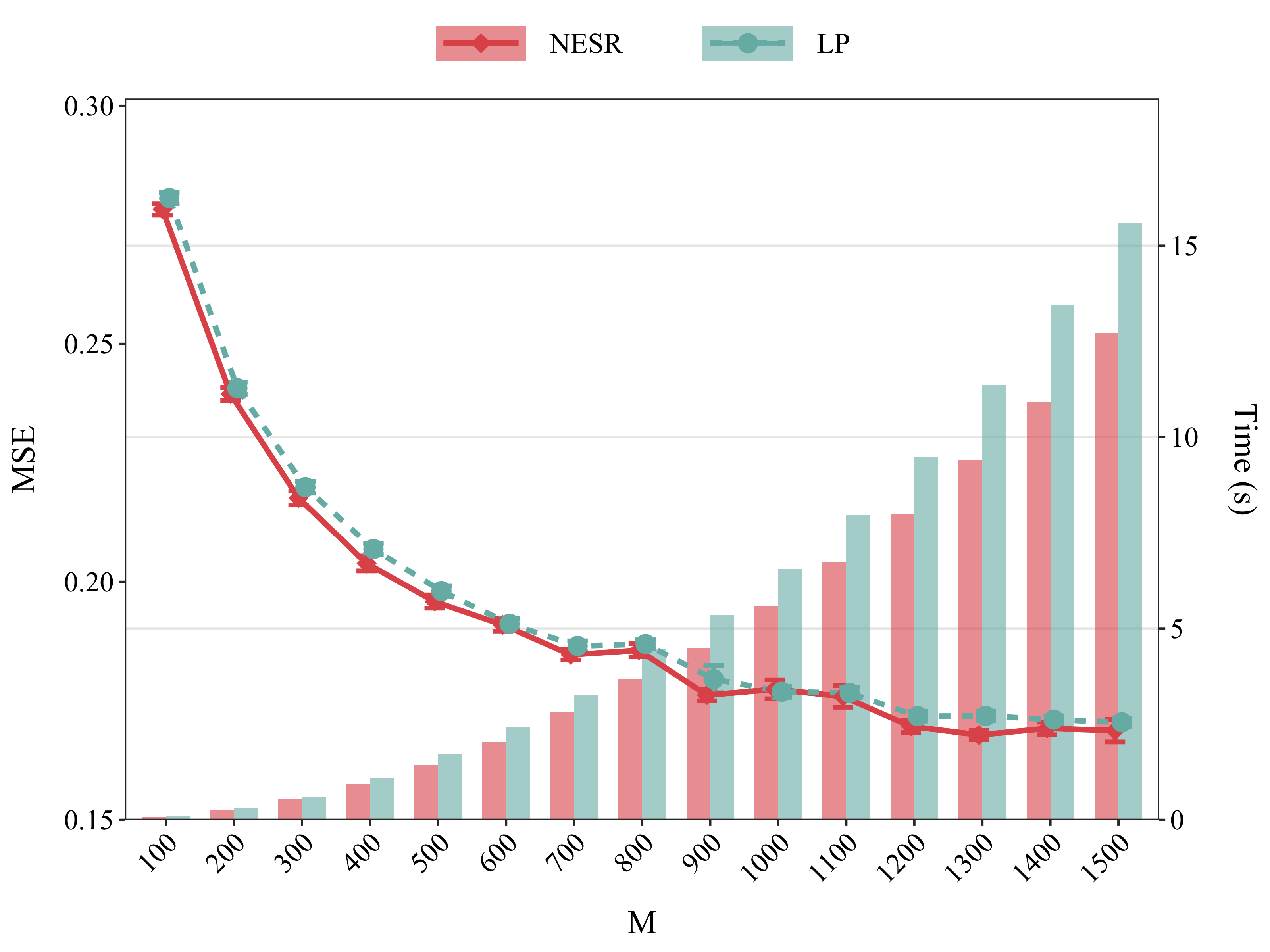}
        \caption{SPCD}
        \label{sim:fig9_b}
    \end{subfigure}
    \hfill
    \begin{subfigure}[b]{0.32\linewidth}
        \centering
        \includegraphics[width=\linewidth]{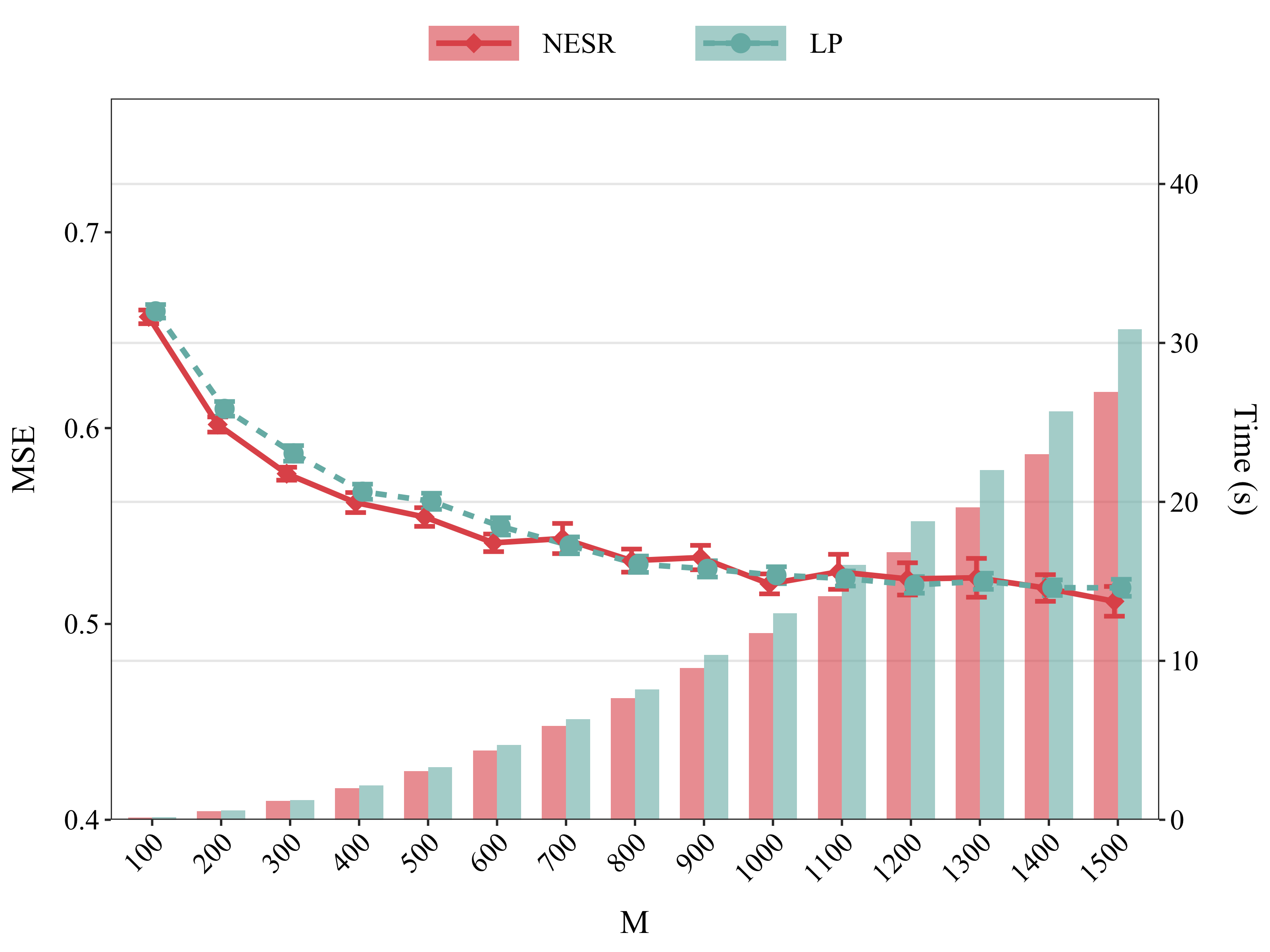}
        \caption{PPOPTS}
        \label{sim:fig9_c}
    \end{subfigure}
    
    \vspace{0.4cm}
    
    % ---- SUSY, HTRU2, MGT ----
    \begin{subfigure}[b]{0.32\linewidth}
        \centering
        \includegraphics[width=\linewidth]{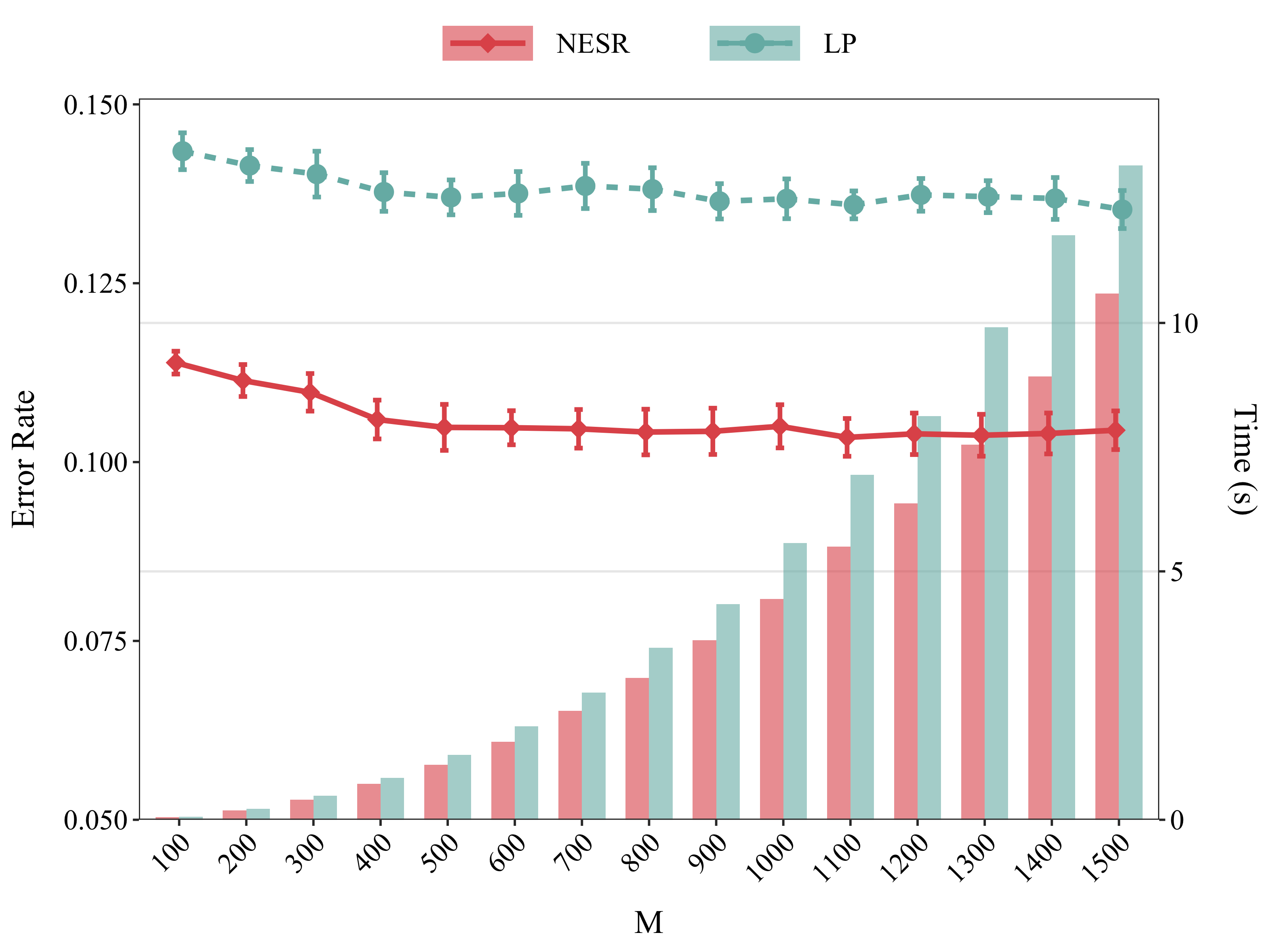}
        \caption{SUSY}
        \label{sim:fig9_d}
    \end{subfigure}
    \hfill
    \begin{subfigure}[b]{0.32\linewidth}
        \centering
        \includegraphics[width=\linewidth]{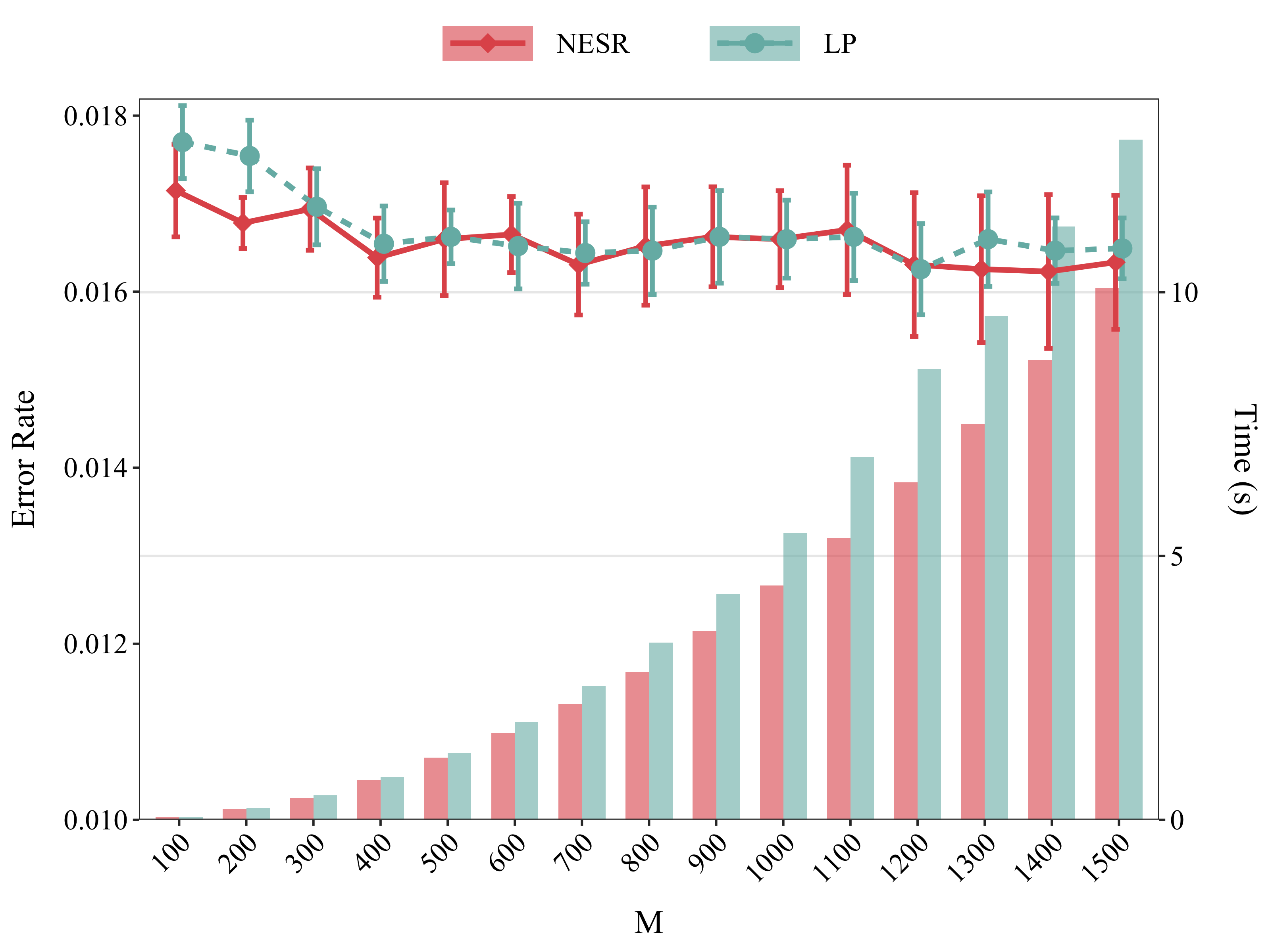}
        \caption{HTRU2}
        \label{sim:fig9_e}
    \end{subfigure}
    \hfill
    \begin{subfigure}[b]{0.32\linewidth}
        \centering
        \includegraphics[width=\linewidth]{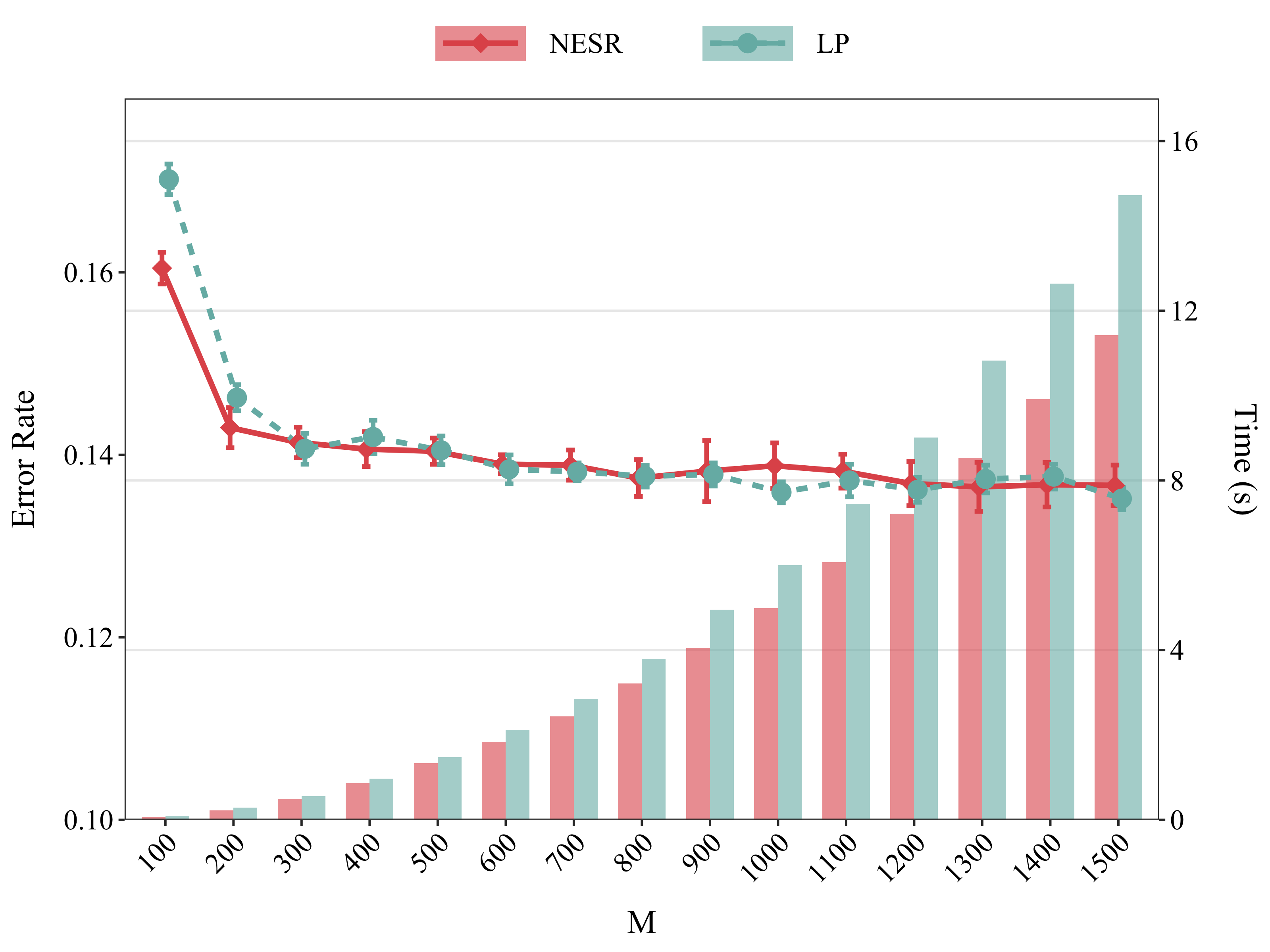}
        \caption{MGT}
        \label{sim:fig9_f}
    \end{subfigure}
    
    \caption{Prediction accuracy and runtime vs the number of random features $M$ on real datasets. Top panel: Regression tasks, where the line plots represent the MSE and the bar plots represent the average runtime over 10 independent trials. Bottom panel: Classification tasks, where the line plots represent the classification error rate and the bar plots represent the average runtime over 10 independent trials.}
    \label{sim:fig9}
\end{figure}

Figures~\ref{sim:fig8} and~\ref{sim:fig9} summarize the empirical performance
of NESR and LP on the six real-world datasets, with all results averaged over
10 independent trials. Figure~\ref{sim:fig8} reports the results obtained by
varying the training sample size $|D|$. For most datasets, the prediction error
of both methods decreases as the training sample size increases. NESR generally
achieves prediction accuracy comparable to or better than LP over the range of
sample sizes considered. Its recorded running time is also lower than that of
LP in most of the reported experiments. The error bars of the two methods are
of similar magnitude across the datasets.

Figure~\ref{sim:fig9} shows the corresponding results as the number of random
features $M$ varies. For the three regression datasets, the prediction error
changes relatively little once $M$ becomes moderately large, with the curves
largely stabilizing around $M=1000$. A similar pattern is observed for the
classification datasets, where the performance tends to level off at somewhat
smaller values of $M$. Across the values of $M$ considered, NESR generally
achieves prediction performance comparable to or better than LP, while requiring
less running time in the reported implementations. Overall, these experiments
suggest that the neighboring comparison rule can provide a favorable balance
between prediction accuracy and computational cost on the datasets considered.

\section{Conclusion}\label{sec-conc}

This paper develops a neighboring early-stopping rule for adaptive
regularization in kernel ridge regression with random features. The proposed
NESR-KRR-RF method compares only adjacent estimators along a grid that is
uniform in inverse regularization, reducing the number of discrepancy
comparisons relative to standard all-pairs Lepskii-type procedures. Both the
neighboring discrepancy and its empirical complexity term can be computed in
the random feature space without constructing the exact kernel Gram matrix. We establish a high-probability comparison bound for neighboring KRR-RF
estimators and show that, under source, capacity, grid-coverage, and
random feature budget conditions, the selected estimator attains the oracle
polynomial learning rate up to logarithmic factors. The result covers both
well-specified and partially misspecified regimes and allows the regularization
parameter to be selected without prior knowledge of the source and capacity
exponents. The numerical experiments illustrate that NESR can achieve
prediction performance comparable to the benchmark methods while requiring
fewer discrepancy comparisons and, in the reported implementations, less
computation. Several questions remain open. In particular, it would be useful to develop
fully data-driven calibration of the stopping threshold and the random feature
budget, and to extend the neighboring comparison principle to other loss
functions and scalable random feature constructions.

\bibliographystyle{abbrv}
\bibliography{ref.bib}

@article{rudi2017generalization,
  title={Generalization properties of learning with random features},
  author={Rudi, Alessandro and Rosasco, Lorenzo},
  journal={Advances in Neural Information Processing Systems},
  volume={30},
  pages={3215--3225},
  year={2017}
}

@article{wang2026generalization,
  title={Generalization Properties of Robust Learning With Random Features},
  author={Wang, Caixing},
  journal={IEEE Transactions on Pattern Analysis and Machine Intelligence},
  year={2026},
  volume={48},
  number={7},
  pages={8199-8215},
  publisher={IEEE}
}

@article{wahba2019representer,
  title={Representer theorem},
  author={Wahba, Grace and Wang, Yuedong},
  journal={Wiley StatsRef: Statistics Reference Online},
  volume={1155},
  pages={1178},
  year={2019}
}

@inproceedings{scholkopf2001generalized,
  title={A generalized representer theorem},
  author={Sch{\"o}lkopf, Bernhard and Herbrich, Ralf and Smola, Alex J},
  booktitle={International Conference on Computational Learning Theory},
  pages={416--426},
  year={2001},
  organization={Springer}
}

@article{blanchard2019lepskii,
  title={Lepskii principle in supervised learning},
  author={Blanchard, Gilles and Math{\'e}, Peter and M{\"u}cke, Nicole},
  journal={arXiv preprint arXiv:1905.10764},
  year={2019}
}

@article{de2010adaptive,
  title={Adaptive kernel methods using the balancing principle},
  author={De Vito, Ernesto and Pereverzyev, Sergei and Rosasco, Lorenzo},
  journal={Foundations of Computational Mathematics},
  volume={10},
  number={4},
  pages={455--479},
  year={2010},
  publisher={Springer}
}

@book{murphy2012machine,
  title={Machine learning: a probabilistic perspective},
  author={Murphy, Kevin P},
  year={2012},
  publisher={MIT Press}
}

@article{lepskii1991problem,
  title={On a problem of adaptive estimation in Gaussian white noise},
  author={Lepskii, OV},
  journal={Theory of Probability \& Its Applications},
  volume={35},
  number={3},
  pages={454--466},
  year={1991},
  publisher={SIAM}
}

@book{engl1996regularization,
  title={Regularization of inverse problems},
  author={Engl, Heinz Werner and Hanke, Martin and Neubauer, Andreas},
  volume={375},
  year={1996},
  publisher={Springer Science \& Business Media}
}

@article{caponnetto2010cross,
  title={Cross-validation based adaptation for regularization operators in learning theory},
  author={Caponnetto, Andrea and Yao, Yuan},
  journal={Analysis and Applications},
  volume={8},
  number={02},
  pages={161--183},
  year={2010},
  publisher={World Scientific}
}

@article{lu2020balancing,
  title={Balancing principle in supervised learning for a general regularization scheme},
  author={Lu, Shuai and Math{\'e}, Peter and Pereverzev, Sergei V},
  journal={Applied and Computational Harmonic Analysis},
  volume={48},
  number={1},
  pages={123--148},
  year={2020},
  publisher={Elsevier}
}

@article{celisse2021analyzing,
  title={Analyzing the discrepancy principle for kernelized spectral filter learning algorithms},
  author={Celisse, Alain and Wahl, Martin},
  journal={Journal of Machine Learning Research},
  volume={22},
  number={76},
  pages={1--59},
  year={2021}
}

@book{gyorfi2002distribution,
  title={A distribution-free theory of nonparametric regression},
  author={Gy{\"o}rfi, L{\'a}szl{\'o} and Kohler, Michael and Krzy{\.z}ak, Adam and Walk, Harro},
  year={2002},
  publisher={Springer}
}

@article{browne2000cross,
  title={Cross-validation methods},
  author={Browne, Michael W},
  journal={Journal of Mathematical Psychology},
  volume={44},
  number={1},
  pages={108--132},
  year={2000},
  publisher={Elsevier}
}

@article{lin2024adaptive,
  title={Adaptive parameter selection for kernel ridge regression},
  author={Lin, Shao-Bo},
  journal={Applied and Computational Harmonic Analysis},
  volume={73},
  pages={101671},
  year={2024},
  publisher={Elsevier}
}

@article{liu2026beyond,
  title={Beyond Cross-Validation: Adaptive Parameter Selection for Kernel-Based Gradient Descents},
  author={Liu, Xiaotong and Lei, Yunwen and Chang, Xiangyu and Lin, Shao-Bo},
  journal={arXiv preprint arXiv:2603.03401},
  year={2026}
}

@article{lin2024lepskii,
  title={Lepskii Principle for Distributed Kernel Ridge Regression},
  author={Lin, Shao-Bo},
  journal={arXiv preprint arXiv:2409.05070},
  year={2024}
}

@article{lin2025adaptive,
  title={Adaptive distributed kernel ridge regression: A feasible distributed learning scheme for data silos},
  author={Lin, Shao-Bo and Liu, Xiaotong and Wang, Di and Zhang, Hai and Zhou, Ding-Xuan},
  journal={Journal of Machine Learning Research},
  volume={26},
  number={108},
  pages={1--54},
  year={2025}
}

@article{smale2007learning,
  title={Learning theory estimates via integral operators and their approximations},
  author={Smale, Steve and Zhou, Ding-Xuan},
  journal={Constructive Approximation},
  volume={26},
  number={2},
  pages={153--172},
  year={2007},
  publisher={Springer}
}

@article{caponnetto2007optimal,
  title={Optimal rates for the regularized least-squares algorithm},
  author={Caponnetto, Andrea and De Vito, Ernesto},
  journal={Foundations of Computational Mathematics},
  volume={7},
  pages={331--368},
  year={2007},
  publisher={Springer}
}

@book{wahba1990spline,
  title={Spline models for observational data},
  author={Wahba, Grace},
  year={1990},
  publisher={SIAM}
}

@article{wang2024communication,
    author = {Wang, Caixing and Li, Tao and Zhang, Xinyi and Feng, Xingdong and He, Xin},
    title = {Communication-Efficient Nonparametric Quantile Regression via Random Features},
    journal = {Journal of Computational and Graphical Statistics},
    volume= {33},
    number= {4},
    pages = {1175--1184},
    year = {2024}
}

@book{rudin2017Fourier,
  title={{F}ourier analysis on groups},
  author={Rudin, Walter},
  year={2017},
  publisher={Courier Dover Publications}
}

@article{rahimi2007random,
  title={Random features for large-scale kernel machines},
  author={Rahimi, Ali and Recht, Benjamin},
  journal={Advances in Neural Information Processing Systems},
  volume={20},
  pages={1177--1184},
  year={2007}
}

@article{zhang2015divide,
  title={Divide and conquer kernel ridge regression: A distributed algorithm with minimax optimal rates},
  author={Zhang, Yuchen and Duchi, John and Wainwright, Martin},
  journal={The Journal of Machine Learning Research},
  volume={16},
  number={1},
  pages={3299--3340},
  year={2015}
}

@book{scholkopf2002learning,
  title={Learning with kernels: support vector machines, regularization, optimization, and beyond},
  author={Sch{\"o}lkopf, Bernhard and Smola, Alexander J},
  year={2002},
  publisher={MIT Press}
}

@article{williams2001using,
  title={Using the Nystroem Method to Speed Up Kernel Machines},
  author={Williams, Christopher and Seeger, Matthias},
  journal={Advances in Neural Information Processing Systems},
  pages={682--688},
  year={2001}
}

@inproceedings{wang2024optimal,
  title={Optimal Kernel Quantile Learning with Random Features},
  author={Wang, Caixing and Feng, Xingdong},
  booktitle={International Conference on Machine Learning},
  pages={50419--50452},
  year={2024},
  organization={PMLR}
}

@article{lin2020distributed,
  title={Distributed Kernel Ridge Regression with Communications},
  author={Lin, Shao-Bo and Wang, Di and Zhou, Ding-Xuan},
  journal={The Journal of Machine Learning Research},
  volume={21},
  pages={3718--3755},
  year={2020}
}

@article{li2021towards,
  title={Towards a unified analysis of random {F}ourier features},
  author={Li, Zhu and Ton, Jean-Francois and Oglic, Dino and Sejdinovic, Dino},
  journal={The Journal of Machine Learning Research},
  volume={22},
  number={1},
  pages={4887--4937},
  year={2021},
  publisher={JMLRORG}
}

@article{gerfo2008spectral,
  title={Spectral algorithms for supervised learning},
  author={Gerfo, L Lo and Rosasco, Lorenzo and Odone, Francesca and Vito, E De and Verri, Alessandro},
  journal={Neural Computation},
  volume={20},
  number={7},
  pages={1873--1897},
  year={2008}
}

@article{lin2020spectral,
  title={Optimal rates for spectral algorithms with least-squares regression over Hilbert spaces},
  author={Lin, Junhong and Rudi, Alessandro and Rosasco, Lorenzo and Cevher, Volkan},
  journal={Applied and Computational Harmonic Analysis},
  volume={48},
  number={3},
  pages={868--890},
  year={2020},
  publisher={Elsevier}
}

@article{lin2017distributed,
  title={Distributed learning with regularized least squares},
  author={Lin, Shao-Bo and Guo, Xin and Zhou, Ding-Xuan},
  journal={The Journal of Machine Learning Research},
  volume={18},
  number={1},
  pages={3202--3232},
  year={2017}
}

@article{rudi2015less,
  title={Less is more: Nystr{\"o}m computational regularization},
  author={Rudi, Alessandro and Camoriano, Raffaello and Rosasco, Lorenzo},
  journal={Advances in Neural Information Processing Systems},
  volume={28},
  pages={1657--1665},
  year={2015}
}

@article{lin2020optimal,
  title={Optimal convergence for distributed learning with stochastic gradient methods and spectral algorithms},
  author={Lin, Junhong and Cevher, Volkan},
  journal={The Journal of Machine Learning Research},
  volume={21},
  number={1},
  pages={5852--5914},
  year={2020}
}

@article{lin2016optimal,
  title={Optimal learning for multi-pass stochastic gradient methods},
  author={Lin, Junhong and Rosasco, Lorenzo},
  journal={Advances in Neural Information Processing Systems},
  volume={29},
  pages={4563--4571},
  year={2016}
}

@book{steinwart2008support,
  title={Support vector machines},
  author={Steinwart, Ingo and Christmann, Andreas},
  year={2008},
  publisher={Springer Science \& Business Media}
}

@article{bach2017equivalence,
  title={On the equivalence between kernel quadrature rules and random feature expansions},
  author={Bach, Francis},
  journal={The Journal of Machine Learning Research},
  volume={18},
  number={1},
  pages={714--751},
  year={2017}
}

@article{rahimi2008weighted,
  title={Weighted sums of random kitchen sinks: Replacing minimization with randomization in learning},
  author={Rahimi, Ali and Recht, Benjamin},
  journal={Advances in Neural Information Processing Systems},
  volume={21},
  pages={1313--1320},
  year={2008}
}

@article{ma2025generalization,
  title={On the Generalization Properties of Learning the Random Feature Models with Learnable Activation Functions},
  author={Ma, Zailin and Yang, Jiansheng and Yang, Yaodong},
  journal={arXiv preprint arXiv:2510.15327},
  year={2025}
}

@article{sonnenburg2006large,
  title= {Large-scale multiple kernel learning},
  author={Sonnenburg, S{\"o}ren and R{\"a}tsch, Gunnar and Sch{\"a}fer, Christin and Sch{\"o}lkopf, Bernhard},
  journal={The Journal of Machine Learning Research},
  volume={7},
  pages={1531--1565},
  year={2006}
}

@article{camps2005kernel,
  title={Kernel-based methods for hyperspectral image classification},
  author={Camps-Valls, Gustavo and Bruzzone, Lorenzo},
  journal={IEEE Transactions on Geoscience and Remote Sensing},
  volume={43},
  number={6},
  pages={1351--1362},
  year={2005},
  publisher={IEEE}
}

@inproceedings{zhao2009multiple,
  title={Multiple kernel clustering},
  author={Zhao, Bin and Kwok, James T and Zhang, Changshui},
  booktitle={Proceedings of the 2009 SIAM International Conference on Data Mining},
  pages={638--649},
  year={2009},
  organization={SIAM}
}

@article{camastra2005novel,
  title={A novel kernel method for clustering},
  author={Camastra, Francesco and Verri, Alessandro},
  journal={IEEE Transactions on Pattern Analysis and Machine Intelligence},
  volume={27},
  number={5},
  pages={801--805},
  year={2005},
  publisher={IEEE}
}

@article{hofmann2008kernel,
  title={Kernel methods in machine learning},
  author={Hofmann, Thomas and Sch{\"o}lkopf, Bernhard and Smola, Alexander J},
  journal={The Annals of Statistics},
  volume={36},
  number={3},
  pages={1171--1220},
  year={2008}
}

@article{avron2017faster,
  title={Faster kernel ridge regression using sketching and preconditioning},
  author={Avron, Haim and Clarkson, Kenneth L and Woodruff, David P},
  journal={SIAM Journal on Matrix Analysis and Applications},
  volume={38},
  number={4},
  pages={1116--1138},
  year={2017},
  publisher={SIAM}
}

@article{rudi2017falkon,
  title={Falkon: An optimal large-scale kernel method},
  author={Rudi, Alessandro and Carratino, Luigi and Rosasco, Lorenzo},
  journal={Advances in Neural Information Processing Systems},
  volume={30},
  year={2017},
  pages={3891--3901}
}

@article{devito2005learning,
  title={Learning from examples as an inverse problem},
  author={De Vito, Ernesto and Rosasco, Lorenzo and Caponnetto, Andrea and De Giovannini, Umberto and Odone, Francesca},
  journal={Journal of Machine Learning Research},
  volume={6},
  pages={883--904},
  year={2005}
}

@article{bauer2007regularization,
  title={Regularization for ill-posed problems and learning algorithms},
  author={Bauer, Frank and Pereverzev, Sergei V and Rosasco, Lorenzo},
  journal={Journal of Complexity},
  volume={23},
  number={1},
  pages={52--72},
  year={2007},
  publisher={Elsevier}
}

@article{blanchard2018optimal,
  title={Optimal rates for regularization of statistical inverse learning problems},
  author={Blanchard, Gilles and M{\"u}cke, Nicole},
  journal={Foundations of Computational Mathematics},
  volume={18},
  number={4},
  pages={971--1013},
  year={2018},
  publisher={Springer}
}

@article{golub1979generalized,
  title={Generalized cross-validation as a method for choosing a good ridge parameter},
  author={Golub, Gene H and Heath, Michael and Wahba, Grace},
  journal={Technometrics},
  volume={21},
  number={2},
  pages={215--223},
  year={1979},
  publisher={Taylor \& Francis}
}

@article{arlot2010survey,
  title={A survey of cross-validation procedures for model selection},
  author={Arlot, Sylvain and Celisse, Alain},
  journal={Statistics Surveys},
  volume={4},
  pages={40--79},
  year={2010},
  publisher={The American Statistical Association, the Bernoulli Society, the Institute of Mathematical Statistics, and the Statistical Society of Canada}
}

@article{blanchard2012discrepancy,
  title={The discrepancy principle for statistical inverse problems with application to kernel methods},
  author={Blanchard, Gilles and Kr{\"a}mer, Nicole and M{\"u}cke, Nicole},
  journal={Journal of Machine Learning Research},
  volume={13},
  pages={1705--1744},
  year={2012}
}

@article{yao2007early,
  title={On early stopping in gradient descent learning},
  author={Yao, Yuan and Rosasco, Lorenzo and Caponnetto, Andrea},
  journal={Constructive Approximation},
  volume={26},
  number={2},
  pages={289--315},
  year={2007},
  publisher={Springer}
}

@article{raskutti2014early,
  title={Early stopping and non-parametric regression: An optimal data-dependent stopping rule},
  author={Raskutti, Garvesh and Wainwright, Martin J and Yu, Bin},
  journal={Journal of Machine Learning Research},
  volume={15},
  number={1},
  pages={335--366},
  year={2014}
}

@article{blanchard2018early,
  title={Early stopping for statistical inverse problems via truncated SVD estimation and concentration inequalities},
  author={Blanchard, Gilles and Hoffmann, Marc and Reiss, Markus},
  journal={Electronic Journal of Statistics},
  volume={12},
  number={1},
  pages={3204--3235},
  year={2018},
  publisher={Institute of Mathematical Statistics and Bernoulli Society}
}

@article{blanchard2018optimaladapt,
  title={Optimal adaptation for early stopping in statistical inverse problems},
  author={Blanchard, Gilles and M{\"u}cke, Nicole},
  journal={SIAM/ASA Journal on Uncertainty Quantification},
  volume={6},
  number={3},
  pages={1040--1063},
  year={2018},
  publisher={SIAM}
}

@article{wei2019early,
  title={Early stopping for kernel boosting algorithms: A general analysis with localized complexities},
  author={Wei, Yuting and Yang, Fang and Wainwright, Martin J and Yu, Bin},
  journal={Advances in Neural Information Processing Systems},
  volume={32},
  pages={6065--6075},
  year={2019}
}

@article{page2018goldenshluger,
  title   = {The Goldenshluger--Lepski Method for Constrained Least-Squares Estimators over {RKHS}s},
  author  = {Page, Stephen and Gr{\"u}new{\"a}lder, Steffen},
  journal = {Bernoulli},
  volume  = {27},
  number  = {4},
  pages   = {2241--2266},
  year    = {2021} 
}

\end{document}